\documentclass[twoside,11pt]{article}
\usepackage{jmlr2e}
\usepackage[utf8]{inputenc}
\usepackage{float}
\usepackage{amssymb,amsmath,array}
\usepackage[usenames,dvipsnames]{color}
\usepackage{wrapfig}
\usepackage{subfig}
\usepackage[toc,page]{appendix}
\definecolor{darkblue}{rgb}{0.10196,0.05098,0.670588}
\usepackage{graphicx}
\graphicspath{{./pics/}{./}{pics/}}
\DeclareGraphicsExtensions{.pdf,.png,.jpg,.jpeg}
\makeatletter
\def\input@path{{./}{./methods/}{./introduction/}{./results/}{./discussion/}}
\makeatother
\usepackage{lmodern}
\usepackage{tabularx,multicol,booktabs,multirow,makecell}[width=\textwidth]
\usepackage{longtable,pdflscape}
\usepackage{fancyhdr}
\usepackage{algorithmicx}
\usepackage[ruled]{algorithm}
\usepackage[noend]{algpseudocode}
\usepackage{footnote}
\usepackage{xfrac}

\newcommand\rsmraise[1]{%
  \ifx#1\displaystyle .9\else
    \ifx#1\textstyle .9\else
      \ifx#1\scriptstyle .9\else
        -.45%
      \fi
    \fi
  \fi}
\usepackage{epstopdf}

\usepackage{enumitem}

\usepackage[]{hyperref} 
\hypersetup{pageanchor=true,pdftitle={AngleLVQ},pdfauthor={Ghosh},colorlinks=true,citecolor=darkblue,linkcolor=black,urlcolor=darkblue,anchorcolor=darkblue}

\title{Interpretable Models Capable of Handling Systematic Missingness in Imbalanced Classes and Heterogeneous Datasets}
\author{S.Ghosh, E.S.Baranowski, M.Biehl, W.Arlt, P.Tino, K.Bunte}
\author{\name Sreejita Ghosh$^{1,5,6}$ \email ghosh.sreejita@gmail.com, s.ghosh1@uu.nl \\
	\AND
	\name Elizabeth S Baranowski$^{2}$ \email e.s.baranowski@bham.ac.uk \\
	\AND
	\name Michael Biehl$^{1,2,3}$ \email m.biehl@rug.nl \\
     \AND
	\name Wiebke Arlt$^{2}$ \email w.arlt@bham.ac.uk \\
	\AND
	\name Peter Tino$^{4}$ \email P.Tino@cs.bham.ac.uk \\
	\AND
     \name Kerstin Bunte$^{1}$ \email k.bunte@rug.nl \\
       \addr $(1)$ Bernoulli Institute of Mathematics, Computer Science and Artificial Intelligence\\\
       University of Groningen,
       Groningen, 9747AG, The Netherlands \\
       $(2)$ Institute of Metabolism and Systems Research \\
	            The University of Birmingham, 
	            Edgbaston, Birmingham B15 2TT, United Kingdom\\
       $(3)$ Systems Modelling and Quantitative Biomedicine, Institute of Metabolism and Systems Research\\  The University of Birmingham, Edgbaston,  Birmingham B15 2TT, United Kingdom\\
       $(4)$  School of Computer Science\\
	            The University of Birmingham, Edgbaston, Birmingham B15 2TT, United Kingdom\\
	   $(5)$ One Health Chemical- Exposome, Institute for Risk Assessment Sciences,\\
	   Utrecht University, Utrecht 3584 CM, The Netherlands\\
	   $(6)$ Circulatory Health Research Center,\\
	   University Medical Centrum Utrecht, 3584 CX Utrecht, The Netherlands
       }
       \editor{}
       
\renewcommand{\vec}[1]{\boldsymbol{#1}}
\definecolor{RoyalBlue}{cmyk}{1, 0.50, 0, 0}

\newcommand{\R}{\rm I\!R}
\usepackage{nomencl}
\makenomenclature

 \renewcommand{\nompreamble}{This list describes several symbols that will be later used within the body of the document}
\begin{document}
\if{0}
\nomenclature{$\chi$}{A dataset} 
\nomenclature{$\chi_i$}{$i^{th}$ instance of $\chi$} 
\nomenclature{$\vec{x}$}{A single sample from set $\chi$}
\nomenclature{$y_i$}{Label of $\vec{x}_i$}
\nomenclature{$D$}{Number of dimensions or features in a dataset}
\nomenclature{$M$}{Reduced rank of dataset} 
\nomenclature{$N$}{Number of instances in dataset}
\nomenclature{$d$}{Dissimilarity or distance between two points}
\nomenclature{$b$}{Cosine dissimilarity}
\nomenclature{$\beta$}{Degree of steepness in exponential transformation of cosine values}
\nomenclature{$g_\beta$}{Exponential transformation function}
\nomenclature{$\vec{w}$}{Prototype}
\nomenclature{$\Lambda $}{Positive semi-definite relevance matrix} 
\nomenclature{$\Omega $}{Projection matrix} 
\nomenclature{$L$}{Discrete set containing closest correct and closest incorrect prototypes}
\nomenclature{$\theta$}{Angle between two vectors}
\nomenclature{$G$}{Origin point of exponential map}
\nomenclature{$\zeta$}{Point on the exponential map}
\nomenclature{$\hat \zeta$}{Point $\zeta$ in the tangent space}
\nomenclature{$\tau$}{Tangent space}
\nomenclature{$\psi$}{Subset containing $\kappa$ nearest neighbours of a selected sample $\tilde x$, and from the same class as $\tilde x$}
\nomenclature{$C$}{Number of defined classes in $\chi$}
\nomenclature{$NaN$}{Not-a-number, missing value}
\nomenclature{$CE$}{Cross-entropy}
\nomenclature{$D_{K||L}$}{Kullbach-Leibler divergence}
\nomenclature{$\vec{y}_i$}{discrete label vector of $\vec{x}_i$}
\nomenclature{$p(c|\vec{x}_i)$}{true probability distribution of $(.)$}
\nomenclature{$\Theta$}{Term in probabilistic ALVQ interpreted as inverse of temperature $\times$ Boltzmann constant}
\nomenclature{$k_B$}{Boltzmann constant}
\printnomenclature
\tableofcontents
\fi
\maketitle
\begin{abstract}
Application of interpretable machine learning techniques on medical datasets facilitate early and fast diagnoses, along with getting deeper insight into the data. Furthermore, the transparency of these models increase trust among application domain experts. Medical datasets face common issues such as heterogeneous measurements, imbalanced classes with limited sample size, and missing data, which hinder the straightforward application of machine learning techniques. In this paper we present a family of prototype-based (PB) interpretable models which are capable of handling these issues. The models introduced in this contribution show comparable or superior performance to alternative techniques applicable in such situations. However, unlike ensemble based models, which have to compromise on easy interpretation, the PB models here do not. Moreover we propose a strategy of harnessing the power of ensembles while maintaining the intrinsic interpretability of the PB models, by averaging the model parameter manifolds. All the models were evaluated on a synthetic (publicly available dataset) in addition to detailed analyses of two real-world medical datasets (one publicly available). Results indicated that the models and strategies we introduced addressed the challenges of real-world medical data, while remaining computationally inexpensive and transparent, as well as similar or superior in performance compared to their alternatives.
\end{abstract}

\section{Introduction}
With the emergence of better and affordable sensors and other data collection tools in various domains the availability of data has exploded. Application of machine learning (ML) techniques in these domains accelerate in-depth analysis of the collected data. ML techniques are being increasingly used in healthcare, judiciary, insurance, logistics, and finance, among other anthropocentric sectors. Therefore it is crucial that the decision-making mechanism of the applied machine learning algorithms are understandable and explainable by human experts. \cite{wang2020should, holzinger2017we, XAItowardsMedicine,sghoshPhDthesis2021} raise an interesting observation that we tend to hold AI to a harsher explanatory standard than we do for drugs and clinicians. The motivation behind this is that clinicians sometimes cannot explain the reason for arriving at a particular diagnosis: a decision may appear intuitive to them but might not actually be explainable. 
Similarly, certain effective drugs had been used widely even before their working mechanism was understood \citep{wang2020should}. Nevertheless, when the decision-making process of a human-expert or a ML algorithm is understandable to the stakeholders, then their trust in the decision increases \citep{XAItowardsMedicine}. Additionally, this ensures improved fairness and prevent biased learning \citep{ghosh2020visualisation, arrieta2020explainable}. Currently, most  ML researchers prioritise results and performance. This tendency, however, comes typically at the cost of reducing transparency. It obscures the inner workings of the ML models, thus precluding any way of verifying the fairness of the system \citep{backhaus2014classification, bibal2016interpretability}. Nevertheless, science and society require much more than just performance metrics of an ML model to be able to adopt it for large scale implementations in the real world, with accountability, fairness, and transparency at the core  \citep{arrieta2020explainable,luo2019balancing,XAItowardsMedicine}. 
This has consequently ushered in the era of Explainable ML or Explainable Artificial Intelligence (XAI) \citep{carvalho2019machine}. 

There are certain terminologies associated with XAI which are used interchangeably in some publications \citep{XAItowardsMedicine}, while others have explained their subtle differences \citep{arrieta2020explainable}. Two such terms are interpretability and explainability. Interpretability, unlike explainability, is not necessarily an active characteristic of a model \cite{arrieta2020explainable, luo2019balancing}. There are three types of interpretability: 
\begin{enumerate}[label=(\Roman*), noitemsep,topsep=0pt] 
\item 
pre-model interpretability, which involves data exploration techniques such as PCA, k-means clustering \citep{parsons2005introduction} and 
t-ditributed Stochastic Neighbor Embedding (tSNE) \citep{maaten2008visualizing};
\item post-model interpretability, in which model-agnostic techniques are applied on black-box models to analyze them locally, 
such as Local interpretable model-agnostic explanations (LIME) \citep{arrieta2020explainable}, DeepView \citep{schulz2020deepview}, Feature Relevance Information (FRI) \citep{pfannschmidt2019fri} and SHapley Additive exPLanations (SHAP)\citep{lundberg2017unified} just to name a few;
\item intrinsically interpretable models are those which are explainable by themselves, such as decision trees (DTs), linear/logistic regression, K-nearest neighbours (KNNs) \citep{arrieta2020explainable}, and nearest prototype based classifiers (NPCs) \cite{ghosh2020visualisation}. 
\end{enumerate}
While neither of the first two techniques give access to the working-logic of the model, the third type includes transparent models \citep{arrieta2020explainable}. However, how the intrinsic interpretability of different classifiers can be compared have often been debated, especially when comparing models of distinct types. \\
Backhaus and Seiffert proposed 3 criteria to answer the aforementioned question \citep{backhaus2014classification, bibal2016interpretability, ghosh2020visualisation}: 
\begin{enumerate}[label=(\arabic*),noitemsep,topsep=0pt]
\item the model's intrinsic ability to select features from the input pattern, 
\item the ability to provide class-specific representative data points, and 
\item model parameters which have information about the decision boundary directly encoded.
\end{enumerate}

\citep{XAItowardsMedicine} describes the criteria (1) as saliency. It further classifies interpretability on the basis of its mathematical structure or whether it is perceptive visually, or verbally, and so on. Unlike \citep{arrieta2020explainable} this paper also uses the terms explainability and interpretability interchangeably. To the best of our knowledge there has not yet been an agreement on the \emph{correct usage} of the aforementioned terminologies in this field, and we do not intend to establish any agreement on this subject in this study. In this paper we present three newly developed prototype-based classifiers which are \textit{competitive} not just in terms of performance, but are also easily and intuitively interpretable. 
Furthermore, they can be visualised, thus allowing intrinsic interpretation and explainability in terms of Backhaus and Seiffert criteria 1-3. These classifiers use a Nearest Prototype Classification (NPC) scheme, where a new sample is assigned to the class of its closest prototype. Techniques implementing this concept, such as Generalized Learning Vector Quantization (GLVQ) \citep{sato} often allow interpretation of the prototypes as representatives of class information, which ensures transparency with regards to (2). Generalized Relevance LVQ (GRLVQ) \citep{Hammer20021059} is an extension of GLVQ which additionally provides feature relevance (criterion 1) by introduction of an adaptive parameterized dissimilarity, which weights features according to their importance for the classification. Further extensions like Generalized Matrix Relevance LVQ (GMLVQ) developed in \cite{Schneider2007,schneider2009} make multi-variate and class-wise feature analysis possible.  The limited rank  version of GMLVQ (LiRaM-LVQ) introduced in \cite{Bunte2011_LiRaMLVQ} facilitates the visualisation of decision boundaries (criterion 3). In addition to verifying model fairness, medical experts are increasingly interested in detailed information about \emph{how} a classification is obtained. 

Certain types of real-life datasets pose the challenges of (a) heterogeneous measurements, (b) missing data, and (c) imbalanced classes, which hinder the straight-forward application of the existing XAI techniques. Heterogeneous measurements arise when data of varying range and types is obtained from different sources. Ignoring that a dataset contains heterogeneous measurements may lead to poorly trained classifier models \citep{tan2016introduction}. Missing data is prevalent in 
control based applications such as traffic monitoring, telecommunications management, financial/business applications, and  biological and medical data analysis \citep{Garcia-Laencina2010}. Missingness can arise due to a variety of reasons. Some causes are \textit{arbitrary}, such as an entry forgotten by medical personnel or a subject choosing to drop out of a study mid-way. Some causes are more \textit{systematic} such as a sensor being unable to measure values beyond a certain range or due to disruption in communication between data collectors \cite{Garcia-Laencina2010}. Little and Rubin have categorised missingness into three types \citep{Garcia-Laencina2010, little2019statistical}: (i) missing completely at random (MCAR), (ii) missing at random (MAR), and (iii) missing not at random (MNAR). These will be discussed in detail in subsections \ref{ss: NaN types} and \ref{sec: NaN Handling}. Lastly, when a dataset being investigated contains significantly unequal numbers of samples per class, it is said to represent an imbalanced class problem. This is common in real-world data sets from sectors such as astronomy, e.g.\ for finding particular types of galaxies \citep{mohammadi2019globular}; telecommunications management, for detection of fraudulent calls; geo-spatial image analysis, for rubble and oil-spills detection \cite{chawla2002smote}; and medicine. If not addressed, class imbalance can cause complications such as a biased and poorly developed classifier. There are classifier formulations that can naturally handle class imbalance, such as Bayesian classifiers employing class priors \citep{mujalli2016bayes}. Others need model-agnostic strategies such as oversampling, undersampling, or boosting exemplified by Synthetic Minority Oversampling TEchnique (SMOTE) presented in \cite{chawla2002smote}.

In \cite{ghosh2017comparison} the authors introduced Angle General Relevance Learning Vector Quantization (AGRLVQ), which is capable of learning from partially observed spaces, thus enabling learning from relatively small data sets containing missing values. This contribution also suggested strategies to deal with imbalanced classes and introduced a geodesic variant of SMOTE \citep{chawla2002smote}. \cite{ghosh2020visualisation} introduced angle-dissimilarity based variant of Generalized Matrix LVQ (GMLVQ) and Local GMLVQ, which in addition to being able to learn from variable dimensional spaces can also tackle more complex problems, while maintaining a superior performance, extracting enhanced knowledge about the dataset they were trained on, and providing visualisation of the classification. The newly introduced angle-dissimilarity based LVQ variants do not require imputation, which saves on time complexity and computational costs for high dimensional datasets while retaining the original information. The intrinsic interpretability of these classifiers lead to intuitive visualisations and knowledge gain. In this contribution we first introduce a probabilistic variant of ALVQ, which in addition to visualization and feature relevance determination, also provides the confidence of the classifier's decision when assigning different class labels to a new sample. Furthermore, we introduce a geodesic average model which can exploit the power of ensembling without compromising on the model interpretability.

In the following sections we discuss common problems associated with biomedical datasets and existing reference standard ML techniques (such as Random Forest (RF), KNN, and LDA), which can partially handle some of the issues. This is followed by the motivations for the newly developed classifiers and the new contributions themselves. We compared our proposed methods to the state-of-the-art shallow ML techniques and RF, on a synthetic dataset and two real-life medical datasets. 
Finally, we present our findings and discuss the extracted knowledge from the real-world medical datasets. 
\section{Challenges and methods for the analysis of biomedical data}\label{s: probs}
As mentioned earlier, many data domains frequently pose challenges such as heterogeneous measurements, missing data, and imbalanced classes, due to limitations in sensor equipment, collection methods or because of high variation in the occurrences of the analysed phenomenon itself. In the medical domain these problems often appear in combination. 
The healthy normal range itself varies due to the physiological features of subjects (such as age or sex or BMI), and the data collection techniques, thereby adding to the heterogeneity of the data. 
If such data is not scaled properly, an unimportant feature which has a higher range of values might be considered more important than it actually is, thereby contributing noisy dimensions. Meanwhile, a feature which is actually important but has values in a lower range might be ignored by the classifier, leading to loss of information. Next, we will discuss in detail the problem of missing values, and imbalanced classes, followed by strategies proposed to handle them.

\subsection{Missing data}\label{ss: NaN types}
As outlined in \citep{little1988test,little2019statistical} there are broadly three categories of missingness. Rubin, in 1976 \citep{Garcia-Laencina2010}, defined the missingness to be of type \emph{missing completely at random (MCAR)} if $f(r|\chi^\textrm{obs}, \chi^\textrm{miss},\Upsilon)= f(r|\Upsilon)$ for all $\chi$ observed ($\textrm{obs}$) and missing ($\textrm{miss}$), where $r$ is the missingness indicator variable, $f$ is probability or density function, 
and $\Upsilon$ is any unknown parameter which caused the missingness. It indicates that the missingness is neither dependent on the observed nor on the missing values of the dataset $\chi \in \R^{N \times D}$ \citep{little1988test, little2019statistical}. 
A common example of MCAR would be a blood vial of a subject from a study that is accidentally broken resulting in blood parameters being not measurable \citep{Garcia-Laencina2010}. On the other hand, Rubin defined missingness to be of type \emph{missing at random (MAR)} if the missingness is independent of the missing values but likely to be dependent on the observed values, i.e., when $f(r|\chi^\textrm{obs}, \chi^\textrm{miss},\Upsilon) = f(r|\chi^\textrm{obs}, \Upsilon)$ \citep{little1988test, little2019statistical}. An example of such missingness is a sensor occasionally failing to acquire data due to power outage. In this scenario the actual variables where data are missing are the cause of some other external influence, such as availability of power, which are recorded \citep{Garcia-Laencina2010}. The third category of missingness, known as \emph{missing not at random (MNAR)} is dependent on the missing values themselves. The cause for this can be systematic, such as the instrument failing to record a parameter when its values are lower than or higher than a certain limit with such data being defined as censored \citep{Garcia-Laencina2010}. Another example of MNAR might be a dataset compounded from different studies or labs, which were not measuring the same parameters.

\cite{Garcia-Laencina2010} broadly defines four strategies to handle missing data:
\begin{enumerate}[label=(\arabic*),noitemsep,topsep=0pt]
\item 
deletion of incomplete cases and performing classification on complete samples only, 
\item 
imputation of missing values using observed data, 
\item 
generative modelling of the data distribution, 
\item 
using ML techniques capable of classifying an incomplete dataset.
\end{enumerate}
Besides, without a doubt, being the most straightforward and simplest strategy, (1) potentially loses a lot of information, 
especially when many instances with partially observed features exist. Furthermore, we often do not have an abundance of data for analysis in many domains such as Medicine, where any loss of information is undesirable, and hence we will not discuss it in this contribution. Strategies to handle missing values of type MNAR are a difficult endeavour, generally requiring knowledge about the process causing the missingness and modeling it accordingly \citep{van2018flexible}. However for the medical datasets which we have come across this is not possible as the mechanisms of missingness are unknown.

\subsubsection{Imputation}\label{sec: NaN Handling}
In most classification tasks of data with missing values it is assumed that missingness is of type MCAR or MAR. Following this assumption the missing values are imputed during pre-processing of data, with strategies broadly divided into (1) single and (2) multiple imputation. These approaches are model agnostic, meaning that afterwards any standard classifier can be applied. 
Imputation generally is a quite common strategy for missing data of type MAR and MCAR \citep{chechik2008max}. 

\paragraph{Single imputation} denotes strategies to fill missing attributes, for example with mean or median of all or a subset of instances that do not miss that feature, such as the k-nearest neighbours (KNN). When the missing variables of interest are correlated with the observed variables from complete samples, regression is the appropriate imputation technique, since it preserves the variance and covariance of the features with missing data. However, for the same reason it fails when imputing missing values in an independent feature, since the imputed value will be correlated and thus changing the original characteristics of the data. Additionally the variance in the dataset is lost when applying this imputation technique \citep{Garcia-Laencina2010}. Two other categories of single imputation are hot and cold deck imputation. 
In hot deck imputation the missing components of a data vector are replaced by the corresponding values found in the complete data vector which is closest to the former data vector (whose missing values are being imputed). 
The disadvantage of this technique is that global properties of the dataset are ignored, since this imputation is based on only the single complete closest data vector. In cold deck imputation the data source to obtain values and the dataset to be imputed are separate datasets \citep{Garcia-Laencina2010}. Single imputation is often adopted due to its simplicity and low complexity. 
However, in contrast to multiple imputation, it provides \emph{one exact value} and can therefore not reflect the uncertainty of the prediction of the missing value \citep{arnab2017survey}.

\paragraph{Multiple imputation (MI)} is used to impute the missing values in the dataset with a set of different \textit{likely} values. A very well respected strategy is a regression-based technique called Multivariate Imputation by Chained Equations 
(MICE)\footnote{The MICE package is publicly available in R \citep{royston2011multiple}} \citep{royston2011multiple}.
It essentially uses a type of hot-deck imputation performed multiple times. Among the available matching techniques for the hot-deck part, predictive mean matching (PMM) proposed by \cite{little1988missing} is often recommended and works as follows: Let $\vec{\chi}^\mathrm{obs}$ denote the $n_1$ observed and $\vec{\chi}^\mathrm{miss}$ the $n_0$ missing entries within one incomplete target variable $\vec{\chi}$. Correspondingly, assume for simplicity $\Xi^\mathrm{obs}$ and $\Xi^\mathrm{miss}$ to be the fully observed $n_1\times q$ and $n_0\times q$ matrix of predictors for the observed and missing data in $\vec{\chi}$, respectively. The first step of PMM bases on Bayesian imputation under the normal linear model, namely it computes the least squares estimate regression weights $\widehat{\varphi}$ from the observed data and draws sample values $\dot{\varphi}$ from the posterior distribution using the standard non-informative priors for each of the parameters. Instead of imputing the linear regression result directly the weights are used to define a matching metric to find a small set of candidate donors, typically 3, 5 or 10, for hot-deck imputation of each missing entry $j=1,\dots,n_0$. Among several possible metrics usually Type 1 is chosen, such that $k$ closest observed candidates are chosen according to the similarity of the estimated value of the observed target entries and the missing value estimate based on the draw from the posterior: $|\Xi^\mathrm{obs}\widehat{\varphi}_{[i]}-\Xi^\mathrm{miss}_{[j]}\dot{\varphi}|$. From the candidate pool of each missing entry $\chi_j^\mathrm{miss}$ a donor is chosen randomly and its value used to impute. Using a posterior sample in the metric considers the sampling variability and the stochastic element also induces between-imputation variation to avoid selecting the same donors too often, which is useful for multiple imputation. Once the incomplete variable $\chi$ is imputed 
the procedure is repeated for the next variable with missing values and so forth. This process is repeated for a user-defined number of times to form multiple imputed sets (in the MICE implementation in R the default is 5 times). Details and extension for multiple regression can be found in \citep{royston2011multiple, van2018flexible}. PMM is often used for two 
main reasons: (1) to prevent imputation by unrealistic values potentially outside the range of available observations, and 
(2) to obviate the need for an explicit model to capture the cause of missingness. 
In practice MI creates several imputed datasets and the same classifier is applied on each of them. The final decision is then made from this ensemble of predictors trained on the different possible \textit{completed} datasets. 

\subsubsection{Machine Learning on incomplete data} \label{sec: LDA LVQ}
Imputation is usually model agnostic and after an incomplete dataset has been imputed, any classifier, such as k-nearest neighbor (KNN), Random Forest (RF), Support Vector Machines (SVM) and so on, can be applied on each of the imputed sets.

\paragraph{Random Forest}\label{ss: RF} (RF) introduced in \citep{breiman2001random} is an ensemble of decision trees (DTs) using bootstrap aggregation (Bagging). A decision tree is a rule based model which can be used for both classification and regression \citep{kubat2017introduction} and due to its transparency it is often used by the medical community. Even though an unpruned decision tree could have a low error rate on the training set, it is prone to overfitting on the validation set. This effect is mitigated in Random Forest because of Law of Large Numbers \citep{breiman2001random}. According to \citep{breiman2001random} the error rate in RF depend on two criteria: (a) the correlation between any two trees, and (b) the strength of each individual tree,  constituting the forest. When the correlation between the trees is high then even increasing the number of trees would not lead to gain in new knowledge, and thus the error rate of RF will not improve. However, error rate of RF decreases with increasing strength of the constituent individual trees of the forest. In Breiman's RF the decision trees are unpruned and each tree learns from a different subset of instances. For classification the final decision is given by the majority vote over an ensemble of all the decision trees. In the Tree Bagger MATLAB implementation the randomness is generated by the random subset selection, which is $30-45\%$ of the training set given as input to the classifier, along with selection of a random subset of predictors (which by default, is equal to the square-root of the original number of predictors) to be evaluated and used at each parent node. Even though it is a robust classifier, due to ensembling RF loses some of the transparency of the decision trees. It also provides only limited information about the decision boundaries and representative examples of classes. One way of estimating the relevance of certain features for the classification is the permutation importance or mean decrease in accuracy (MDA), in which observations of a variable are randomly permuted and the influence on the performance computed. If a feature is not important the permutation should not increase the error made by the model significantly. Conversely if the permutation causes the error to be high it implies that the feature is important \citep{fisher2019all}. The other strategy for finding feature importance is Gini Importance or mean decrease in impurity (MDI) in which, given a predictor the decrease in impurity is averaged over all the trees. Among its drawbacks, it is biased in the presence of correlated features and favour categorical variables with multiple categories \citep{scornet2020trees}. The Tree Bagger in MATLAB uses the former strategy, i.e., MDA which resolves the aforementioned issues of MDI. Random Forest cannot handle missing data directly in its original formulation. Therefore, one can apply multiple imputation on a dataset with missing values before classification with Random Forest.

However multiple imputation is expensive with regards to time and memory with increasing amounts of missingness. 
Especially in a cross-validation setting this is costly, since it needs to be performed for every training set independently to obtain the parameters for imputing the corresponding test set for fair comparison of the generalization error. To avoid imputation of any kind machine learning techniques, which deal with partially observed data were introduced. Prominent examples of strategies based on generative modeling followed by Linear Discriminant Analysis (LDA), as for example analyzed by \citep{Marlin:2008:MDP:1925592}. These methods show promising results for missing data of ignorable types MCAR and MAR and cannot necessarily be assumed to work well on MNAR. Alternatively, prototype based strategies have recently emerged to deal with datasets containing missing values \citep{NaNLVQ, ghosh2020visualisation}. 

\paragraph{Generative modeling strategies}
\label{ss: PPCA LDA}
are often used for (un)supervised data analysis or as preprocessing for partially observed data. 
When dealing with high dimensional data containing a relatively small number of instances, factor analysis (FA) is often used for structured covariance approximation. FA, which is one of the most common latent variable models, assumes that a set of \textit{latent} or \textit{unobservable} factors $t_j,j=1\dots Q$ are linearly combined to generate $\chi$. FA aims to relate a D-dimensional observed data vector $\vec{\chi}$ to its corresponding $Q$-dimensional vector of latent variables $\vec{t}$ ($Q <D$) \citep{tipping1999mixtures,Marlin:2008:MDP:1925592}. 
Vectors $\vec{\chi}$ and $\vec{t}$ are related by
\begin{align}
 \vec{\chi}=\daleth \vec{t} + \mu +\epsilon
\end{align}
Conventionally $\vec{t} \sim \mathcal{N}(0,I)$ (with Identity matrix $I$) and $\epsilon \sim \mathcal{N}(0,\Psi)$, i.e., both the latent variables and the noise model are Gaussian. The latent variants are also independent of each other by convention and $\Psi$ is a square diagonal matrix. $\daleth$ contains the factor loadings and is of dimension $D \times Q$. Therefore the observed variables $\vec{\chi} \sim \mathcal{N}(\mu, \Sigma)$ where $\Sigma=\daleth\daleth^\top+ \Psi$. The parameters $\daleth$, $\Psi$ and $\mu$ are optimized for a dataset using the expectation maximization (EM) algorithm. This model illustrates the dependencies between the data variables $\vec{\chi}$ through the latent variables $\vec{t}$ \citep{tipping1999mixtures,Marlin:2008:MDP:1925592, severson2017principal}. In other words, when variables in the input space are highly correlated, it can be assumed that they have a common source. Additionally FA has a term to explain what was not explainable by the factors, denoted by $\epsilon_i$. Probabilistic Principal Component Analysis (PPCA) is a special case of FA, where instead of the diagonal matrix $\Psi$ the covariance is simplified to $\sigma^2 I$. Since the covariance matrix is assumed to be spherical, PPCA is rotation-invariant with regards to the observed data \citep{Marlin:2008:MDP:1925592, tipping1999mixtures}. Note, that classical PCA is a special case of probabilistic PCA where the noise limit or covariance $\sigma$ is zero.

For supervised analysis these generative model strategies are followed by classification, for example with Linear Discriminant Analysis (LDA) \citep{Marlin:2008:MDP:1925592}. Even though LDA can classify data containing missing values, when the dataset is high dimensional or has small sample size, it is preferable according to \citep{Marlin:2008:MDP:1925592} to use a structured covariance approximation, such as that given by FA and PPCA. Since our medical dataset is both high dimensional and only has a few samples in certain conditions, we followed the suggestion in \citep{Marlin:2008:MDP:1925592}. Hence we use LDA on the Q-dimensional dataset ($\vec{t}$), which in addition to being of lower dimension does not contain missingness. We use PPCA instead of classical PCA because the former is a generative probabilistic model, which makes it amendable to missing data \cite{tipping1999mixtures, severson2017principal}. Further interesting information comparing using PPCA and MICE for learning from data containing missing values can be found in \cite{HEGDE2019100275}.

\paragraph{Prototype-based machine learning methods}
\label{ss:GMLVQ} can intuitively deal with missing data by adapting prototypes and comparing to new data samples based on the observed dimensions only. A powerful family of prototype based classifiers is based on the concept of Learning Vector Quantization (LVQ), which follows a Nearest Prototype Classification (NPC) scheme, where a new vector is assigned the class label of the prototype to which it is closest, according to a chosen dissimilarity measure. Assume the data consist of $N$ instances 
$\vec{x}_i\in\R^D$ accompanied by labels $y_i$ denoting one of $C$ classes and let $\vec{w}^j\in\R^D$ denote one of $C$ prototypes with labels $c(\vec{w}^j)$. Now, Generalized LVQ (GLVQ) performs a supervised training procedure aimed at minimizing the following cost function \citep{sato}, which exhibits a large margin principle \citep{Hammer2005}:
\begin{align}
\label{eq:GLVQ}
E = \sum_{i=1}^N f\left(\lambda
_i\right)
\text{, where  } \lambda
_i =\frac{d_i^J-d_i^K}{d_i^J+d_i^K} \enspace. 
\end{align}
Here, the dissimilarity of each data sample $\vec{x}_i$ to its nearest correct prototype with $y_i=c(\vec{w}^J)$ is defined by $d_i^J$ and by $d_i^K$ for the nearest wrong prototype ($y_i\neq c(\vec{w}^K)$). $f$ is a monotonic function and we use the identity ($f(a) = a$) in this contribution. Extensions to GLVQ introduced parameterized dissimilarity measures, such as the quadratic form:
\begin{align}
\label{eq:d_Lambda}
d^L_i = (\vec{x}_i-\vec{w}^L)^\top\Lambda(\vec{x}_i-\vec{w}^L) \quad\text{ with }\sum_i\Lambda_{ii}=1\enspace,
\end{align}
with a positive semi-definite matrix $\Lambda\in\R^{D\times D}$ containing additional parameters for optimization. 
This led to a family of relevance and matrix extensions (GRLVQ and GMLVQ) that provide intrinsic interpretability 
in the form of relevance of the features for classification determined by the diagonal of $\Lambda$ \citep{Hammer20021059,Schneider2007,schneider2009} and discriminant visualization using low-rank decompositions of $\Lambda$ 
\citep{Bunte2011_LiRaMLVQ}.

In \cite{ghosh2017comparison} the authors introduced two variants of Generalized Matrix LVQ (GMLVQ) which can deal with missing values. The first variant called NaN-GMLVQ bases on the intuitive idea that one can update the prototypes $\vec{w}^L$ and matrix $\Lambda$ in the observed dimensions only for each training sample $\vec{x}_i$. Accordingly, a new sample is classified with the label of the closest prototype computing the distance Eq.~\eqref{eq:d_Lambda} without the missing dimensions. This is achieved by applying the Partial Distance strategy (PDS), shown in \citep{Dixon,doquire2012feature,EIROLA2013115,NaNLVQ}, on Eq.\ \eqref{eq:d_Lambda}.
It introduces a weighting factor proportional to the number of mutually observed dimensions that can be used in the distance Eq.\ \eqref{eq:d_Lambda} between the incomplete training sample with observed dimension indices $D_\mathrm{obs}$ and a prototype by:
\begin{align}
\label{eq: norm_PD_Euclid}
\widehat{d}_i^L&= \frac{D}{|D_\mathrm{obs}|} \sum_{m,n\in D_\mathrm{obs}} (x_{i,m}-w_m^L)\Lambda_{m,n}(x_{i,n}-w_n^L) \enspace .
\end{align}
However, PDS ignores the general variability of the data and has a tendency to underestimate distances due to using only locally known components. The effect is generally more severe when comparing vectors that both have missing components and hence restricting only to mutually known dimensions. This is typically avoided with prototype-based techniques, since only the samples are expected to be incomplete. It practically requires a feature being missing for all samples within a class to result in prototypes with missingness (which has more negative implications for the learning than a mismatch in scale). Note, that assuming the prototypes never miss any dimensions, the PDS factor is only dependent on the sample $\vec{x}_i$ and hence the same for any prototype and 
$d_i^J$ and $d_i^K$ in Eq.~\eqref{eq:GLVQ}. Therefore, it effectively cancels in the computation of the costs and derivatives. 
However, a large variation in the number of missing features across different classes can still lead to stronger repulsion of prototypes of classes with more missingness, effectively pushing prototypes away from classes with less missingness. Countering these effects served as motivation for the development of an LVQ method that classifies on the hypersphere, instead of Euclidean space,
based on an angular dissimilarity measure (ALVQ) as detailed in section \ref{ss:ALVQ}. 

\subsection{Imbalanced classes}

In many domains we face the situation that occurrences of instances from different classes vary in frequency and, on top of it, experts are often most interested in samples of the minority class(es). In the medical field for example, while it is promising that there are more healthy subjects than reported patients, this fact generally poses a challenge in training machine learning models. 
The issue of class imbalance is even more pronounced when the investigated conditions are rare diseases. The main difficulty with training in the presence of class imbalance is that many classifiers tend to become biased towards the majority class. This is due to the fact that the minority class is under-represented or possibly even absent during training. Moreover, performance evaluation measures can also be affected, e.g.\ when looking at one overall accuracy. Literature, e.g.\  \cite{parsons2005introduction},  suggests that the most prominent strategies to handle imbalanced data comprise of bagging, boosting, and sampling, 
including undersampling and oversampling. In \cite{ghosh2017comparison} we introduced a geodesic oversampling strategy and a strategy of penalizing certain misclassifications which yielded promising results. These are explained in the following sections.

\subsubsection{Synthetic Minority Oversampling}\label{sec:SMOTEg}
A well known oversampling method is Synthetic Minority Over-sampling Technique (SMOTE) \citep{chawla2002smote}. It increases the sample size of the minority classes by creating randomized artificial new training samples between $k$ nearest neighbours of the same class. More formally: \begin{equation} \vec{x}_\mathrm{new}=\vec{x} + \alpha\cdot(\vec{x}_\psi-\vec{x}),\end{equation} where $\alpha \in ]0,1[$, $\vec{x}_\mathrm{new}$ is a generated synthetic sample, and $\vec{x}_\psi$ is one of the $k$ nearest neighbours of $\vec{x}$. However this simple solution might not be the best choice when the applied classifier operates on a manifold as  in \citep{ghosh2017comparison}. In such a case SMOTE can be performed on that manifold. For example, the authors introduced a geodesic variant of the original SMOTE, which synthesized samples on the hypersphere instead of Euclidean space, since the transformed data points were known to lie on a hypersphere. To achieve this an important tool of Riemannian geometry is used, which is the exponential map \citep{Fletcher2004,WilsonHPD14pami}. The exponential map has an origin $G$, which defines the point for the construction of the tangent space $\tau_G$ of the manifold. Let $\zeta$ be a point on the manifold and $\hat{\zeta}$ the corresponding point in the tangent space with $\hat{\zeta}=\text{Log}_G(\zeta)$, $\zeta=\text{Exp}_G(\hat\zeta)$ and $d_g(\zeta,G)=d_e(\hat \zeta,G)$ with $d_g$ being the geodesic distance between the points on the manifold and $d_e$ being the Euclidean distance on the tangent space. $\text{Log}$ and $\text{Exp}$ denote a mapping of points from the manifold to the tangent space and vice versa. As described in \citep{ghosh2017comparison} we present a point $\vec{x}$ from class $c$ on the unit sphere with fixed length $\lVert\vec{x}\rVert=1$, that becomes the origin of the tangent space. Next, $k$ nearest neighbours of the selected sample $\vec{x}$ are found from the same class $\vec{x}_\psi\in\mathcal{N}_{\vec{x}}$ using the geodesic distance between the vectors $\theta = \cos^{-1}(\sfrac{(\vec{x}^\top\vec{x}_\psi)}{r^2})$ and (in our case) $r=1$. 
\begin{figure}[t]
\centering
\includegraphics{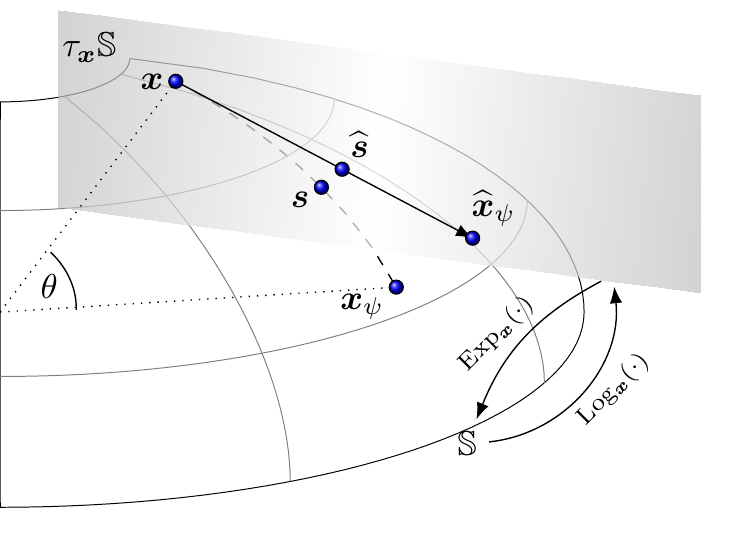}
\caption{Depiction of geodesic SMOTE to generate synthetic samples $\vec{s}$ on the hypershpere to oversample minority classes for imbalanced data using Riemannian geometry.}
\label{fig:geodesSMOTE}
\end{figure}
Each random neighbour $\vec{x}_\psi$  is then projected onto that tangent space using only the available features and the $\text{Log}$ transformation for spherical manifolds:
\begin{align}
\hat{\vec{x}}_\psi = \mathrm{Log}_{\vec{x}}(\vec{x}_\psi) = \frac{\theta}{\sin\theta}(\vec{x}_\psi-\vec{x}\cos \theta) \enspace. 
\end{align}
Finally, a synthetic sample is produced either on the tangent space with a formula similar to the original SMOTE, namely $\hat{\vec{s}}=\alpha\cdot\hat{\vec{x}}_\psi$, that is subsequently projected onto the sphere via Exp transformation: 
$\vec{s} = \mathrm{Exp}_{\vec{x}}(\hat{\vec{s}})$. Or we produce the new samples on the geodesic directly using the new angle $\hat\theta=\lVert{\widehat{\vec{x}}_\psi}\rVert$ and the $\text{Exp}$ transformation: 
\begin{align}
s = \vec{x}\cos(\hat\theta\alpha)+\frac{\sin(\hat\theta\alpha)}{\hat\theta}\cdot\widehat{\vec{x}}_\psi \qquad\text{with }\alpha\in]0,1[\enspace .
\end{align}
This procedure of synthetic sample generation is depicted in figure \ref{fig:geodesSMOTE} and repeated with other random samples from the class until the desired number of training samples is reached. We propose to oversample each of the minority classes in the training set until they are equivalent in size to the majority class. This avoids the original SMOTE hyperparameter selection, namely the percentage of oversampling for each minority class.

\subsubsection{Variable penalty/reward cost weight matrix}\label{sss:PenaltyMat}

Unlike the oversampling strategy which is model agnostic, this strategy to handle imbalanced classes is integrated in the LVQ model training. The LVQ cost function induces an update of the model parameters based on a presented training sample. Therefore, majority classes with significantly more samples can introduce bias in the final model by simply causing more updates to the parameters during training than the minority classes. An intuitive way to circumvent this is by introducing a weighting dependent on the number of samples in the class, effectively reducing the update strength for majority class samples. This principle can be furthermore used to incorporate expert knowledge and preferences in cases where an error free classification cannot be achieved. Some errors might be more costly than others, such as a misclassification of a patient as healthy that would not get treated. A misclassification of a disease for another where the treatment is similar on the other hand might be more acceptable. The model can be incentivised to reduce certain misclassifications by making the error costlier with higher weights. Following the suggestion in \citep{Pazzani1994}, a hypothetical cost matrix $\Gamma=\{\gamma_{cp}\}_{c,p=1,...,C}$ with $\sum_{c,p}\gamma_{cp}=1$ was introduced, so as to boost learning of difficult or minority classes, thus enabling enhanced differentiation between minority classes (all disease classes) and the majority class (healthy class). The rows of this matrix correspond to the actual classes $c$ and columns denote the predicted classes $p$ of the current model parameters. When user-defined costs are unavailable and one simply wants to correct for the class imbalance, equal costs can be assigned to all $\gamma_{cp}$. This ensures that the weight contribution of each class is inversely proportional to the class strength. These costs are included in our cost function Eq.(\ref{eq:cwALVQ}), as shown below:
\begin{equation}
\label{eq:cwALVQ}
\hat{E}=
\sum_{c=1}^C 
\frac{1}{n_c} \left[\sum_{\vec{x}_i, s.t. y_i = c} 
\gamma_{c, {\hat y_i} } 
\lambda_i\right]
\enspace,
\end{equation}
where $c=y_i$ is the class label of training sample $\vec{x}_i$, $n_c$ defines the number of samples within that class, $\hat y_i$ is the predicted label (label of the nearest prototype $c(\vec{w}^J)$), and $\lambda_i$ is the cost function value of sample $i$ Eq.\ \eqref{eq:GLVQ}. To chose the matrix one can run the algorithm with equal entries first and adapt it according to the undesired misclassifications observed. 

\section{Biomedical motivation}
In this section two biomedical datasets, exhibiting typical problems, such as missing data and imbalanced classes, which provided motivation for our research are described: 
(1) a real-world medical dataset containing urinary steroid excretion data measured by Gas Chromatography–Mass Spectrometry (GC-MS) measurements in patients with inborn disorders of steroidogenesis and healthy controls, from the Institute of Metabolism and Systems Research (IMSR), University of Birmingham; and (2) a publicly available real-world heart disease dataset from the UCI repository. 
\subsection{Urine steroid metabolite dataset}
\begin{figure}[t]
\centering
\includegraphics[width=0.85\textwidth, trim={1cm 0.3cm 1cm 0.5cm},clip]{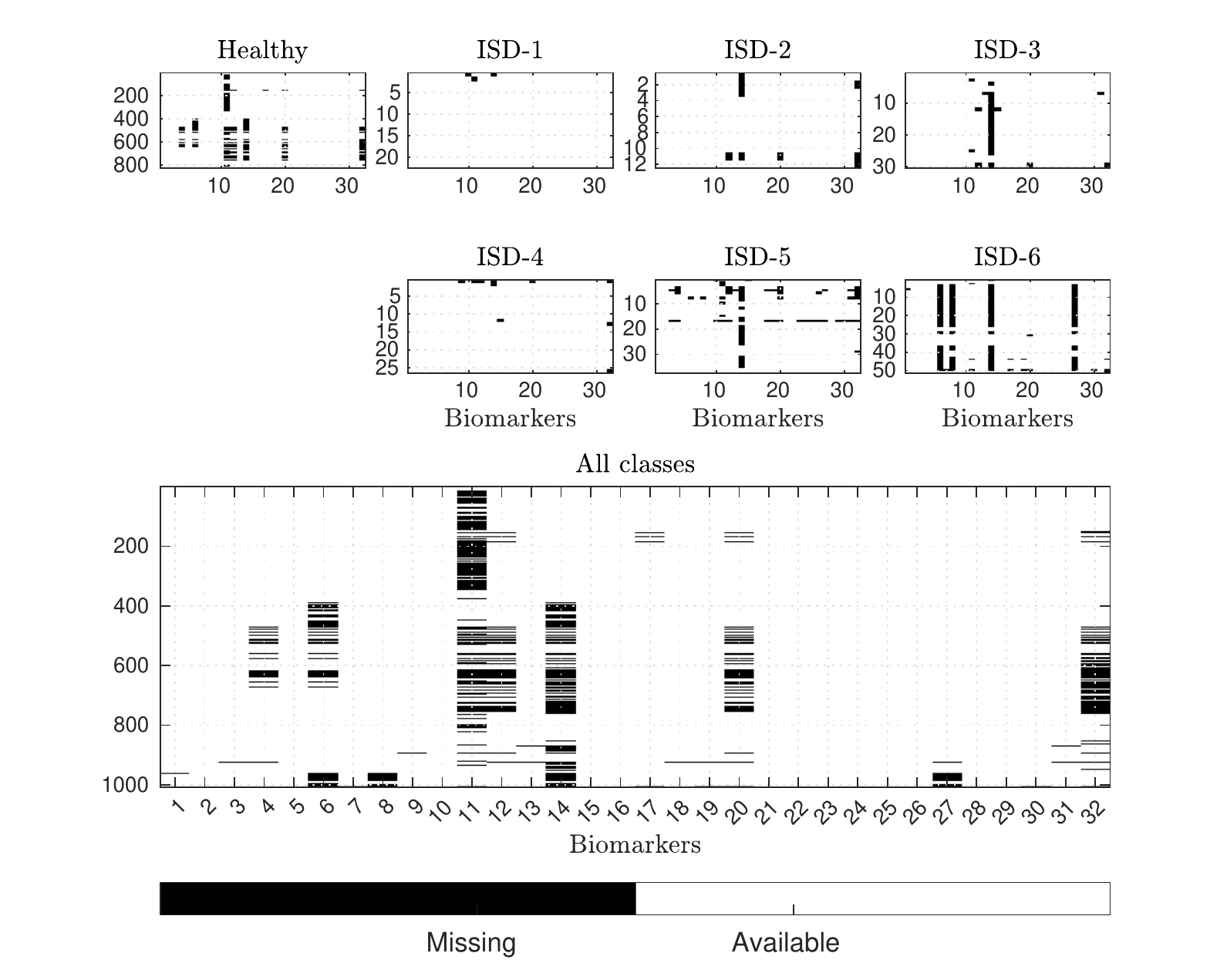}
\caption{Heat-maps showing the presence of systematic missingness in each of the conditions of steroidogenic disorders and healthy subjects contained in the GCMS dataset.
}
\label{fig:sysMissGCMS}
\end{figure}

Inborn disorders of steroidogenesis are genetic diseases which affect the Endocrine system that synthesizes hormones for a variety of bodily functions, such as blood pressure regulation, stress response, sex differentiation and puberty. Mutations in genes encoding distinct enzymes can cause blockages in hormone production leading to several forms of Congenital Adrenal Hyperplasia (CAH) and Differences in Sex Development (DSD) \citep{baranowski2018monogenic}. Early detection is essential, since some of these rare conditions can be life-threatening. Rapid diagnosis would allow life-saving treatment to be delivered in  a more efficient manner, thereby reducing the distressing time of diagnostic uncertainty for patients and their families. Furthermore, it would also enable doctors to plan and advice future treatment strategies more promptly. Accurate biochemical diagnosis can be made by measuring characteristic patterns of individual steroid metabolites altered in these enzyme deficiencies, however, the complexity of this data means computer aided approaches for diagnosis are highly desirable. The IMSR at the University of Birmingham, UK, collected a unique and extensive dataset of urinary steroid metabolite excretion data in patients with inborn steroidogenic disorders, which were collected over a period of two decades. As often seen for the analysis of rare diseases, the data exhibits several common difficulties for straightforward approaches for computer-aided diagnosis. For example, in some of the samples in the dataset certain steroid metabolites were not measured as at the time of analysis these steroids were not yet part of the assay used for steroid multi-profiling. Since the data was collected over a long period of time the clinicians' understanding of which are important metabolites have improved, as has the GCMS method itself. Together, these issues gave rise to systematic missingness in this dataset. In this database, 32 steroid metabolite concentrations, referred henceforth as biomarkers, have been measured using GCMS. The dataset contains measurements from 829 healthy controls and 178 patients with inborn disorders of steroidogenesis (ISD-1: 22, ISD-2: 12, ISD-3: 30, ISD-4: 26; ISD-5: 37; ISD-6: 51). The number of subjects in each class clearly shows the presence of high levels of class imbalance. The class imbalance in this dataset arises from the opportunistic nature of how these samples were collected, rather than the imbalance being representative of the population prevalence of these diseases. 
The third challenge is the presence of very heterogeneous measurements. 
Large variations in biomarker profiles are observed across subjects even within the same condition class due to individual physiological features, such as age, sex, etc. 
There is also heterogeneity in sample collection method, including single urine sample collections, urine extracted from nappies for babies, and full 24-hour urine collections. 
It has been proposed that using ratios of metabolites reduces some of this heterogeneity, allowing direct comparison of results obtained from different 
urine collection methods \citep{arlt2004congenital, storbeck2019steroid, baranowski2018monogenic}. 
Hence, we also used this approach, but from a completely data-driven perspective, and 496 potentially informative ratios were built by pair-wise combinations of the 32 biomarkers. 
The heatmaps in figure \ref{fig:sysMissGCMS} illustrate the missingness in each condition of the GCMS dataset. 
 
\subsection{Cleveland heart disease dataset from UCI repository}
This dataset contains 13 features from 164 healthy subjects and 139 subjects with varying degrees of heart problems. 
The predictor variable is originally 5 unique values, 0 indicating healthy (164), while 1 (55 subjects), 2 (36 subjects), 3 (35 subjects), and 4 (13 subjects) indicating patients with different heart conditions. Furthermore, six subjects contain missing values. According to \cite{DavidWAha1988} the missing values in the data were replaced by a value of $-9$. Exploratory analysis showed that while there is a very good separation between healthy and HD subjects considered in binary classification, the multi-class problem differentiating between the 4 classes of HD patients turns out to be remarkably difficult. In this study we investigated this dataset for the five class problem, as suggested in \cite{ghosh2020visualisation, sghoshPhDthesis2021}. 
The dataset originally consisted of 76 features but most research has been done on the publicly available subset of 13 of these. 
Further details about them can be found at the UCI repository \cite{DavidWAha1988}. 

In classification problems addressing any type of missingness is challenging, because for most mainstream classifiers managing missing values is not straightforward \citep{Marlin:2008:MDP:1925592}. As seen in Figure \ref{fig:sysMissGCMS} the urine GCMS dataset contains both random and systematic missingness. For the GCMS dataset the systematic missingness arose from different studies measuring different metabolites and the time when the measurement was made. Information about the cause of missingness in the heart disease dataset is unavailable to us. As mentioned in \ref{ss:GMLVQ} the presence of missingness, especially systematic missingness, cannot be straightforwardly handled by existing intrinsically interpretable classifiers to the best of our knowledge, and imputation is likely to induce bias in the data. The combination of complications arising in biomedical problems such as these, motivated the development of a novel family of geodesic prototype-based classification strategies as outlined in the following.

\section{Geodesic prototype-based classification}

In \cite{ghosh2020visualisation} the authors introduced a prototype-based classification method using a parameterized angular dissimilarity classifying on the hypersphere. The Angle Learning Vector Quantization (Angle LVQ) strategy (denoted henceforth as $LVQ^{A}$) shows promising results facing systematically missing values and very heterogeneous data 
where the absolute values are not informative, while enabling intrinsic interpretability by biomarker detection and visualization of the decision boundaries. In this contribution we systematically investigate the influence of missing values of types MCAR and MNAR, the amount of missingness and the training set size to compare the classification performance of several common strategies to deal with such problems. Furthermore, we extend the $LVQ^A$ algorithm to a geodesic prototype-based classification framework\footnote{Matlab code is made publicly available at \url{https://github.com/kbunte/geodesicLVQ_toolbox}} including:
(1) A probabilistic variant 
which provides better interpretability to the user in terms of confidence of the classifier's decision; 
(2) a rank-preserving average of matrix LVQ models, formulated using the geodesic on the Riemannian manifold the parameters lay on; and
(3) a strategy to cluster LVQ models based on the geodesic distance of their metric tensors to identify and interpret local optima.
Interestingly, the rank-preserving mean often shows 
a more robust performance than that of a single classifier, however, unlike an ensemble approach, it retains the 
interpretability and transparency of an individual LVQ model.

\subsection{Angle LVQ} 
\label{ss:ALVQ}

Angle GRLVQ and angle GMLVQ \citep{ghosh2020visualisation} were developed as the angle-based variants of their Euclidean counterparts optimizing the same cost function as the GRLVQ and GMLVQ, namely Eq.\ \eqref{eq:GLVQ}. The angle based variants (referred to as $LVQ^{A}$ henceforth) replace the quadratic form $d_i^{\{J,K\}}$ in Eq.\ \eqref{eq:d_Lambda} by a parameterized angle-based dissimilarity:

\begin{align} \label{eq:ALVQ_diss}
d_i^L &= g_\beta\left(b\right) \text{ with } & 
\notag
b = b_\Lambda(\vec{x_i},\vec{w}^L)= \frac{\vec{x}_i^\top\Lambda\vec{w}^L}{{\lVert\vec{x}_i\rVert}_\Lambda\lVert{\vec{w}^L\rVert}_\Lambda} \\ 
\text{ where }\\
\notag
\lVert\vec{v}\rVert_{\Lambda} & =\sqrt{\vec{v}^\top\Omega^\top\Omega \vec{v}} \enspace, &   \Lambda =\Omega^\top\Omega \enspace, \\ 
\notag
g_\beta(b) &= \frac{e^{-\beta (b-1)}-1}{e^{(2\beta)}-1},  &\sum_i\Lambda_{ii}=1 \text{ and } L\in\{J,K\}\enspace .
\end{align}
Here, the exponential function $g_\beta(b)$ transforms the cosine $b=\cos\theta\in[-1,1]$ into dissimilarities in the range [0,1]. The dissimilarity measure $d_i^L$ itself can be parameterized enabling several powerful extensions with varying potential for further interpretation \citep{ghosh2020visualisation, sghoshPhDthesis2021}. 
This includes the number of prototypes used to represent each class (which is fixed to one throughout this contribution) and the choice of the metric tensor. The simplest choice for the metric tensor is restricting $\Lambda$ to a diagonal matrix with $\Lambda_{ij}=0$ $\forall i\neq j$ and $\Lambda_{ij}>0$ $\forall i=j$ to learn the relevance of each feature for the classification. More complex is the use of a global metric tensor trained by decomposing $\Lambda=\Omega^\top\Omega$ with $\Omega \in \R^{M\times D}$ for $M\le D$ to ensure positive semi-definiteness of $\Lambda$.
Strictly speaking, if $M\neq D$ we work with a pseudo-Riemannian, also called a semi-Riemannian manifold \citep{Amari}. 
For simplicity we still refer to the general positive semi-definite $\Lambda$ as ``metric", abusing the mathematical terminology slightly. In addition to the weighting of the individual dimensions $\Lambda_{ii}$ this enables rotating the coordinate system towards discriminant directions for classification 
\citep{biehl2013distance} and the linear transformation $\Omega$ allows for visualization of the decision boundaries if $M\in\{2,3\}$ similar to \citep{ghosh2020visualisation}.\par

The cost function Eq.\ \eqref{eq:GLVQ} is non-convex and can for example be optimized using stochastic gradient descent or conjugate gradient methods with the following derivatives 
for the parameters $\Phi\in\{\{\vec{w}^j\}_{j=1}^k,\Omega\}$: 
\begin{align}
\frac{\partial E}{\partial \Phi} =& \sum_{i=1}^N\frac{\partial f}{\partial \Phi}\frac{\partial \mu_i}{\partial \Phi} \qquad \text{with} \\
\frac{\partial\mu_i}{\partial \Omega} =& \frac{2 d^K}{(d^J + d^K)^2} \cdot \frac{\partial d^J}{\partial \Omega} - \frac{2 d^J}{(d^J + d^K)^2}\cdot \frac{\partial d^K}{\partial \Omega} \\ 
\frac{\partial \mu_i}{\partial \vec{w}^J} =& \frac{2 d^K}{(d^J + d^K)^2}\cdot \frac{\partial d^J}{\partial \vec{w}^J} \qquad \text{and}\qquad 
\frac{\partial \mu_i}{\partial \vec{w}^K} = \frac{-2 d^J}{(d^J + d^K)^2}\cdot \frac{\partial d^K}{\partial \vec{w}^K}\\
\frac{\partial d_i^L}{\partial \Phi} =& \frac{\partial g_\beta(b_\Omega)}{\partial b_\Omega}\cdot \frac{\partial b_\Omega}{\partial \Phi} \\ 
\label{eqn:deriv bw}
\frac{\partial b_\Omega(\vec{x}_i,\vec{w}^L)}{\partial \vec{w}^L} = &
\frac{\vec{x}_i\Omega^\top\Omega\lVert\vec{w}^L\rVert_\Omega^2-\vec{x}_i\Omega^\top\Omega \vec{w}^L\cdot \vec{w}^L\Omega^\top\Omega}{\lVert {\vec{x}_i\rVert}_\Omega\lVert {\vec{w}^L\rVert}_\Omega^3}\\ 
\notag\frac{\partial b_\Omega(\vec{x_i},\vec{w}^L)}{\partial \Omega_{mn}} =  
&\frac{x_{i,n}\sum_j^D\Omega_{mj} {w^L_j}+
{w^L_{n}} 
\sum_j^D\Omega_{mj}x_{i,j}}{\lVert {\vec{x}_i\rVert}_\Omega \lVert {\vec{w}^L\rVert}_\Omega} \\
&-\vec{x}_i\Omega^\top\Omega \vec{w}^L \left[
\frac{x_{i,n}\sum_j\Omega_{mj}x_{i,j}}{\lVert{\vec{x}_i\rVert}^3_\Omega\lVert{\vec{w}^L\rVert}_\Omega} + 
\frac{{w^L_n} \sum_j \Omega_{mj} w^L_j}{\lVert\vec{x}_i\rVert_\Omega\lVert \vec{w}^L\rVert^3_\Omega}\right] \enspace, \label{eqn:deriv bo}
\end{align}

where $x_{i,n}$ denotes dimension $n$ of vector $\vec{x}_i$ and $m=1,\dots,M$.
In the presence of missing data the cosine dissimilarity $b$ and its derivatives are computed with the available dimensions only. This aspect is similar to the Euclidean version, referred to as NaNLVQ, which was presented in \cite{ghosh2017comparison, sghoshPhDthesis2021}. However, in contrast to NaNLVQ which uses the normalization strategy in Eq.\ \eqref{eq: norm_PD_Euclid}, this parameterized angle measure contains a normalization that corrects the comparison of vectors of different length more robustly especially for increasing missingness. The generalization bounds can be estimated using the Rademacher complexity similar to LGMLVQ \citep{schneider2009}.

In \cite{ghosh2020visualisation} we also introduced the angle variant of the localized GMLVQ (LGMLVQ), denoted hereon by $LVQ^{LE}$ \cite{schneider2009} by attaching metric tensors $\Lambda^c$ to each prototype or each class. 
The diagonal of the local metric tensors $\Lambda^j=\Omega^{j\top}\Omega^j$ contain local or class-wise feature relevances, 
which enables more complex modeling in addition to providing class-specific discriminative information. 
The local $LVQ^{A}$ extension (denoted by $LVQ^{LA}$) is therefore written as:
\begin{align}
\label{eq:bOmegaL}
b=b_{\Omega^L} =& \frac{\vec{x}_i^\top\Omega^{L\top}\Omega^L\vec{w}^L}
{\lVert\vec{x}_i\rVert_{\Omega^L}\lVert\vec{w}^L\rVert_{\Omega^L}} \enspace,
\intertext{with corresponding derivatives of $b_{\Omega^L}$: 
}
\frac{\partial b_{\Omega^L}}{\partial \vec{w}^L} =& 
\frac{\vec{x}_i \Omega^{L \top} \Omega^L \lVert \vec{w}^L \rVert_{\Omega^L}^2-\vec{x}_i \Omega^{L\top} \Omega^L \vec{w}^L \cdot \vec{w}^L \Omega^{L\top}\Omega^L}
{\lVert \vec{x}_i\rVert_{\Omega^L}\lVert\vec{w}^L\rVert_{\Omega^L}^3} \\
\frac{\partial b_{\Omega^L}}{\partial \Omega^L_{{{mn}}}}  =& 
\frac{x_{i,n} \sum_j^D \Omega_{mj}^L w_{j}^L +w_n^L \sum_j^D \Omega_{mj}^L x_{i,j}}
{\lVert \vec{x}_i \rVert_{\Omega^L}\lVert \vec{w}^L\rVert_{\Omega^L}} 
- \notag\\
&\vec{x}_i \Omega^{L\top} \Omega^L \vec{w}^L 
\left[\frac{x_{i,n} \sum_j^D \Omega_{mj}^L x_{i,j}}{\lVert \vec{x}_i \rVert^3_{\Omega^L}\lVert \vec{w}^L\rVert_{\Omega^L}} + \frac{w_n^L\sum_j^D \Omega_{mj}^L w_j^L}
{\lVert \vec{x}_i \rVert_{\Omega^L}\lVert \vec{w}^L\rVert^3_{\Omega^L}}\right] \enspace .
\end{align}

The update rules of $LVQ^{A}$, similar to their Euclidean predecessors, contain forces attracting the closest correct prototype towards each data sample, and forces of repulsion pushing away the closest one with a different class label. In an imbalanced class problem the Euclidean variant might push the minority class prototype far away from the data all together, since it is being repelled more often by the majority class than attracted by the minority class. However, the $LVQ^{A}$ variants classify on the surface of the hypersphere. Whereas in Euclidean space repelled prototypes can increase their distance to all prototypes simultaneously, which may lead in some cases to infinite repulsion. This cannot happen in $LVQ^{A}$ since a repelled prototype inevitably gets closer to another prototype due to the nature of the hypershpere, leading to a more stable behaviour when facing imbalance. Furthermore, the hyper-parameter $\beta$ in Eq.\ \eqref{eq:ALVQ_diss} influences the slope of the dissimilarity conversion. Therefore, $\beta\rightarrow 0$ leads to a near linear relationship between the update strength dependent on the distance of the sample to the the corresponding prototype. The $\beta$ in the exponential function influences the strictness of the classifier's decision boundary. The larger the value of $\beta$ the more effectively it reduces the contribution of a sample to the update of a prototype from which is it very far away, and increases the influence of a nearby sample. In other words, the greater the distance between a sample and a prototype, the lesser is the contribution of that sample towards the update strength of the prototype, and the value of $\beta$ determines how much greater or lesser the contribution is based on the distance. In this contribution we use $\beta=1$ unless explicitly stated otherwise, and therefore denote the angle LVQ simply by $LVQ^A$ instead of $LVQ^{A^\beta}$.

\subsection{A probabilistic approach to classifying data with missingness}\label{ss:pALVQ}
In the medical domain, patients can have multiple comorbidities instead of a single crisp condition, they may be on the borderline between two or more conditions, or they can have a diagnosis which shows phenotypic similarity or overlap with other conditions. 
If the classifier could estimate the probability of a patient belonging to condition-1 and the probability of belonging to condition-2 then this would constitute useful information, for instance for the planning of further, often more expensive, confirmatory investigations or for treatment planning. 
Moreover, some diseases may be difficult to diagnose, which may result in different labels when several experts are consulted. This can be expressed as probability of a class dependent on the fraction of experts that agree.
Therefore, we develop a probabilistic version of $LVQ^{A}$, which allows to express our uncertainty about the class label, given an input, in the form of conditional probability distribution over the classes. 

Authors of \cite{villmann2018probabilistic} and \cite{schneider2011multivariate} used information theoretical principles to generalize Robust Soft LVQ (RSLVQ), by using maximum likelihood and the Cross-Entropy (CE) as the cost function. In our formulation we estimate the class when the sample $\vec{x}$ is given, by minimizing the difference between the true class and our estimate i.e., by minimizing the Kullback-Leibler (KL) divergence ($D_{KL}$) in the cost function. It is closely related to the CE used in \cite{villmann2018probabilistic}. It is interpreted in information theory as the additional number of bits required to convey encode the data \citep{tse2005appendix}. 
Consider the unknown joint distribution $p(\vec{x},c) = p(c|\vec{x}) p(\vec{x})$ over the inputs and labels that generated our training set $\{ (\vec{x}_i,c_i) \}_{i=1}^N$. 
Our discriminative model produces an estimate $\hat p(c|\vec{x})$ of $p(c|\vec{x})$.
The expected KL divergence measures the mismatch between 
$\hat p(c|\vec{x})$ and 
$p(c|\vec{x})$ can be approximated through the training sample as
\begin{equation}
 \begin{split}
 \label{eq:pALVQ3}
 H(\hat p(c|\cdot)) & = E_{p(\vec{x})}[D_{KL}(\hat{p}(c|\vec{x})\parallel p(c|\vec{x}))] \\
                    & \approx \frac{1}{N}\sum_{i=1}^N \sum_{c=1}^C \hat{p}(c|\vec{x}_i)[\ln \hat{p}(c|\vec{x}_i)-\ln p(c|\vec{x}_i)]
                   =  \frac{1}{N}\sum_{i=1}^N \sum_{c=1}^C \hat{p}(c|\vec{x}_i)\ln \frac{ \hat{p}(c|\vec{x}_i)}{p(c|\vec{x}_i)} \enspace .              
 \end{split}
\end{equation}
Since we do not have access to the true distributions $p(c|\vec{x}_i)$ the cost function is often formulated by considering only the generated labels $c_i$. 

For the case of the generated sample $(\vec{x}_i,c_i)$ being noise-free $p(c_i|\vec{x}_i) = 1$ (for example when the diagnosis is genetically confirmed) other classes have a probability of 0 and KL cannot be used. 
In such cases one can either simplify the cost function considering only $c_i$ for $\vec{x}_i$:
$\frac{1}{N}\sum_{i=1}^N  \hat{p}(c_i|\vec{x}_i)\ln \hat{p}(c_i|\vec{x}_i)$ or introduce some noise by substracting $\epsilon$ from the class and adding $\epsilon/(C-1)$ to the others. In the following we assume the latter and provide the detailed derivatives for noisy labels.

For sample $\vec{x}_i$ the $\hat p(c|\vec{x}_i)$ is computed by the following parameterized softmax function:
\begin{align}
\label{eq:E_pALVQ}
\hat p(c|\vec{x}_i) &= \frac{g_\Theta\left(
\frac{\vec{x}_i \Lambda \vec{w}^{c\top}}{
{\lVert\vec{x}_i\rVert}_\Lambda\lVert{\vec{w}^c\rVert}_\Lambda}\right)
}{\sum_j^C g_\Theta\left(
\frac{\vec{x}_i \Lambda \vec{w}^{j\top}}{
{\lVert\vec{x}_i\rVert}_\Lambda\lVert{\vec{w}^j\rVert}_\Lambda
}\right)} \quad
\text{with }\ g_\Theta(b) = \frac{e^{\Theta (b+1)}-1}{e^{(2\Theta)}-1} \enspace .
\end{align}

The parameter $ \Theta $ can be interpreted as $\frac{1}{k_BT}$ where $k_B$ is the Boltzmann constant and $T$ is the absolute temperature. The derivatives of $D_{KL}(\hat{p}(c|\vec{x})\parallel p(c|\vec{x}))$ (Eq.~\ref{eq:pALVQ3}) with $\lVert\vec{v}\rVert_{\Omega}=\sqrt{\vec{v}^\top\Omega^\top\Omega\vec{v}}$ are: 
\begin{equation}
 \frac{D_{KL}(\hat{p}(c|\vec{x}_i)\parallel p(c|\vec{x}_i))}{\partial \Omega}  = \sum_{c=1}^C  \frac{\partial \hat{p}(c|\vec{x}_i)}{\partial \Omega}\cdot 
 \left(1+ \ln \frac{\hat{p}(c|\vec{x}_i)}{p(c|\vec{x}_i)} \right)
\end{equation}
and
\begin{equation}
 \frac{D_{KL}(\hat{p}(c|\vec{x}_i)\parallel p(c|\vec{x}_i))}{\partial \vec{w}^j}  = \sum_{c=1}^C  \frac{\partial\hat{p}(c|\vec{x}_i)}{\partial \vec{w}^j}\cdot 
 \left(1+ \ln \frac{\hat{p}(c|\vec{x}_i)}{p(c|\vec{x}_i)} \right)
\end{equation}
Now $\frac{\partial \hat{p}(c|\vec{x}_i)}{\partial \Omega}$ can be expanded to
\begin{equation}
 \frac{\partial \hat{p}(c|\vec{x}_i)}{\partial \Omega}=\frac{\frac{\partial g_\Theta(\vec{x}_i,\vec{w}^c)}{\partial \Omega}\cdot 
 \sum_{j=1}^C g_\Theta(b_\Omega(\vec{x}_i,\vec{w}^j))-g_\Theta(b_\Omega(\vec{x}_i,\vec{w}^c))\cdot \sum_{j=1}^C
 \frac{\partial g_\Theta(b_\Omega(\vec{x}_i,\vec{w}^j))}{\partial \Omega}}{(\sum_{j=1}^C g_\Theta(b_\Omega(\vec{x}_i,\vec{w}^j)))^2} \enspace ,
\end{equation}
similarly $\frac{\partial \hat{p}(c|\vec{x}_i)}{\partial \vec{w}^j}$ for $j=c$ we have
\begin{equation}
 \frac{\partial \hat{p}(c|\vec{x}_i)}{\partial \vec{w}^{c}}= \frac{\frac{\partial g_\Theta(\vec{x}_i,\vec{w}^c)}{\partial \vec{w}^{c}}\cdot 
 \sum_{j=1}^C g_\Theta(b_\Omega(\vec{x}_i,\vec{w}^j))-g_\Theta(b_\Omega(\vec{x}_i,\vec{w}^c))\cdot \frac{\partial g_\Theta(b_\Omega(\vec{x}_i,\vec{w}^{c}))}{\partial \vec{w}^{c}}}{\Big(\sum_{j=1}^C g_\Theta(b_\Omega(\vec{x}_i,\vec{w^j}))\Big)^2} \enspace ,
\end{equation}
and the derivative for $j\neq c$ is given by
\begin{equation}
 \frac{\partial \hat{p}(c|\vec{x}_i)}{\partial \vec{w}^{j}}=\frac{-g_\Theta(b_\Omega(\vec{x}_i,\vec{w}^c))\cdot \frac{\partial g_\Theta(b_\Omega(\vec{x}_i,\vec{w}^{j}))}{\partial \vec{w}^{j}}}{\left(\sum_{k=1}^C g_\Theta(b_\Omega(\vec{x}_i,\vec{w}^k))\right)^2}
\end{equation}
where 
\begin{equation}
 \frac{\partial g_\Theta(b_\Omega(\vec{x}_i,\vec{w}^j))}{\partial \Phi}=\frac{\Theta e^{\Theta(b_\Omega+1)}}{e^{2\Theta}-1}\cdot \frac{\partial b_\Omega}{\partial \Phi} \quad \text{  with } \Phi \in [\Omega, \vec{w}^j] \enspace.
\end{equation}
The partial derivatives $\frac{\partial b_{\Omega}}{\partial \vec{w}}$ and $\frac{\partial b_{\Omega}}{\partial \Omega}$ are defined as in Eq.\ \eqref{eqn:deriv bw} and \eqref{eqn:deriv bo}.

This probabilistic variant of $LVQ^{A}$ will henceforth be abbreviated as $PLVQ^{A^\Theta}$ in contrast to the deterministic variant $LVQ^{A^\beta}$). In both subsections \ref{ss:pALVQ} and \ref{ss:ALVQ} the cost weight matrix introduced in \ref{sss:PenaltyMat} could be introduced as an alternative to minority class oversampling, to handle class imbalance in an  efficient manner, avoiding an increase in the number of training samples.

\subsubsection{Influence of \texorpdfstring{$\Theta$}{T} on classifier confidence}
\begin{figure}[t]
 \centering
 \includegraphics[width=0.5\textwidth, height=0.3\textwidth]{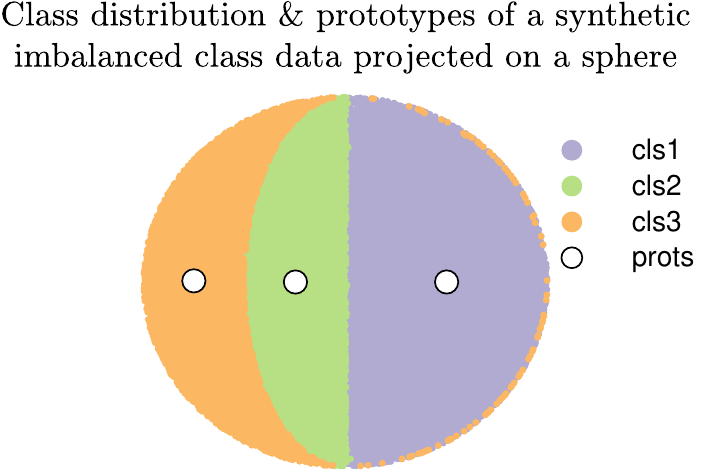}
 \caption{Mollweide projection of a 3D dataset for investigation of the effect of the $\Theta$ value.}
 \label{fig: toyData}
 \includegraphics[width=\textwidth]{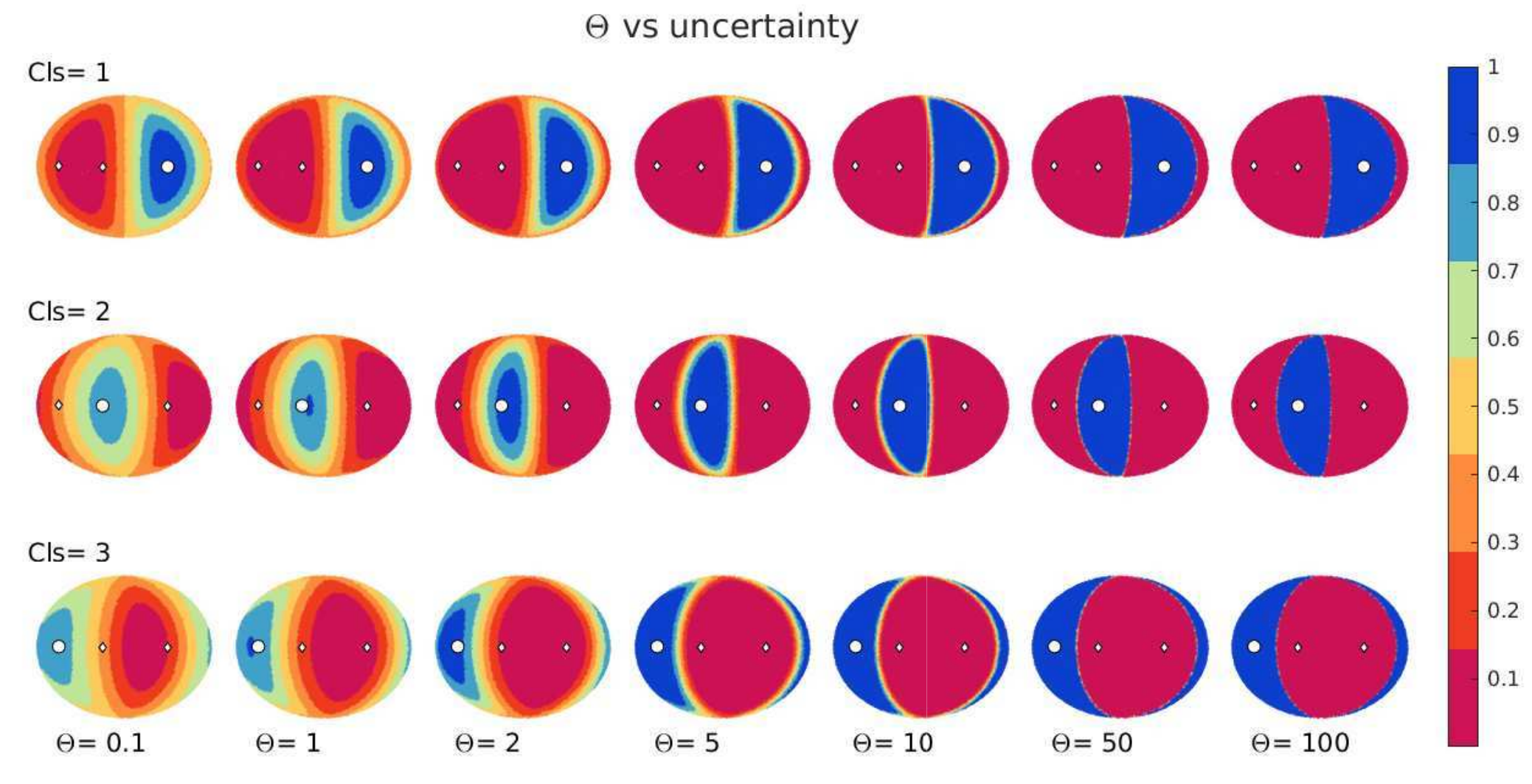}
 \caption{Effect of the value of $\Theta$ in classification uncertainty. The \textbf{o} represents the prototype of the class highlighted in each row.}
 \label{fig: ThetaVariaion}
\end{figure}
We created a three-dimensional synthetic dataset, each sample of which lies on the surface of a sphere. This toy dataset contained 85000 samples which were distributed into three classes in the proportion of 2:1:1, as shown in the Mollweide projection of this dataset (figure \ref{fig: toyData}). Since we were interested in studying the effect of $\Theta$ alone on the region of significant influence and area of regions of uncertainty, we fixed the $\Omega$ and the $\vec{w}^c$ (where $c=1,..,C$ and $C=3$) of a trained $PLVQ^{A^\Theta}$ model and varied only the value of its $\Theta$ before applying the model on the mentioned toy data. In figure \ref{fig: ThetaVariaion} each column corresponds to a value of $\Theta\in\{0.1,1,2,5,10,50,100\}$ and each row depicts the regions of probability for class 1, 2 and 3. In each sub-figure the Mollweide projection of the samples of the toy dataset are coloured according to the confidence of the model in assigning that sample the label of the class whose prototype is highlighted (big white circle). The heatmaps illustrate how with increasing value of $\Theta$ the regions of uncertainty decreased for all the three classes resulting in more and more crisp decisions. Since we aimed for non-crisp decisions we kept the value of $\Theta$ much below 20 in our experiments on real-world datasets. 

\subsection{Geodesic average model}
\label{sec: Average}
Ensembling is a well known strategy to avoid overfitting and improve on the generalization of machine learning algorithms \citep{breiman1996bagging, parsons2005introduction}. 
However, the improved performance by combining independently trained models comes at the cost: 
first, increased computational and memory cost needed keeping all models the ensemble consists of and second, loosing interpretability even if the individual models provide it. 
As mentioned before the Random Forest is an example of an ensemble classifier based on decision trees build from random subsets of the data. 
The memory and computational costs using the Random Forest grows with the number of trees used and the interpretation in form of feature relevances is proposed as post-processing. 
The transparency of an individual tree as rules for classification is largely lost in the Forest, since the ensemble is a 
nontrivial combination of partially overlapping subspaces. 
In this section we propose and investigate a different strategy, namely to build a geodesic average model 
that retains interpretability while avoiding overfitting effects by combining parameter information of independently trained models. 

\subsubsection{Geodesic average over model parameters}

In order to build an average of $k$ models we compute the geometric mean of each of the model parameters, namely the trained prototypes of each class 
$W_c=\{\vec{w}_{\{c,k\}}\}_{i=1}^k$ and the positive semi-definite matrices $\Lambda_k$. 
We restrict the description for one prototype per class here, each initialized close to the class means. With random initialization one might need to rotate the coordinate system to align the prototypes before averaging. 
If using several prototypes per class the correct index for averaging can be found using the geodesic distance of the set of prototypes within each class.
The $\beta$ (or $\Theta$) parameter is a positive scalar and typically fixed or found by line search. 
Classification by geodesic $LVQ^A$ variants (Eqs.~\eqref{eq:ALVQ_diss}, 
\eqref{eq:bOmegaL} and \eqref{eq:E_pALVQ}) 
takes place on the hypershpere and the geometric mean of the model prototypes of each class $\overline{\vec{w}}_c\in \mathcal{M}$ 
in the Riemannian interpretation, known as Karcher mean \citep{Karcher1977}, is the point in $\mathcal{M}$ that minimizes the sum of squared geodesic distances:
\begin{equation}
\label{eq:mean_wc}
\overline{\vec{w}}_c = \arg\min_{\vec{w}\in\mathcal{M}} \sum_{i=1}^k d_\mathrm{geod}(\vec{w}_{\{c,i\}},\vec{w})^2 \qquad \text{with }c\in\{1,\dots,C\} \enspace ,
\end{equation}
with $\vec{w}_{\{c,i\}}$ being the prototype of class $c$ of individual model $i$.
In the Euclidean LVQ variants, GRLVQ, GMLVQ and LGMLVQ, the geodesic distance is simply Euclidean.
In case of $\mathcal{M}$ being the hypersphere the geodesic distance is $d_\mathrm{geod}(\vec{w}_i,\vec{w}_j)=\cos^{-1}(\frac{\vec{w}_i\vec{w}_j}{\lVert \vec{w}_i\rVert\lVert\vec{w}_j\rVert})$. 
This mean exists and is uniquely defined only as the set of prototypes $W_c$ is contained in an open half-sphere, which means a convexity radius of $\pi/2$, and is typically computed by non-linear optimization methods \citep{Karcher1977,Kendall1990,Krakowski07onthe}. 
However, computing the geometric mean of the positive semi-definite matrices $\Lambda_k$ is less straightforward. 

The computation of geometric means of positive definite (PD) matrices as proposed by \cite{ANDO2004305} has received considerable attention 
due to its relevance for numerous applications, ranging from control theory, convex programming, mercer kernels and diffusion tensors in medical imaging. 
However, the computation of this \textit{Ando mean} is not rank-preserving, resulting almost surely in  a rank null for matrices with rank $M<D/2$ \citep{BONNABEL20133202}. Due to the growing interest in low-rank approximations in large-scale applications, \cite{Bonnabel2009,BONNABEL20133202} introduced and extended the geometric mean to the set of positive semi-definite (PSD) matrices $S^+(M,D)$ of fixed rank $M$ using a Riemannian framework. 
Their approach bases on the decomposition of each of the $k$ metric tensors
\begin{equation}
\label{eq:PSD_decomp}
\Lambda_i = U_iR_i^2U_i^\top \qquad \text{for }i=1\dots k
\end{equation}
exhibiting the geometric interpretation of PSD matrices in $S^+(M,D)$ as flat $M$-dimensional ellipsoids in $\R^D$.
Here $U_i$ is element of the Stiefel manifold $St(M,D)$, which denotes the set of all orthonormal $M$-frames in $\R^D$.

Thus, the columns of each $U_i$ forms an orthonormal basis of the $M$-dimensional subspace the corresponding flat ellipsoid is embedded in 
and each $R_i^2$ is an $M\times M$ PD matrix that defines the ellipsoids shape in that low rank cone. 
\cite{BONNABEL20133202} proposes that the Karcher mean of the $k$ $M$-dimensional subspaces $U_i$ serves as a basis for the mean of the $\Lambda_k$ 
where all flat ellipsoids are brought to by a minimal rotation. 
In that common subspace the problem reduces to the computation of the geometric mean of $k$ rank $M$ PD matrices. The implementation of their proposed mean for an arbitrary number of PSD matrices is outlined in 
Algorithm \ref{alg:PSDmean}\footnote{We provide the Matlab code at \url{https://github.com/kbunte/geodesicLVQ_toolbox}}.
For more information about the rank preserving PSD mean and its properties we refer the reader to \cite{BONNABEL20133202}.

\alglanguage{pseudocode}
\begin{minipage}[t]
{0.96\textwidth}
\renewcommand*\footnoterule{}
\begin{savenotes}
\begin{algorithm}[H]
\small
\caption{Computation of geometric rank preserving PSD mean}
\label{alg:PSDmean}
\begin{algorithmic}[1]
\Procedure{$\mathbf{PSDmean}$}{$\{\Lambda_i\}_{i=1}^k$}
    \For {$i = 1 \to k$}
      \State compute eigenvalue decomposition $\Lambda_i = U_iR_i^2U_i^\top$
    \EndFor
    \State compute an orthonormal basis $V$ on the Stiefel manifold $St(M,D)$ of the Karcher mean of the $k$ subspaces $U_i$
    \footnote{The Karcher mean of a set of $M$-dimensional subspaces of $\R^D$ on Grassmann manifold $Gr(M,D)$ is unique in a geodesic ball of radius less than $\pi/(4\sqrt{2})$ \citep{Afsari2011}     
    and can be found by minimal rotation, as provided in the 
    SuMMET package \citep{Marrinan2014}. 
    }
    \For {$i = 1 \to k$}	      
      \State compute two orthogonal matrices $O_i$ and $O_i^V$ by 
      SVD of $U_i^\top V$
      \footnote{These bases remove ambiguity in the definition of the PSD mean chosing particular bases $Y_i$ of the fibers $U_iO(M)$ and bases $V_i$ of the mean subspace fiber $VO(M)$ 
      building the endpoints of the geodesic in the Grassmann manifold \citep{BONNABEL20133202}.}
      \State compute bases $Y_i=U_iO_i$
      \State compute bases $V_i=VO_i^V$
      \State with $\Psi_i^2=Y_i^\top \Lambda_iY_i$ the ellipsoid of $\Lambda_i$ rotated to the mean subspace is $V_i\Psi_i^2V_i^\top$ 
      \State express the ellipsoids in a common basis $V$: $T_i^2=V^\top V_i\Psi_i^2V_i^\top V$ 
    \EndFor
    \State compute the ando mean $\overline{A}(T_1^2,\dots,T_k^2)$ in the low-rank cone \footnote{Methods are proposed in \cite{ANDO2004305,ARNAUDON20121437,Bini} and  
    we used the mmtoolbox implementation by the latter.}
    \State \textbf{return} the geometric mean $\overline{\Lambda} = V\overline{A}(T_1^2,\dots,T_k^2)V^\top$
\EndProcedure
\end{algorithmic}
\end{algorithm}
\end{savenotes}
\end{minipage}

\subsubsection{Convex combinations of models}
LVQ models approximate the solution to non-convex problems and as such may converge to different local optima in independent training runs and the complexity of the problem. 
We expect that the model resulting from averaging over models from different local optima might exhibit inferior performance  compared to its original contributors. 
Therefore we investigate convex combinations of Matrix LVQ models empirically and propose a clustering strategy to distinguish models to build local averages. 
For the prototypes of the models the Karcher mean, Eq.\ \eqref{eq:mean_wc}, can be generalized to a weighted mean or convex combination:
\begin{equation}
\label{eq:weighted_wc}
\widehat{\vec{w}}_c = \arg\min_{\vec{w}\in\mathcal{M}} \sum_{i=1}^k \alpha_i d_\mathrm{geod}(\vec{w}_{\{c,i\}},\vec{w})^2 \quad \text{with }c\in\{1,\dots,C\}\text{, }
\alpha_i\ge0\text{ and }\sum_i\alpha_i=1 \enspace .
\end{equation}
To the best of our knowledge an analytical solution for the weighted mean does not exist and several iterative strategies were proposed
\citep{Clark1984,wagner1990,wagner1992,Watson1983,Alfeld1996BernsteinBzierPO}. 
Two fast iterative solutions exhibiting linear and quadratic convergence for spheres can be found in \cite{Buss01sphericalaverages}. 

\cite{BONNABEL20133202} provided an analytical solution for the weighted average of two positive semi-definite matrices $\Lambda_1$ and $\Lambda_2$ $\in S^+(M,D)$, 
which can be summarized as follows. 
It bases on the same decomposition as stated in Eq.\ \eqref{eq:PSD_decomp}, i.e.\ $\Lambda_1=U_1R^2_1U_1^\top$ and $\Lambda_2=U_2R^2_2U_2^\top$ defined up to an orthogonal 
transformation $O\in O(M)$\footnote{$O(M)$ denotes the general orthogonal group in dimension $M$} 
and hence $A_i=U_iR^2_iU_i^\top=U_iO_i(O_i^\top R^2_iO_i)O_i^\top U_i^\top$. 
The equivalence classes $U_iO(M)$, called fibers, denote all bases that correspond to the same $M$-dimensional subspace $U_iU_i^\top$. 
While the orthongonal transformations do not affect the Grassmann\footnote{Grassmann $Gr(M,D)$ denotes the space of all $M$-dimensional linear projectors in $\R^D$} 
mean of subspaces they do effect the Ando mean of the low-rank PD matrices 
$\overline{A}(R^2_1,R^2_2)\neq\overline{A}(R^2_1,O^\top R^2_2O)$
which causes the problems with the definition of a geometric mean. 
To deal with the ambiguity \cite{BONNABEL20133202} proposed to compute particular representatives $Y_1=U_1O_1$ and $Y_2=U_2O_2$ as bases of the fibers, 
obtained by SVD of $U_1^\top U_2 = O_1(\cos\Sigma)O_2^\top$ using the matrix cosine. 
These two bases correspond to the endpoints of the geodesic in the Grassman manifold that minimize the distance between two fibers in the 
Stiefel manifold $St(M,D)$. These are than used to define a geodesic between $Y_1$ and $Y_2$ containing the convex combinations or $t$-weighted mean
\begin{align}
\label{eq:tweightedMean}
Y(t) = Y_1\cos\Sigma t+X\sin\Sigma t \quad\text{with }t\in[0,1]\enspace ,
\end{align}
where $\Sigma$ is the diagonal matrix containing all principal angles and $X=(Y_2-Y_1\cos\Sigma)(\sin\Sigma)^{-1}$. 
Note that the half-way point  $Y(0.5)$ is the Riemannian mean of $Y_1$ and $Y2$. 
Than the representative PD matrices for the $M$-dimensional ellipsoids in the low rank cone in the corresponding subspaces are given by 
$\Psi_i=Y_i^\top\Lambda_iY_i$. 
Following \cite{Lawson2016} the convex combination (or $t$-weighted mean denoted by $\#_t$) of these two PD matrices is computed as
\begin{align}
\Psi_1\#_t\Psi_2 = \Psi_1^{1/2}\left( \Psi_1^{-1/2}\Psi_2\Psi_1^{-1/2} \right)^{t}\Psi_1^{1/2}\enspace .
\end{align}
Finally, having all the necessary ingredients, the convex combination of the SDM matrices $\Lambda_1$ and $\Lambda_2$ is computed by the $t$-weighted mean \citep{BONNABEL20133202}:
\begin{align}
\Lambda(t) = Y(t) (\Psi_1\#_t\Psi_2) Y(t)^\top \enspace .
\end{align}

\subsubsection{Clustering of Matrix LVQ models}
\label{sec:clustering}
\begin{figure}
\includegraphics[width=\textwidth]{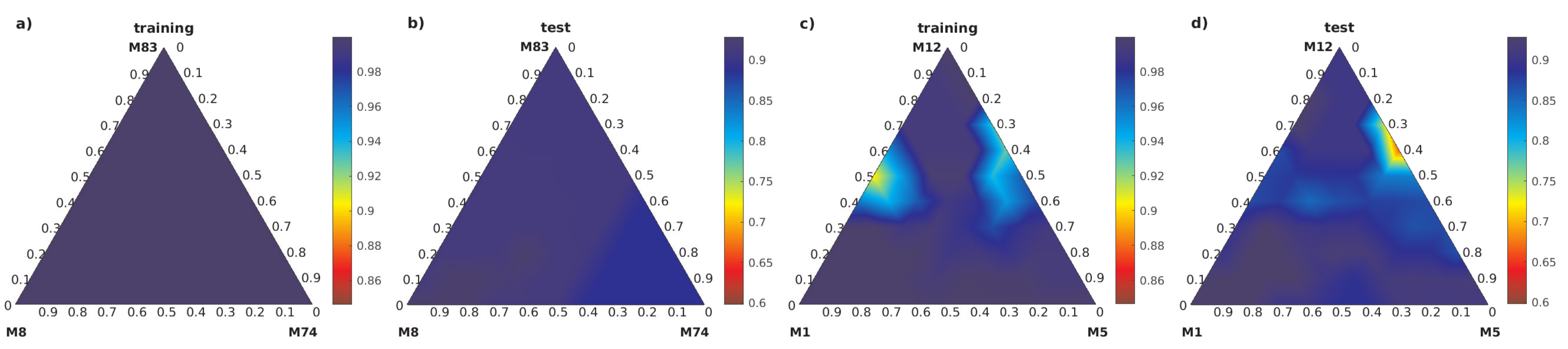}
\caption{Visualizations of macro averaged accuracies in training and test of the convex hull build by three $PLVQ^A$ models from the same 
(a,b) 
and three different clusters 
(c,d). 
}
\label{pic:ConvexCombinations}
\end{figure}

In order to avoid averaging across local optima we propose a clustering strategy based on the Grassmann distance between the bases of the fibers $U_i$ from the decomposition of the metric tensors $\Lambda$, see Eq.\ \eqref{eq:PSD_decomp} and the text above \eqref{eq:tweightedMean}.
The Grassmann distance $d_{Gr}(U_i,U_j) = \lVert\vec{\Sigma}\rVert_2$ is computed using the principal angles $[\Sigma_1,\dots,\Sigma_m]$, which are collected in the diagonal matrix $\Sigma$ 
obtained by SVD of the product of the subspaces $U_i^\top U_j = O_i(\cos\Sigma)O_j^\top$. 
In case of localized class-wise metric tensors $\Lambda^c=\Omega^{c\top}\Omega^c$ we compute the Grassmann distance for each of the $c$ projectors and use the 
average distance for clustering.
We employ agglomerative hierarchical clustering using Ward Linkage on the pairwise Grassman distances and extract cluster memberships varying the numbers of clusters. 
Afterwards we compute the geodesic average model using only members of the same cluster and compute the macro averaged accuracy on the training set to select the best clustering. 
Of course different cluster methods could be used as well, such as for example variations of Grassman k-Means \citep{Turaga2011,Shirazi2012,Carson2017}. 
Furthermore, the Matlab ManOpt toolbox\footnote{\url{http://www.manopt.org}} 
provides a rich collection of algorithms for a variety of manifold optimization problems. 
However, we decided to use hierarchical clustering, since we have typically a comparable low number of models, such that the squared complexity with the number of instances does not state a problem and it avoids further introduction of local optima as is expected using k-Means or Gaussian Mixture Model approaches. 
Furthermore, the cluster memberships for different numbers of clusters can be easily extracted without the need of re-running the method. 
Figure \ref{pic:ConvexCombinations} shows the macro averaged accuracies of the convex hull build by three probabilistic $LVQ^A$ models trained on the GCMS data with rank $M$ set to three. 
The first two panels depict the training and test set performances of the closest models within the same cluster, while the latter 2 panels show 
the performance of models taken from three different clusters.
It can be seen that the convex combination of metric tensors from different clusters can lead to inferior performance, while it can improve using models from the same cluster. 
Therefore, we propose to extract 2-$k$ clusters, compute the average model of each and look at an elbow in the training performance.

\section{Synthetic datasets and experiments} 
\label{sec: Synthetic}

This section describes a synthetic dataset\footnote{The synthetic data is made publicly available in \url{https://git.lwp.rug.nl/cs.projects/angleLVQtoolbox.git}} we modeled to simulate the aforementioned problems, such as low amounts of training data and missing values, often encountered in biomedical data analysis. On this synthetic dataset we introduce once the MNAR and once the MCAR type of missingness, vary the amount of missing data, and the sample size of the training data to study the influence of each of these variations on different classifiers discussed in the previous sections. 
\subsection{Synthetic dataset description}
\begin{figure}[t]
	\centering	
 \parbox{0.42\textwidth}{ 
\includegraphics[width=0.95\linewidth,clip,trim={2.cm 0.25cm 0.8cm 0.8cm}]{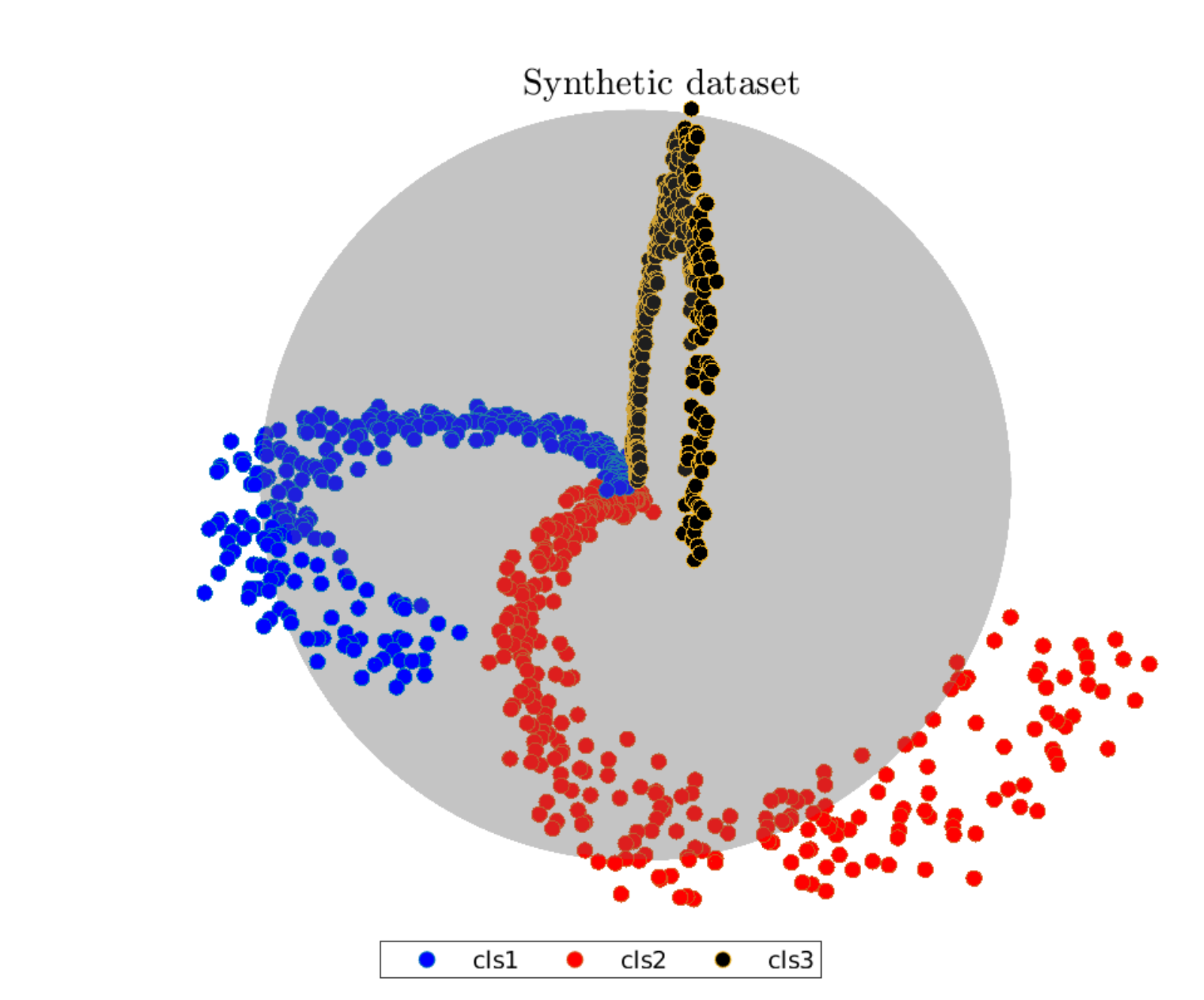}
}\parbox{0.58\textwidth}{ 
\includegraphics[width=0.95\linewidth,clip,trim={0.33cm 0.33cm 1.7cm 0.25cm}]{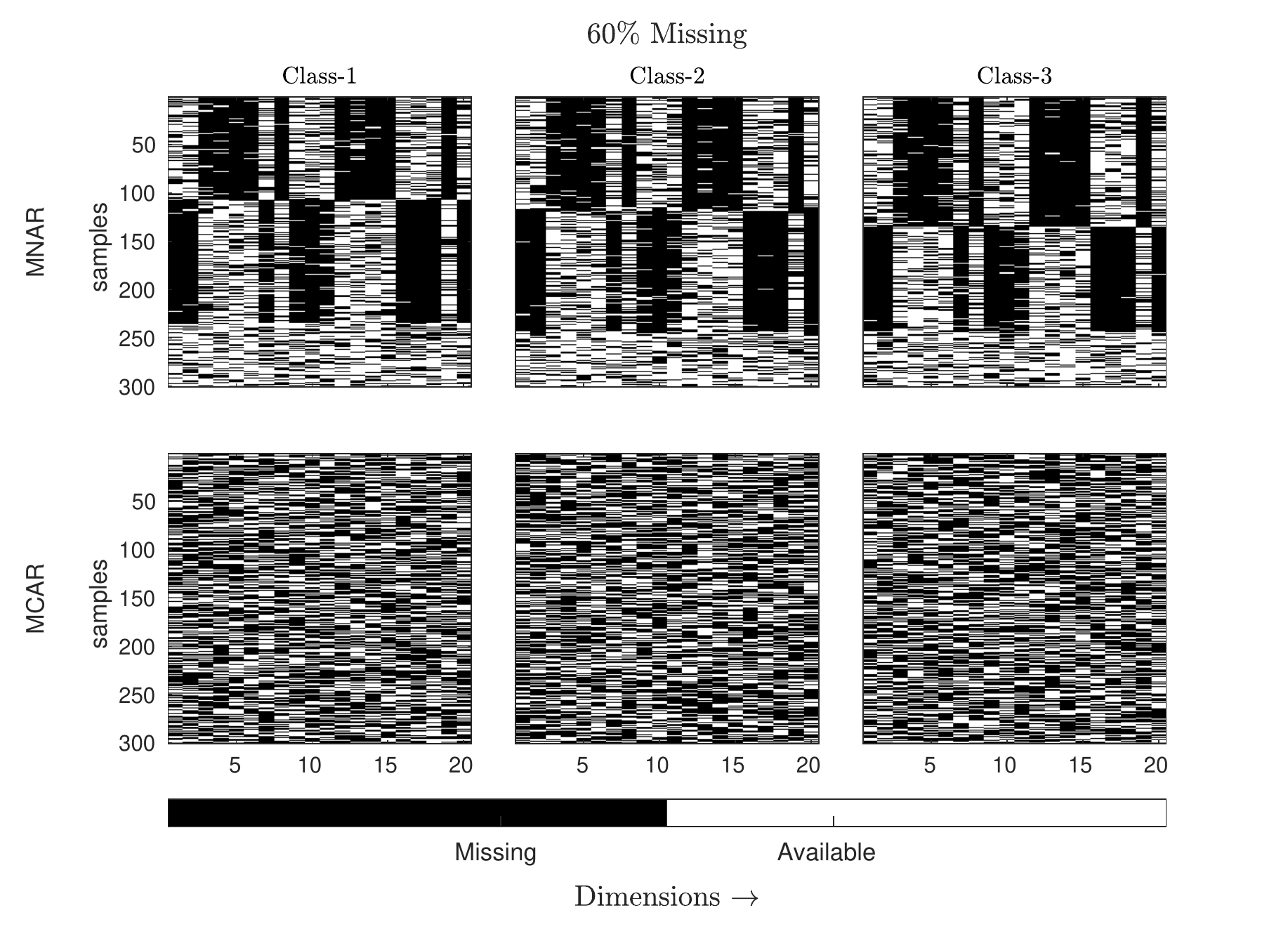} 
}
\caption{Synthetic data: plot of the 3 informative dimensions (left) and class-wise heatmaps of subject-wise average 60\% missingness of type MNAR and MCAR (right). }\label{fig:synData}
\end{figure}

The synthetic dataset $\chi_\textrm{syn}$ was created with three informative dimensions in which three classes are arranged on two-dimensional manifold arcs bending in 3D and overlapping with their narrow parts in the center of a sphere. Similar to the real biomedical dataset the absolute values are not very informative in this arrangement. We created 300 samples per class as shown in the left panel of Figure \ref{fig:synData}. An independent test set consists of 30,072 samples generated similarly. 
To increase the complexity we augmented it with nonlinear transformations of the three informative dimensions and five dimensions of uniform random noise. The non-linear copies were created by taking the base 10 logarithmic transform, and two exponential transforms $e^\chi$, $\chi^3$ and $\chi^5$, resulting in 20 dimensions in total. Next the dataset is successively treated with increasing amount of missingnes of type MCAR and MNAR starting from 10\% to 60\% in steps of 10. For the latter category the dataset was divided into 3 groups randomly, such that the proportion of subjects in the groups were $0.4:0.4:0.2$. Each group could be thought of as a different laboratory or study from which the data was collected. The first two studies measure a few mutually exclusive features and the third group measures all features of the 20 dimensional synthetic data. However, with passage of time each of the first two labs started measuring a few more dimensions than they initially used to, and thus we have a time and study-dependent (systematic) missingness. 
In reality randomly missing samples can exist in addition to systematically missing ones, so we added some random missingness as well. The right panel in Figure \ref{fig:synData} shows the most complicated case with $60\%$ MNAR, where black indicates missing values, while available information is marked white. 
\subsection{Synthetic data experiments}
To study the effect of the training set size on the generalization performance of the classification we successively reduced the amount of data from 80\% to 20\% of the original 900 samples in steps of 10. Now we compare several strategies for classification in the presence of missing data on the synthetic datasets explained above by 10-fold cross-validation (CV). The first strategy bases on generative modeling, namely learning a PPCA model from the data as proposed by \cite{tipping1999mixtures} followed by 
classification by LDA as proposed in \cite{Marlin:2008:MDP:1925592}. The algorithm is abbreviated by $LDA^Q$ in the following, where the subscript $Q$ denotes the latent dimension for Probabilistic PCA. PPCA performed on the full training sets suggests an intrinsic dimensionality of 10 for each percentage of missingness. Another common strategy is multiple imputation, which is model agnostic and can be performed as preprocessing. We imputed each training set using MICE following the predictive mean matching (PMM) strategy \citep{royston2011multiple,azur2011multiple} and generated 10 imputed sets. The resulting model was used to impute the validation and hold-out test set of each CV fold accordingly\footnote{a recent out-of-sample extension for MICE called \emph{mice.reuse} is available at \url{https://github.com/prockenschaub/Misc/tree/master/R/mice.reuse}}. After imputation any classifier such as Random Forest (RF) and $k$-nearest neighbour (KNN) can be used. For the KNN classifier we varied number of nearest neighbours $k$ and type of distance used, namely Euclidean and Mahalanobis and abbreviate the method with $iKNN^{E_k}$ and $iKNN^{M_k}$ respectively. 
\citep{lall1996nearest} suggests that the value of $k$ should be chosen as the square root of the number of training instances. However, since we varied the size of the training set and simultaneously wanted to eliminate the effect of different values of this hyperparameter for the different sizes of the training set, we selected the upper limit of $k\approx\sqrt{162} \approx 12$ for all the sets according to the smallest set being 20\% of the original samples. For RF we selected the number of decision trees to be $150$, which is large enough for a strong ensemble classifier and still smaller than the smallest training set. 

For prototype-based classification we compare the original Euclidean prototype based classifiers GMLVQ with rank $M$ on the imputed data abbreviated by $iLVQ^{E_M}$ and the NaNLVQ able to deal with missing values, accordingly referred to as $LVQ^{E_M}$. 
The geodesic Angle LVQ extension ($LVQ^{A_M}$) is performed on the original and imputed data ($iLVQ^{A_M}$) as well to show the influence of the imputation on the performance. The novel probabilistic Angle LVQ variant based on the Kullback-Leibler divergence is in the following abbreviated by $PLVQ^{A_M^\Theta}$. In this experiment we set the hyperparameter $\Theta=1$. Additionally we reduced the rank $M$ to 10 for direct comparison with the LDA strategy. The prototype-based strategies are repeated 5 times with random initialization on each training set. 

\subsection{Synthetic data results}

Table \ref{tab:synResMNAR} reports the performance in terms of classification error (and standard deviation) averaged over the 10 folds CV, when applied on the datasets with MNAR values. The classifier names are abbreviated as introduced before together with the main hyperparameters shown in the subscript and superscript. Prefix $i$ denotes that the classifier is trained and tested on the imputed datasets. In the column names, $T_{fN}^{z}$ exhibits the training error and $HO_{fN}^{z}$ the corresponding hold-out test error, where the factor $f$ indicates the fraction of the number of samples $N$ used for training and $z$ marks the average percentage of missingness per sample. Table \ref{tab:synResMNAR} shows that RF exhibits the lowest error in the hold-out test test. 
However we also observed that RF suffers from significant overfitting effect. This table further indicates that throughout the experimental settings (variation of amounts of missingness and available data for training), the performance of $LVQ^{A_{10}}$ is more stable than that of its Euclidean counterpart even for the lowest rank of $\Omega$ matrix. With regards to the KNN, the choice of distance measure seems to have a stronger effect than the choice of $k$ for this data. Comparing the LDA and the LVQs we find that the effect of the number of principal components is more pronounced in the former than the effect of rank of $\Omega$ is for the LVQs.
\def\myColW{0.095\textwidth}
\begin{table}[t] 
\caption{Selected average training $T_{fN}^z$ and hold-out test $HO_{fN}^z$ errors for fraction $f$ number of training samples $N$ and on average $z$\% of missingness of type MNAR.}
\label{tab:synResMNAR}
\begin{tabularx}{\textwidth}{
@{\extracolsep{\fill}}
@{}>{\raggedright\scriptsize}p{0.09\textwidth}@{} *{6}{@{}>{\raggedright\scriptsize}p{\myColW}@{}} >{\scriptsize\arraybackslash}r@{} 
}
\toprule
Classifier & 
\makecell[c]{$T_{N}^{0\%}$} & \makecell[c]{$HO_{N}^{0\%}$} & \makecell[c]{$HO_{0.2N}^{0\%}$}  & \makecell[c]{$HO_{N}^{30\%}$} & 
\makecell[c]{$HO_{0.2N}^{30\%}$} & \makecell[c]{$HO_{N}^{60\%}$} & \makecell[c]{$HO_{0.2N}^{60\%}$} \\
\toprule
$ikNN^{E_{12}}$ & 0.06 (0.01) & 0.15 (0.01) & 0.23 (0.05) & 0.22 (0.03) & 0.28 (0.04)  & 0.41 (0.01) & 0.45 (0.02) \\
$ikNN^{M_{12}}$ & 0.02 (0)    & 0.08 (0.01) & 0.23 (0.07)  & 0.23 (0.04) & 0.34 (0.05) & 0.45 (0.01) & 0.52 (0.02) \\
$ikNN^{E_{5}}$  & 0.05 (0.01) & 0.17 (0.01) & 0.26 (0.05)   & 0.26 (0.03)  & 0.30 (0.04) & 0.43 (0.01)  & 0.47 (0.02) \\
$ikNN^{M_5}$    & 0.02 (0)    & 0.12 (0.01) & 0.25 (0.06)   & 0.28 (0.04)  & 0.36 (0.05) & 0.48 (0.01)  & 0.53 (0.02) \\
$iRF_{150}$     & 0 (0)       & 0.01 (0)    & 0.02 (0.01)   & 0.06 (0.01)  & 0.08 (0.02) & $\mathbf{0.25 (0.01)}$ & $\mathbf{0.30 (0.01)}$ \\
$iLVQ^{E_{10}}$  & 0.02 (0)   & 0.02 (0)  & 0.07 (0.04)   & 0.15 (0.02)  & 0.21 (0.04) & 0.36 (0.01) & 0.43 (0.03) \\
$iLVQ^{A_{10}}$  & 0 (0)        & 0.01 (0)  & 0.08 (0.05)   & 0.14 (0.02)  & 0.20 (0.04) & 0.35 (0.01) & 0.41 (0.03) \\
$iLVQ^{E_{20}}$  & 0.02 (0.01)  & 0.02 (0.01) & 0.07 (0.03)   & 0.15 (0.02)  & 0.21 (0.04) & 0.36 (0.01) & 0.43 (0.03) \\
$iLVQ^{A_{20}}$  & 0 (0)        & 0.01 (0)  & 0.08 (0.05)   & 0.14 (0.02)  & 0.20 (0.04) & 0.35 (0.02) & 0.42 (0.04) \\
\midrule
$LDA^{Q_{10}}$     & 0.01 (0.01) &0.17 (0.03) &0.26 (0.07) &0.25 (0.03) & 0.30 (0.05) &0.38(0.03) &0.40 (0.03)  \\
\midrule
$LVQ^{E_{20}}$   & 0.02 (0.01)  & 0.02 (0.01) & 0.07 (0.03)   & 0.15 (0.01)  & 0.21 (0.04)  & $\mathbf{0.30 (0.01)}$& $\mathbf{0.35 (0.03)}$ \\
$LVQ^{A_{20}}$   & 0 (0)        & 0.01 (0.01) & 0.07 (0.05)   & 0.14 (0.02)  & 0.20 (0.02)  & $\mathbf{0.27  (0.01)}$ & $\mathbf{0.35 (0.05)}$ \\
$LVQ^{E_{10}}$   & 0.02 (0)   & 0.02 (0)  & 0.07 (0.04)   & 0.15 (0.01)  & 0.23 (0.04)  & 0.31 (0.01)& 0.37 (0.03) \\
$LVQ^{A_{10}}$   & 0 (0)        & 0.01 (0)  & 0.08 (0.05)   & 0.14 (0.02)  & 0.21 (0.04)  & $\mathbf{0.27 (0)}$ & $\mathbf{0.35 (0.04)}$ \\
$PLVQ^{A_{10}^1}$&  0 (0)       & 0.01 (0)   & 0.01 (0)    & 0.15 (0.03)  & 0.16 (0.03) & $\mathbf{0.27 (0.01)}$   & $\mathbf{0.28 (0.02)}$\\
$LVQ^{LA_{10}}$ & 0.01 (0) & 0.05 (0.03)& 0.13 (0.07) & 0.13 (0.02) & 0.23 (0.05) & 0.24 (0.02) & 0.36 (0.04)\\
\bottomrule
\end{tabularx} 
\end{table}

We investigate whether the 
superior performance by RF is due to ensembling. Therefore we train 
a system of 150 $LVQ^{A_{20}}$ on the exact same imputed subsets of training data that each of the 150 DTs of the RF had trained on, 
on the most difficult setting (60$\%$ MNAR and training set reduced to 20$\%$ of its original size). 
The mean generalization error from the system of $iLVQ^{A_{20}}$ is \textbf{0.39 (0.02)} and that from $LVQ^{A_{20}}$ is \textbf{0.32 (0.01)} against RF's \textbf{0.30 (0.01)}. This additionally confirms that imputation does adversely affect the performance of $LVQ^{A}$ classifiers. Since ensembling compromises with the interpretability of a classifier we applied geodesic averaging to our classifier, which resulted in a generalization error of 0.31 (0.01), thus comparable to RF with 150 DTs trained on the exact same subset of training data, indicating that ensembling and averaging strategies are indeed beneficial. Next we compare and discuss the performance of the classifiers on the aforementioned MCAR datasets. For each of the classifiers, only the most promising hyperparameter settings (based on the validation set performance) were applied. Hence, in the following experiments we omit imputation for algorithms that handle the missingness internally. Also since we know that there are 10 intrinsic dimensions 
based on PPCA and EVD, we restrict the rank of $\Omega$ to 10, and keep the latent dimension of covariance matrix 10 for the LDA. 
\begin{table}[t]
\caption{Selected average training $T_{fN}^z$ and hold-out test $HO_{fN}^z$ errors for fraction $f$ number of training samples $N$ and on average $z$\% of missingness of type MCAR.
}
\label{tab:synResMCAR}
\begin{tabularx}{\textwidth}{@{\extracolsep{\fill}}
@{}>{\raggedleft\scriptsize}p{0.09\textwidth}@{}*{6}{@{}>{\raggedright\scriptsize}p{\myColW}@{}} >{\scriptsize\arraybackslash}l@{}
}
\toprule
Classifier & 
\makecell[c]{$T_{N}^{0\%}$} & \makecell[c]{$HO_{N}^{0\%}$} & \makecell[c]{$HO_{0.2N}^{0\%}$}  & \makecell[c]{$HO_{N}^{30\%}$} & 
\makecell[c]{$HO_{0.2N}^{30\%}$} & \makecell[c]{$HO_{N}^{60\%}$} & \makecell[c]{$HO_{0.2N}^{60\%}$} \\
\toprule
$ikNN^{E_{12}}$ & 0.06 (0.01)	& 0.15 (0.01)	&	0.23 (0.05)	&	0.23 (0.01)	&	0.23 (0.01)	&	0.38 (0.01)	&	0.38 (0.01) \\
$iRF_{150}$ & 0 (0)	&	0.01 (0)	&	0.02 (0.01)	&	0.06 (0)	&	0.06 (0)	&	0.23 (0.01	&	$\mathbf{0.23 (0.01)}$\\
\midrule
$LDA^{Q_{10}}$ & 0.1 (0.01)	&	0.17 (0.03)	&	0.26 (0.07)	&	0.21 (0.03)	&	0.22 (0.03)	&	0.33 (0.03)	&	0.33 (0.03)\\
\midrule
$LVQ^{E_{10}}$ & 0.02 (0)	&	0.02 (0)	&	0.07 (0.04)	&	0.14 (0.01)	&	0.16 (0.02)	&	0.28 (0.01) 	&	0.28 (0.02)\\
$LVQ^{A_{10}}$ & 0 (0)	&	0.01 (0.01)	&	0.08 (0.05)	&	0.13 (0.01)	&	0.14 (0.02)	&	0.28 (0.02)	    &	0.28 (0.03)\\
$PLVQ^{A_{10}}$ &  0 (0)    & 0.01 (0)   & 0.01 (0)    & 0.15 (0.02)   & 0.16 (0.03)  & 0.29 (0.02) & 0.29 (0.02)\\ 
$LVQ^{LA_{10}}$ & 0.01 (0) & 0.05 (0.03) & 0.13 (0.07) & 0.15 (0.02) & 0.16 (0.03) & 0.27 (0.02) & 0.35 (0.03)\\
\bottomrule
\end{tabularx}
\end{table}

\begin{figure}[t]
\centering
\includegraphics[width=\textwidth,trim={2.8cm 0.8cm 2.5cm 0.5cm},clip]{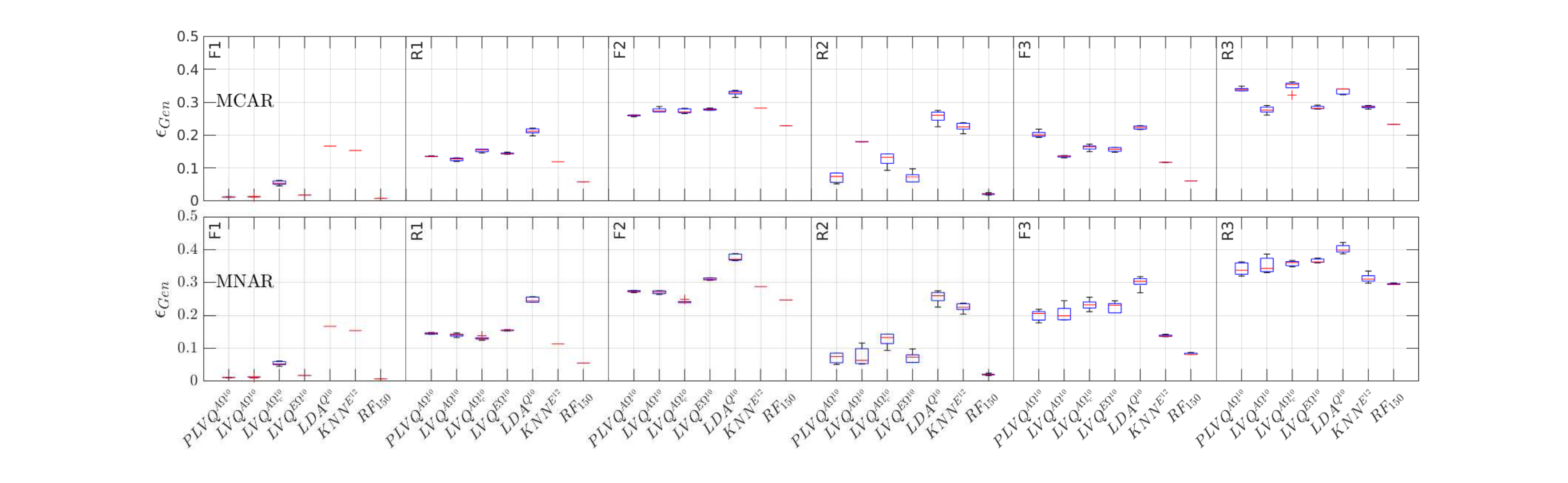} 
\caption{Overview of the classification error plots of 6 methods on the hold-out test set for missingness of type MCAR (top) and MNAR (bottom). The letters F and R mark the full and to 20\% of the original size of the training set, while the number 1-3 indicates 0\%, 
30\% and 60\% missingness respectively.}
\label{fig:MNARvsMCAR}
\end{figure}

Figure \ref{fig:MNARvsMCAR} provides visual summary of the generalization performance of the aforementioned classifiers, i.e., KNN with $k=12$ neighbours, $LVQ^A$ and GMLVQ trained on the unimputed data, and LDA with latent dimension of 10.
Figure \ref{fig:MNARvsMCAR} shows the performance of $PLVQ^A$, $LVQ^A$, $LVQ^E$, and LDA trained on unimputed data and KNN and RF on the imputed dataset. Comparison of tables \ref{tab:synResMNAR} and \ref{tab:synResMCAR}, and figure \ref{fig:MNARvsMCAR} illustrate that the difference is performance of the LVQ classifier with parameterized cosine dissimilarity measure and that with Euclidean distance measure is prominent for systematic missingness only. Similarly KNN and LDA are also less prone to error when the missingness type is MCAR. Thus, while the $LVQ^A$ classifier shows similar performance for MCAR missingnes, it is superior with respect to its Euclidean counterparts when the missingness is of type MNAR. Even though RF with 150 DTs have a slightly lower error rate that of the LVQ classifiers, our invetigation confirmed that it is because of ensembling. Since the motivation behind table \ref{tab:synResMNAR} was to show the difference in influence of the MCAR and MNAR type of missingness, we have not repeated the experiment with ensembling for this part.

\section{Computer aided diagnosis of inborn disorders of steroidogenesis, based on urine steroid metabolite excretion data} 
\begin{figure}[t]
\centering 
\includegraphics[width=\textwidth,trim={2cm 0cm 1.2cm 0.2cm},clip]{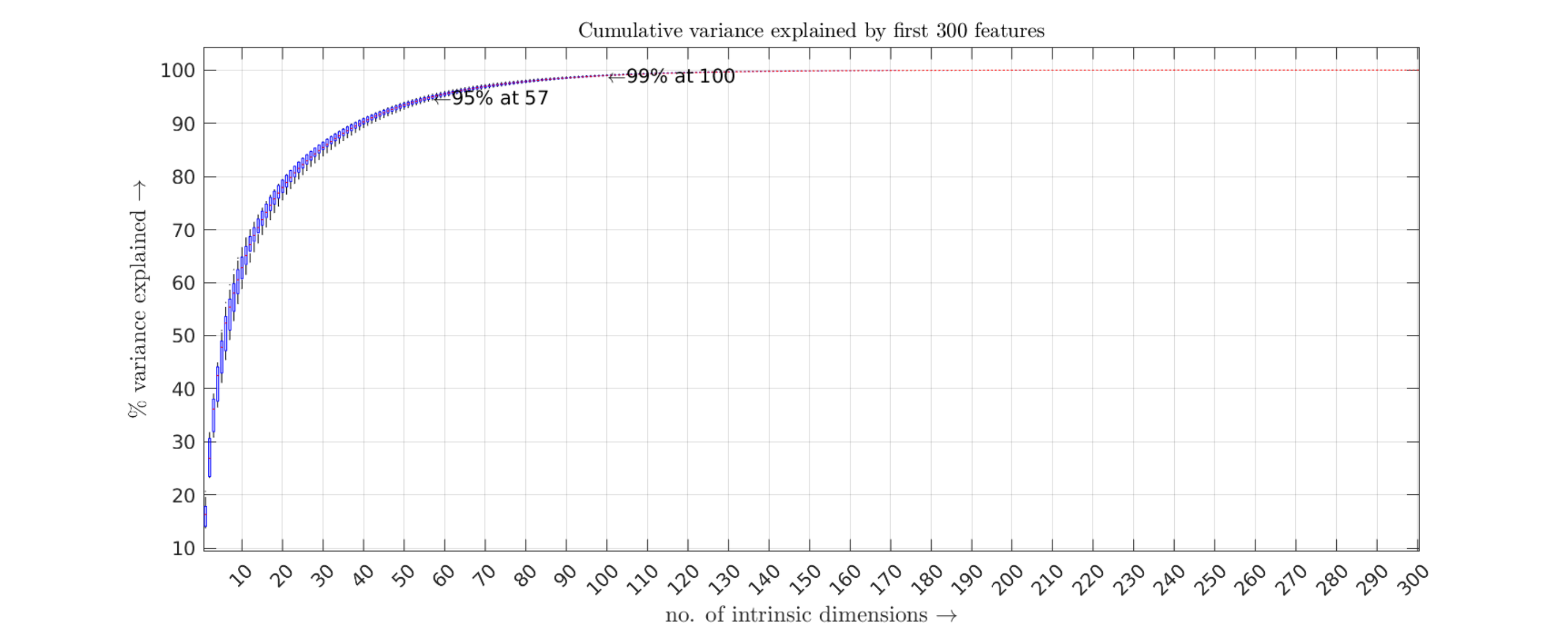} 
\caption{The cumulative variance shows that the 57 dimensions together explain $95\%$, and 100 dimensions explain more than $99\%$ of the variance of the dataset.}
\label{fig:PPCA}
\end{figure}

Due to the limited number of samples for the rare disorders of steroidogenesis it is impossible 
to keep a hold-out test. Therefore, we validate the performance of the classifiers using 5-fold cross-validation, dividing the folds with a comparable number of subjects from each condition (preserving the class distribution). 
Following the outcome from the series of experiments performed on the synthetic datasets, 
we did not use imputation on this real-life data for any algorithm which can handle missing data implicitly. 
The data is preprocessed by z-score transform with the mean and standard deviation determined by each training set and consecutively used in the corresponding test set. 
An exploratory analysis with EVD and probabilistic PCA (with the number of latent dimensions $Q=300$) is performed, to estimate the number of intrinsic dimensions of the data. Both PPCA (figure \ref{fig:PPCA}) and EVD suggest that $\approx 57$ and $100$ are able to explain $95\%$ and $99\%$ of the variances of the dataset, respectively. EVD of $\Lambda=\Omega^\top \Omega$ with $\Omega\in\R^{57\times D}$ revealed an intrinsic dimensionality for classification of $M=6$. We additionally experimented with $\Omega\in\R^{3\times D}$ that allows the visualization of the decision boundaries. Experiments on each fold were repeated at least 5 times with random initialization of elements $\Omega_{ij}\in[-1,1]$.

\begin{table}[!t]
\scriptsize
\caption{Experiments on the GCMS data. 
All experiments were performed on 5 folds of cross-validation, using stratified sampling, and repeated random initializations 
per fold. 
} 
\label{tab:GCMS_Exp} 
\begin{tabularx}{\textwidth}{@{\extracolsep{\fill}} l m{0.8\textwidth}  } 
\hline
 \textbf{Algorithms} & \textbf{Hyperparameters and experiment description}\\  
\toprule
$LDA^{Q_M}$         & PPCA for latent dimension of $M$=100 and 57, SMOTE (imbalance)\\
$iKNN_\kappa^{E}$   & MICE (imputation), SMOTE (imbalance), and $k\in\{3,5,7\}$\\
$iRF_t$             & MICE (imputation), SMOTE (imbalance), number of DTs $t\in\{7,50,100\}$\\
\midrule
$LVQ^{E_M}$         & SMOTE (imbalance), 1 prot/class, Rank$(\Lambda)$ $M\in\{100,57,6,3\}$\\
$LVQ^{A^\beta_M}$   & Geodesic SMOTE (imbalance), 1 prot/class, Rank$(\Lambda)$ $M\in\{100,57,6,3\}$\\
$LVQ^{LA^\beta_M}$  & Geodesic SMOTE (imbalance), 1 prot/class, Rank$(\Lambda^L)$ $M\in\{100,57,6,3\}$\\
$PLVQ^{A_M^\theta}$ & Geodesic SMOTE (imbalance), 1 prot/class, Rank$(\Lambda)$ $M\in\{100,57,6,3\}$\\
$cPLVQ^{A_M^\theta}$& Cost weight matrix $\gamma_{cp}$ (imbalance), 1 prot/class, Rank$(\Lambda)$ $M\in\{100,57,6,3\}$\\
\midrule
$(cP)LVQ^{A_M^{\{\beta,\theta\}}}_{e\eta}$  & Ensembling (majority vote) of $\eta\in\{5,100\}$ iterations of $(cP)LVQ^{A_M^{\{\beta,\theta\}}}$ for each fold with Rank$(\Lambda)$ $M\in\{100,57,6,3\}$\\
$(cP)LVQ^{A_M^{\{\beta,\theta\}}}_{\#\eta_{\mathrm{\upsilon}}}$ & Geodesic average of $\mathrm{\upsilon}$ clusters over $\eta=100$ 
$(cP)LVQ^{A_M^{\{\beta,\theta\}}}$ models for each fold with Rank$(\Lambda)$ $M\in\{6,3\}$\\
\bottomrule
\end{tabularx}
\end{table}
Table \ref{tab:GCMS_Exp} summarizes the experiments performed on the metabolite ratios of the GCMS dataset. 
GMLVQ using Euclidean distance is indicated by $LVQ^E$, GMLVQ using angle-based dissimilarity is represented by $LVQ^{A_M^\beta}$, where $M$ denotes the rank of metric tensor $\Lambda$ and hyper-parameter $\beta$. For this dataset we also experimented with the more complex local variant, namely $LVQ^{A}$ with local metric tensors, which are referred to in the tables as $LVQ^{LA}$. 
The probabilistic $LVQ^{A}$ variant is abbreviated by $PLVQ^{A_M^\Theta}$. For the top-most block all the classifiers, except LDA, were applied on 10 imputed sets of training data per fold. The variability in these classifiers arises from both the imputed sets and the oversampling per iteration. The LDA performance variation per iteration in a fold stem from oversampling only, since it can handle missing data intrinsically. Block 2 investigates the LVQ variants with different hyperparameter settings, trained with at least 5 random initializations in each fold. Most experiments used (geodesic) SMOTE to tackle the class imbalance in a comparable way. Alternatively a cost weight matrix (as introduced in section \ref{sss:PenaltyMat}) can be used, that can penalize certain classification errors more than others. We observed comparable results to the use of geodesic SMOTE and added one result with costs $\gamma_{cp}=\sfrac{1}{6}$ for demonstration. 
The diagonal $\gamma_{c=p}$ and misclassified healthy controls cost is $\sfrac{2}{3}$, 
misclassifications of ISD-4 and ISD-5 for any other disease is $\sfrac{1}{3}$, 
and the highest penalty is induced by a patient being misclassified as healthy, setting the corresponding column off-diagonal elements to 1. The third block indicates the ensembling experiment settings using majority vote and the geodesic averaging strategy. We concentrate on the best hyperparameter settings (based on training performances) of the newly presented intrinsically interpretable classifiers ($LVQ^A$, $LVQ^{LA}$, $PLVQ^A$ and $cPLVQ^A$) from the block before. 
The ensembling experiments are added for two reasons: first they constitute a fairer performance comparison to RF (which is an ensemble of $\eta$ DTs), 
and second the average models (see section \ref{sec: Average}) lend themselves for easy interpretation. For all LVQ models we ensembled first over the 5 initially trained models only, to get a first estimate of the performance gain.  
We only trained additional models for the most promising and interesting hyperparameter settings.
The experiments demonstrating the clustering strategy before building the geodesic average model (see section \ref{sec:clustering}) 
are abbreviated by $(P)LVQ^{A_M^{\{\beta,\theta\}}}_{\#\eta_{\mathrm{\upsilon}}}$. 
$\upsilon=1$ indicates that all models were used and $\upsilon>1$ indicates the average was build in clustered subsets. 
The former is the default setting since the latter is only beneficial if there are significantly different local optima found, 
which usually is encountered when the complexity of the model (global metric tensor, rank and/or number of prototypes) is too small for the classification problem. 
The rank 3 restriction that allows visualization of the decision boundaries is a typical situation where the clustering can be beneficial.

\subsection{Performance comparison on the GCMS ratios data} 
In this section we present the results of the best hyperparameter settings (selected based on training performance) described in table \ref{tab:GCMS_Exp} for LDA, KNN and RF. For LVQ classifiers, their performances on both imputed and unimputed (original ratios) were compared, however as was seen for the synthetic dataset, imputation has an adverse effect and we do not show them. We performed grid search to optimize the hyper-parameter settings of all methods with respect to the training data. For KNN the performance corresponding to Euclidean distance with $\kappa=5$ nearest neighbours has been reported. For the $PLVQ^{A^\Theta_M}$, with rank 3 we found $\Theta=10$ and for rank $M=6$ we found $\Theta=15$ to be good choices. We present the generalization performance from 
those experimental settings which had best training performance, were easy to interpret, and helped in considerable knowledge gain by the medical community. 

\def\myColW{0.081\textwidth}
\LTcapwidth=\linewidth
\begin{table}[!t]
\caption{\scriptsize{GCMS: mean validation performance (and standard deviation) across 5 folds. Evaluation measures include sensitivity (all conditions versus healthy), macro-averaged accuracy (MAvg), and class-wise accuracy. The last 2 blocks show the performance of majority vote ensembling ($(cP)LVQ_{e\eta}$) and the fold-wise average model ($(cP)LVQ_{\#\eta}$).}}
\label{tab:perfAllClassifiers}
\begin{tabularx}{\linewidth}{
@{\extracolsep{\fill}} 
@{}>{\raggedright\tiny}p{0.10\textwidth}@{} *{8}{@{}>{\raggedleft\tiny}p{\myColW}@{}} >{\tiny\arraybackslash}r@{}
}
\toprule
Method	& \makecell[c]{Sensitivity} & \makecell[c]{MAvg} & \makecell[c]{Healthy} & \makecell[c]{ISD-1} & \makecell[c]{ISD-2} &  \makecell[c]{ISD-3} & \makecell[c]{ISD-4} & 
\makecell[c]{ISD-5} & \makecell[c]{ISD-6} \\
\toprule
$iRF_{100}$        & $\mathbf{94.1 (0.03)}$ & $\mathbf{92.4 (0.03)}$ & 99.8 (0.00) & 88.5 (0.16) & 96.7 (0.07) & 90.7 (0.14) & 89.3 (0.17) & 87.1 (0.09) & 94.8 (0.06) \\ 
$iKNN^E_5$              & 86.2 (0.06) & 79.6 (0.06) & 98.0 (0.01) & 61.1 (0.22) & 82.7 (0.30)  & 82.7 (0.13) & 84.9 (0.13) & 66.6 (0.14) & 80.9 (0.14) \\
$LDA^{Q_{100}}$         & 87.7 (0.03) & 80.3 (0.02) & 97.7 (0.01) & 63.0 (0.29) & 76.7 (0.22)  & 83.3 (0.12) & 72.7 (0.23) & 78.6 (0.07) & 90.2 (0.00) \\
\hline
$LVQ^{E_6}$	        & 89.4 (0.04) & 86.2 (0.06) & 99.3 (0.00) & 78.0 (0.14) & 93.3 (0.15)  & 83.3 (0.12) & 76.7 (0.33) & 78.6 (0.15) & 94.0 (0.05) \\
$LVQ^{A_6^{1}}$  & 95.1 (0.04) & 91.3 (0.03) & 98.9 (0.01) & 84.0 (0.22) & 98.7 (0.03)  & 93.3 (0.10) & 89.9 (0.12) & 78.6 (0.17) & 95.7 (0.05) \\  
$PLVQ^{A^{10}_3}$& 95.8 (0.02) & 89.8 (0.03) & 98.2 (0.01) & 81.0 (0.25) & 95.3 (0.10)  & 90.7 (0.13) & 86.9 (0.14) & 81.6 (0.16) & 94.5 (0.06) \\
$PLVQ^{A^{15}_6}$& $\mathbf{96.6 (0.03)}$ & $\mathbf{91.8 (0.03)}$ & 98.1 (0.01) & 85.0 (0.24) &  100 (0.00)  & 91.3 (0.12) & 88.8 (0.17) & 80.5 (0.13) & 98.8 (0.03) \\
$cPLVQ^{A^{15}_6}$      & $\mathbf{97.3 (0.02)}$ & $\mathbf{91.1 (0.04)}$ & 97.2 (0.02) & 84.8 (0.20) & 97.2 (0.10) & 92.4 (0.09) & 89.6 (0.15) & 81.6 (0.14) & 95.0 (0.06) \\
$LVQ^{LA^{1}_3}$	& 95.2 (0.03) & 91.1 (0.03) & 99.0 (0.01) & 85.0 (0.18) & 97.3 (0.06)  & 92.7 (0.09) & 88.0 (0.18) & 78.2 (0.12) & 97.3 (0.05) \\ 
\hline
$LVQ_{e100}^{A^{1}_3}$    & 94.8 (0.03) & 91.7 (0.02) & 99.2 (0.01) & 81.0 (0.21) & 100 (0.00)  & 96.7 (0.07) & 88.0 (0.18) & 78.9 (0.14) & 98.0 (0.04)\\
$LVQ_{e100}^{A^{1}_6}$    & 94.3 (0.03) & 91.4 (0.02) & 99.0 (0.01) & 81.0 (0.21) & 100 (0.00)  & 96.7 (0.07) & 88.0 (0.18) & 78.9 (0.14) & 96.2 (0.05)\\
$PLVQ_{e100}^{A_3^{10}}$         & $\mathbf{96.6 (0.03)}$ & $\mathbf{93.1 (0.03)}$ & 98.9 (0.01) & 86.0 (0.22) & 100 (0.00) & 96.7 (0.07) & 88.0 (0.18) & 81.8 (0.17) & 100 (0.00)\\
$PLVQ_{e100}^{A_6^{15}}$         & $\mathbf{96.6 (0.03)}$ & $\mathbf{93.4 (0.02)}$ & 98.7 (0.01) & 86.0 (0.22) & 100 (0.00) & 96.7 (0.07) & 88.0 (0.18) & 84.3 (0.14) & 100 (0.00)\\
$cPLVQ_{e100}^{A^{15}_6}$        & $\mathbf{97.2 (0.02)}$ & $\mathbf{92.7 (0.03)}$ & 98.4 (0.02) & 91.0 (0.12) & 100 (0.00) & 93.3 (0.09) & 88.0 (0.18) & 81.8 (0.17) & 96.0 (0.05) \\
$LVQ_{e100}^{LA^{1}_3}$   & 94.4 (0.02) & 90.2 (0.02) & 99.3 (0.01) & 76.0 (0.25) & 100 (0.00) & 93.3 (0.09) & 88.0 (0.18) & 78.9 (0.14) & 96.2 (0.05)\\
\hline
$LVQ_{\#100_5}^{A^{1}_3}$ & 94.4 (0.02) & 85.7 (0.08) & 94.9 (0.09) & 72.6 (0.20) & 85.3 (0.23) & 92.7 (0.09) & 83.7 (0.15)& 73.3 (0.18) & 97.3 (0.04)\\
$LVQ_{\#100_5}^{A^{1}_6}$ & 94.6 (0.02) & 91.4 (0.01) & 99.0 (0.01) & 78.5 (0.22) &  100 (0.00) & 96.7 (0.07) & 88.0 (0.18)& 81.6 (0.12) & 96.2 (0.05) \\
$PLVQ_{\#100_4}^{A_3^{10}}$      & $\mathbf{96.5 (0.01)}$ & 89.0 (0.03) & 98.1 (0.01) & 77.8 (0.24) & 92.5 (0.07) & 92.5 (0.07) & 86.0 (0.22) & 81.0 (0.14)& 95.5 (0.04)\\
$PLVQ_{\#100_1}^{A_6^{15}}$      & $\mathbf{96.6 (0.02)}$ & $\mathbf{92.6 (0.02)}$ & 98.4 (0.01) & 86.4 (0.20) & 99.9 (0.00) & 93.8 (0.08) & 88.0 (0.17) & 82.8 (0.14)& 99.3 (0.00)\\
$cPLVQ_{\#100_1}^{A^{15}_6}$     & $\mathbf{97.8 (0.01)}$ & $\mathbf{92.9 (0.04)}$ & 98.1 (0.01) & 91.0 (0.12) & 100 (0.00) & 93.3 (0.09) & 88.0 (0.18) & 81.8 (0.17) & 98.0 (0.04) \\
$LVQ_{\#100_1}^{LA^{1}_3}$& 95.5 (0.01) & 91.2 (0.02) & 99.3 (0.01) & 81.0 (0.21) & 100 (0.00) & 93.3 (0.09) & 88.0 (0.18) & 78.9 (0.14) & 98.2 (0.04) \\
\bottomrule
\end{tabularx}
\end{table}
Table \ref{tab:perfAllClassifiers} shows the most interesting selection of performances of Angle LVQ 
(global and local), RF (with $100$ trees), imputed KNN with $\kappa=5$, LDA with latent dimension $Q=100$, and the original matrix $LVQ^{E_M}$ (imputed and NaNLVQ) \citep{Bunte2011_LiRaMLVQ, NaNLVQ}. Additionally the table also shows the performance from the $PLVQ^{A_M^\Theta}$ to enable comparison between deterministic and probabilistic versions. For a fair comparison with RF, which is an ensemble of DTs, we also report the performances of the majority vote ensemble and geodesic average model of 100 trained $LVQ^{A_M^\beta}$ and $PLVQ^{A_M^\Theta}$ models within each fold. Since the class-wise accuracy of the healthy condition is the same as the specificity we only report the former. The evaluation measure macro-averaged accuracy (MAvg) is the mean of the class-wise accuracies of all the classes. Table \ref{tab:perfAllClassifiers} shows that the imputed RF is superior to imputed KNN or LDA. Furthermore, the experiments demonstrate that the use of the angular dissimilarity in the LVQ models is beneficial for this dataset. Note, that the ALVQ models are also fairly robust with respect to the hyper-parameter setting, with the exception of the rank 3 visualization models that trade some of the performance  for additional interpretability by visualization. Interestingly, the probabilistic variant shows slightly better performance than the original formulation across the settings. $PLVQ^{A_M^\Theta}$ even with ranks 3 and 6 of $\Lambda$ and $LVQ^{LA_M^\beta}$ with rank 3 local metric tensors $\Lambda^L$ show comparable performance to RF. Following an ensembling strategy with 100 LVQ models leads to a fair comparison and especially the $PLVQ^{A}$ models achieve similar or superior performance and exhibit higher sensitivity and macro-averaged accuracy when compared to RF.  
However, any majority vote ensemble loses interpretability and hence we also reported the clustered average model performance, as explained in section \ref{sec: Average}. The overall best performance is achieved with $PLVQ$ ensembles and the average model over 100 random initializations (just one cluster). The cost weight matrix $cPLVQ$ is a viable alternative to oversampling and shows comparable performance. It can steer the training with prior knowledge and avoids creating synthetic samples, which makes it cheaper to run.
\subsection{Knowledge extraction from \texorpdfstring{$LVQ^{A}$}{LVQA} models}\label{ss:ALVQ interpretable}

\subsubsection{Visualisation of decision boundaries} 
\begin{figure}[ht]
\centering
\includegraphics[width=0.62\textwidth, height=0.52\textwidth]{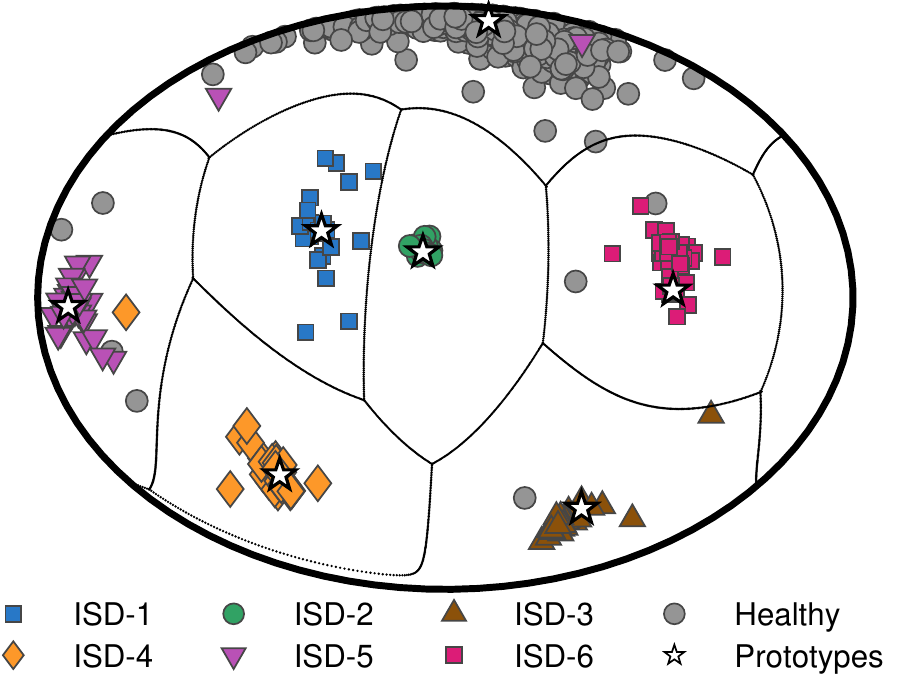}
\caption{Mollweide projection of the decision boundaries, prototypes and samples induced by one of the cluster models $PLVQ^{A_3^{10}}_{\#100_{4}}$ averaged over 32 individual models in fold 1.} 
\label{fig:SoD}
\end{figure}

The angle LVQ variants $LVQ^{A}$ (both probabilistic and deterministic) with $\Lambda$ of rank 3 can be used to visualize the decision boundaries between the conditions, the positions of the prototype of each class, and the subjects on a sphere. The average model (explained in section \ref{sec: Average}) provides the advantages of an ensemble while preserving interpretability. The rank 3 model is not complex enough for the GCMS classification problem resulting in trade-off in performance and several local optima, that we clustered before averaging. For figure \ref{fig:SoD} we selected a model from $PLVQ^{A_3^{10}}_{\#\eta_{\mathrm{\upsilon}}}$ 
cluster $4$ which averaged over 32 constituent models of fold 1. We visualize the sphere in two dimensions using the Mollweide projection\footnote{Matlab code available at \url{https://github.com/drSreejitaGhosh/classificationSphereMollweide}}. 
The reduction of the hypersphere of 496 dimensions to 3 dimensions for visualisation purpose slightly compromised with the sensitivity and class-wise accuracies. However, this illustration provides an effective visual explanation for collaborators of how $PLVQ^{A_3^{10}}_{\#100_{4}}$ performs classification on the hypersphere and highlights the position of subjects which lie close to decision boundaries which may be challenging to accurately classify. 

\subsubsection{Biomarker extraction}
 
\begin{figure}[ht]
\centering
\includegraphics[width=\textwidth,trim={2.1cm 0cm 2.1cm 0cm},clip]{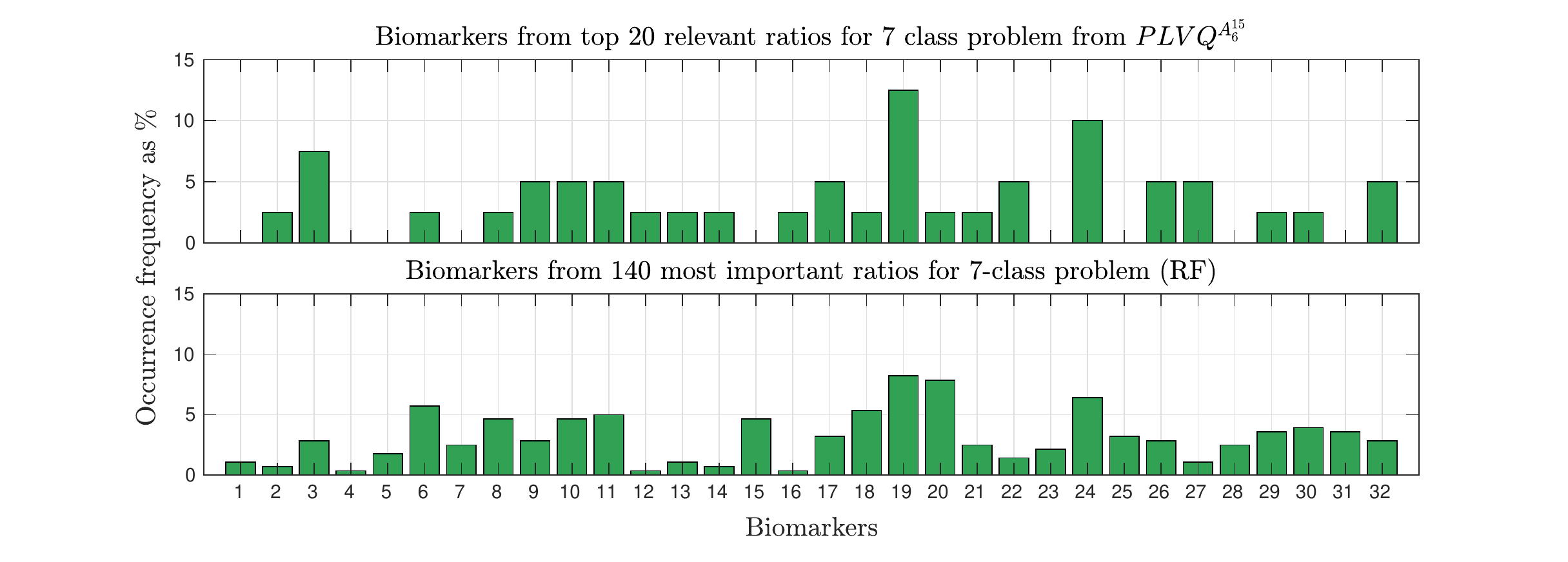}
\caption{Biomarker relevance for inborn disorders of steroidogenesis as occurrence frequency extracted from top 20 and 140 most relevant ratios with $PLVQ^A$ and RF respectively. Numbers 1-32 indicate the 32 metabolites extracted from urine for investigation.
}
\label{fig: pALVQ6D_rel}
\end{figure}
 
Matrix LVQ models enable the extraction of feature relevance information from the given dataset. One can obtain similar general relevance information from the RF through a comparatively laborious process, using the model-agnostic MDA strategy explained in section \ref{ss: RF}. Figure \ref{fig: pALVQ6D_rel} shows the feature relevance in form of occurrence frequency of each metabolite measured in urine for the classification  of inborn disorders of steroidogenesis, extracted from the largest 20 ratios on diagonal of $\Lambda$ in comparison with those extracted from 140 most important ratios of the RF (to cover 20 ratios $\times$ 7 classes). We reassuringly find that the feature relevance and importance profiles obtained from $PLVQ^A$ ad RF look very similar, which one would expect for two classifiers with very similar performance. 

While it is theoretically possible to extract condition specific information from RF, it is not as intuitive or straightforward a process as it is to extract similar information from all $LVQ^A$ variants. Class-specific biomarker information is often vital for clinician understanding and interpretation, but often difficult to obtain from classification models. Being prototype based classifiers, we can also extract how the measured biomarkers varied in magnitude to each other for each of the disease conditions.
The information about the prototype profiles can be obtained by both the global and local versions of $LVQ^{A}$. However, the local version also provide information about which biomarker ratios (i.e. which feature vectors) are relevant for distinguishing a particular condition from the rest, therefore providing clinicians with a unique ``fingerprint'' pattern of biomarkers most specific for each condition. The feature relevance information for each condition is easily extracted from the diagonal of the matrix LVQ variants with classwise local metric tensors $\Lambda^L$. However, both probabilistic and deterministic LVQ variants can provide the domain expert with far superior interpretability in this regard. Using the average models we can perform a descriptive analysis of the classification terms (explained in sec \ref{sec:descriptiveGCMS}) to investigate the statistical relevance of features 
for the classification of a particular condition or any individual sample. Extraction of such knowledge can give clinicians insight into the disease mechanisms for each of these rare diseases and generate trust in the decision made by these classifiers by transparency, and demonstration that the extracted knowledge makes clinical and biological sense.

\subsubsection{Descriptive analysis of Probabilistic LVQ decisions}
\label{sec:descriptiveGCMS}
Any matrix LVQ model allows the analysis of its decision making based on the classification terms, which essentially serve similar purpose to that by the classification activation map (CAM) as explained in \citep{XAItowardsMedicine}. For the proposed variant $PLVQ^A$ (Eq.~\eqref{eq:E_pALVQ}) for example a classification term is the product of the sample vector dimension $\vec{x}_{i,d1}$, 
the relevance matrix element $\Lambda_{d1,d2}$, and a prototype dimension $\vec{w}_{d2}^c$, together with additional factors or transformations dependent on the dissimilarity measure. A sample $\vec{x}_{i}$ is classified as class $c$ if the sum of classification terms ($T^{ic}$) including prototype $\vec{w}^c$ is larger than for any other prototype:
\begin{align}
\notag \hat p(c|\vec{x}_i) &= \frac{g_\Theta\left(\sum_{F1=1,F2=1}^DT_{F1,F2}^{ic}\right) 
}{\sum_j^C g_\Theta\left(\sum_{F1=1,F2=1}^DT_{F1,F2}^{ic}\right)} \quad 
\text{with }T_{F1,F2}^{ic} =\frac{x_{i,F1}\Lambda_{F1,F2}w^c_{F2}}{
\lVert\vec{x}_i\rVert_\Lambda\lVert\vec{w}^c\rVert_\Lambda} \enspace .
\end{align}
While the generalization performance is demonstrated in the previous section we show here the decision making statistics over the full data set, and hence a descriptive analysis. Therefore, we build one model from the 5 fold $cPLVQ^{A_6^{15}}_{\#100_1}$ models, using the geodesic averaging strategy explained in section \ref{sec: Average}, that represents the average statistics of the trained decision making process across all folds. Extracting the feature-wise relevances from the diagonal of $\overline{\Lambda}$ and sorting in descending order reveals that 394 metabolite ratios (out of 496) already contain over 95\% of the total relevance and equivalent accuracy. We remove the unimportant dimensions resulting in a reduced model for the following analysis, that only misclassifies 15 out of 1007 samples in total. Among the latter are 10 healthy controls, 1 ISD-1 patient missed as healthy, 2 ISD-3
patients missed as ISD-5 and 2 ISD-5 patients misclassified as ISD-4. The $\hat p(c|\vec{x}_i)$ provides the probability for  sample $i$ to belong to  class $c$ and we can see that the second most likely class is often the correct one. However, we are interested in the ratios and metabolites and how much (on average) they contribute to the decision. 

\begin{figure}
\includegraphics[width=\textwidth]{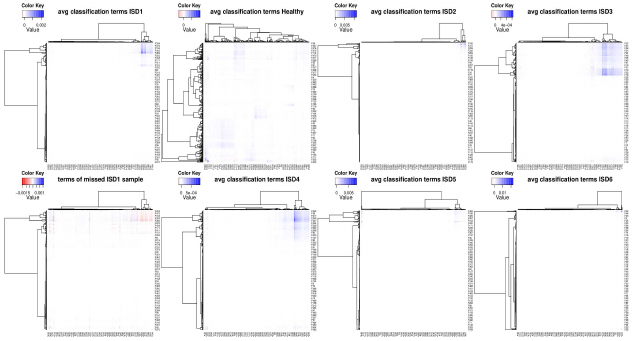}
\caption{Biclusters of classification terms $T^{ic}$ for the 394 metabolite ratio features $F$ averaged for all samples of $\vec{x}_i$ with $y_i=c$ for every condition $c$. 
The terms of the missed ISD-1 
sample (lower left panel) are sorted according to the condition bicluster (top left).}
\label{fig:biclusters}
\end{figure}

The matrix of classification terms $T^{ic}$ contain positive or negative entries indicating the correlation of $\vec{x}_i$ with the prototype $\vec{w}^c$ induced by the metric tensor $\Lambda$. Since the classification decision is based on the biggest sum over all  $\underset{c}{\arg\max}(\sum_{F1,F2}T_{F1,F2}^{ic})$ the terms can be sorted. Figure \ref{fig:biclusters} shows biclusters of classification terms $T^{ic}$ averaged for all samples of $\vec{x}_i$ with $y_i=c$ for every condition $c$. The rows and columns are clustered using the agglomerative Ward2 cluster algorithm \citep{Ward1963,ward2}, grouping similar entries simultaneously in rows and columns. Note that most of the classification decisions for each condition are only based on comparably few metabolite ratios as many terms are close to zero. In contract to Healthy Controls the disease samples show mostly a clear important block of pairwise ratios dominating the decision. The misclassified ISD-1 sample is shown in the lower left panel with the sorting adopted from its condition's average classification terms showing clearly that important ratios differ significantly from the respective prototype.

\begin{figure}
\centering
\includegraphics[width=0.46\textwidth]{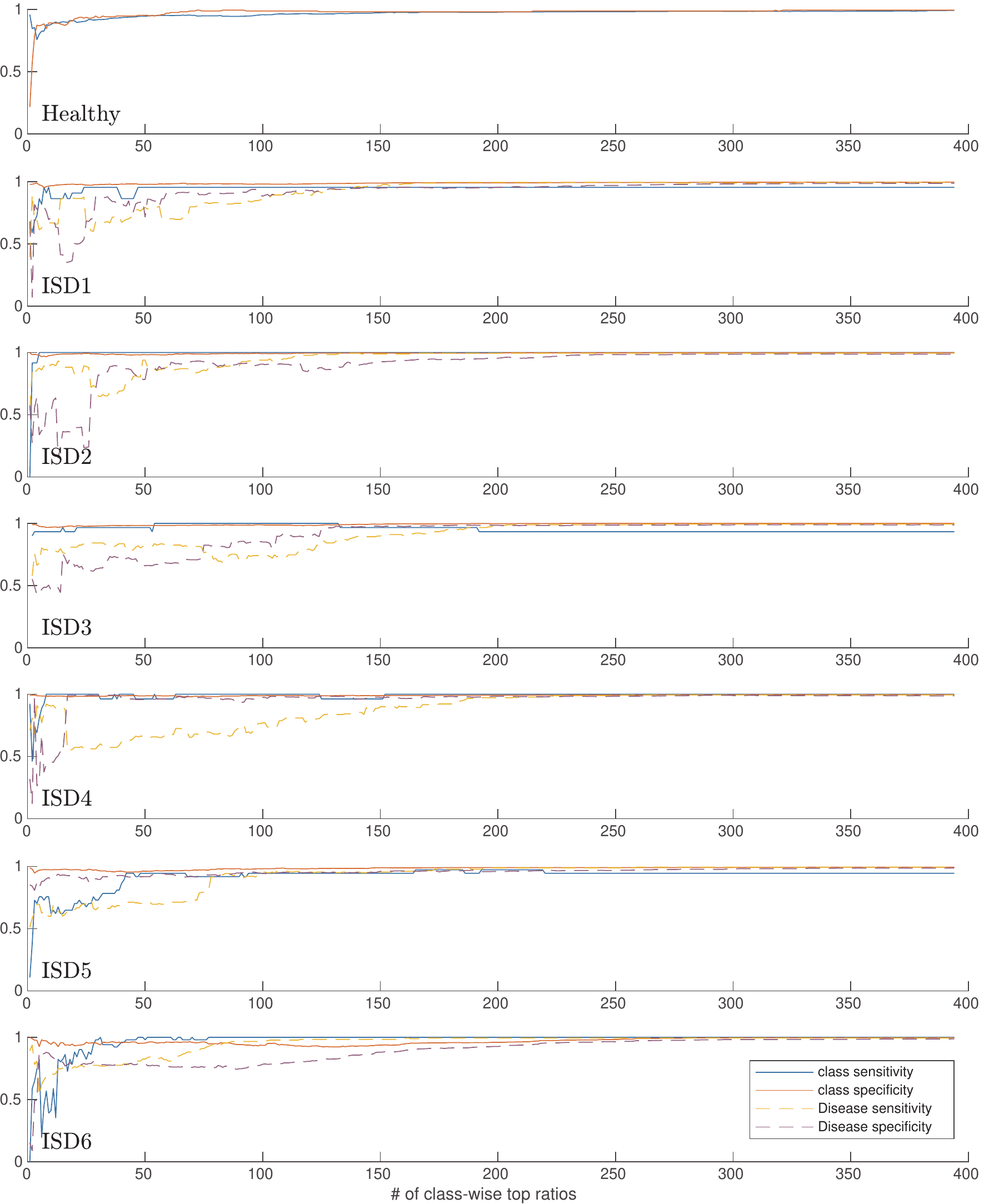}
\includegraphics[width=0.53\textwidth]{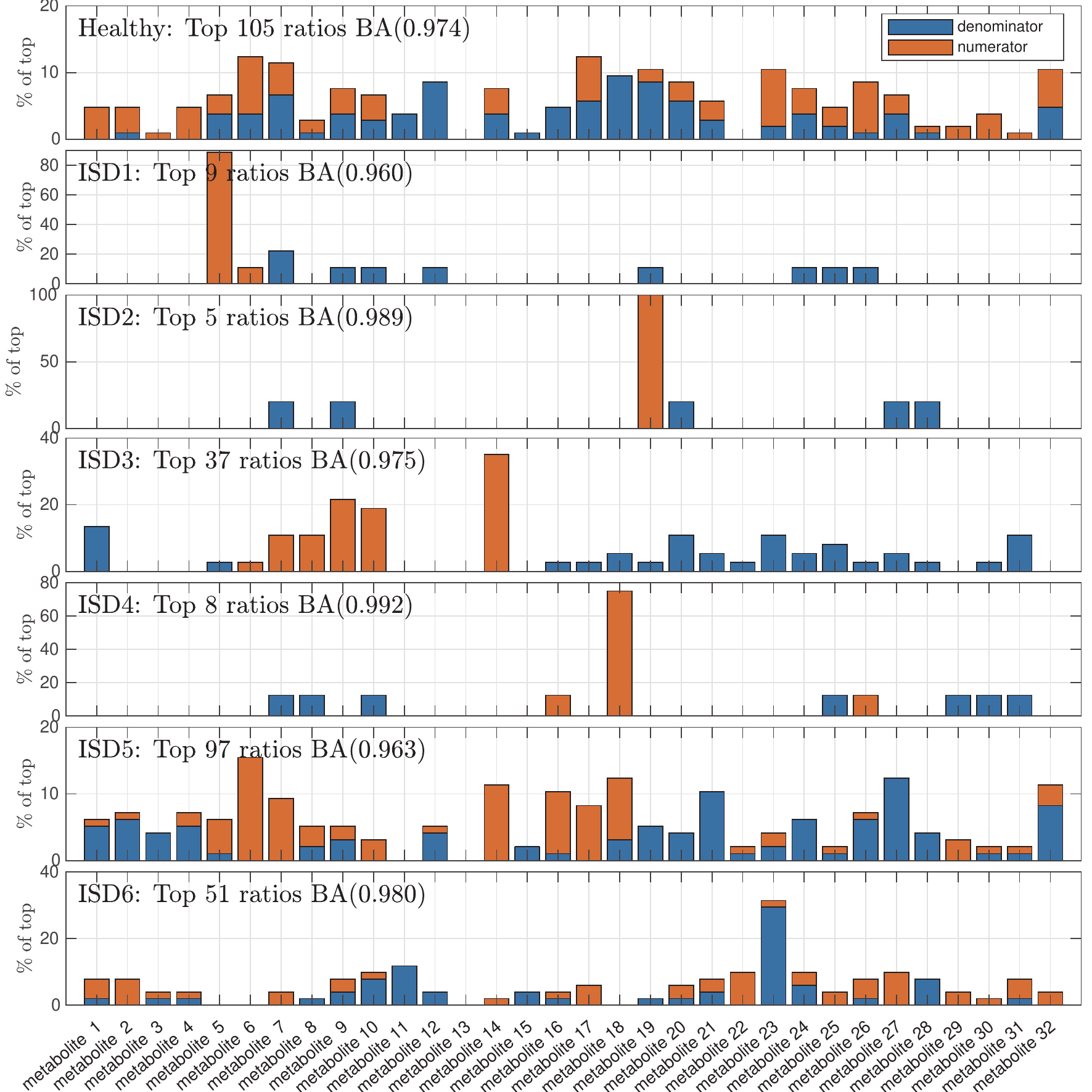}
\caption{Descriptive analysis of ratio contribution (left) and metabolite frequency among top ratios (right) for the classification of the inborn disorders of steroidogenesis. 
}
\label{fig:classTermStats}
\end{figure}

Figure \ref{fig:classTermStats} depicts the performance change dependent on the number of top ratios used in the model and the frequency of metabolites for each class. The left panel shows the respective class specific sensitivity and specificity (as well as overall sensitivity and specificity in terms of healthy vs disease) achieved by the model being reduced to the top $x$ ratios extracted from the sorted average classification terms $T^{ic}$ per class $y_i=c$. And the right panel depicts the frequency of metabolites among the top ratios and their appearance in nominator or denominator, that indicates if it is over- or under-produced in the respective class. We additionally report the number of top ratios per class that achieve $98\%$ of the balanced accuracy (BA, arithmetic mean of the sum of sensitivity and specificity of the class versus the other classes combined) within each panel.
Notably, the classification decision for patients of some of the conditions, namely ISD-1, ISD-2, and ISD-4, is very accurate with over 0.95 BA and on average based on fewer than 10 metabolite ratios (9, 5 and 8 respectively). Within those top ratios an overwhelming majority of over $80\%$ contain one specific metabolite in the numerator indicating an excess compared to healthy controls, which constitutes an interesting biomarker for each of these conditions. The other conditions ISD-3, ISD-6, ISD-5, and Healthy are more heterogeneous and an increasing number of ratios and consequently metabolites are necessary to distinguish them. In summary, the classifier exhibits excellent performance and readily provides ample insight into the contribution of each feature for the decision of each individual samples, as well as the feature statistics over all samples per class. This transparency constitutes an important characteristic for the purpose of medical education, potentially biomarker discovery and to gain trust for computer aided diagnosis with machine learning within the medical community.

\section{Classification and analysis of the 
UCI heart disease dataset}
\label{sec: Exp HD5Cls}
In this section we compare the most promising classifiers as demonstrated on the real-world urine GCMS data set on the publicly available UCI heart disease (HD) data. As mentioned earlier, unlike many of the the existing classifiers the angle LVQ variants can learn even in the presence of missing values, thus obviating the need for imputation or case deletion, or the use of a pseudo value, such as -9 as suggested for this data set \cite{DavidWAha1988}. We follow the recommendation with the pseudo value for the RF training. Many previous publications report performances simplifying the five classes, Healthy and HD1-HD4, to the binary disease versus healthy problem, due to the difficulty. Since we were interested in investigating this dataset as a multi-class problem and the smallest minority class contained only 13 subjects, we use 5-fold stratified training-validation split for cross validation. We use z-score transformation in each fold using the mean and standard deviation of the corresponding training set. As before, we compare two strategies for handling class imbalance: 
(a) SMOTE as described in \cite{chawla2002smote} (or geodesic SMOTE$^g$ for the angle LVQ variants, see section \ref{sec:SMOTEg}), and 
(b) assigning user-defined variable costs of misclassification (see section \ref{sss:PenaltyMat}). For RF we used only (a). We ensured that all the minority classes in the training set were oversampled to contain the same number of samples as the majority class (Healthy) and based on line search we chose $k=3$ nearest neighbours for both SMOTE and SMOTE$^g$. For option (b) we set the cost-weight entries to 1, since we do not have expert information about the severity of certain misclassifications and hence only handle the imbalance. We varied the number of trees for the RF$_t$ from $t=$50 to 200 and the hyper-parameters $\beta$ and $\theta$ for angle LVQ$^{A_M^{\{\beta, \theta\}}}$ and the probabilistic version PLVQ$^{A_M^\theta}$ is set to 1 and 2 respectively after grid-search. To be comparable to the RF, which is an ensemble of decision trees, we trained 100 $LVQ^A$ models in each fold and additionally reported the performance of their majority vote ensemble, abbreviated by (cP)LVQ$^*_{e100}$.
We trained the 100 LVQ models with a full metric tensor rank of $M=13$ in each fold and also built a one cluster geodesic average model abbreviated by (cP)LVQ$^*_{\#100_1}$ for further analysis.
Since the majority of the probabilistic model tensors exhibited a rank of 12 after training, and we need equal rank to build the average model, we limited the rank to 12 for all of them. 
\renewcommand{\arraystretch}{0.92}
\begin{table}[t]
\caption{\footnotesize{UCI HD data: mean performance (std), in terms of Sensitivity (Healthy versus H1-4 combined), macro averaged (MAA) and class-wise accuracies, of RF$_t$ with $t$ trees and angle LVQ models $(cP)LVQ^{A_M^{\beta,\theta}}$ with rank $M$, and $c$ indicating the use of cost weights, $P$ probabilistic cost function \eqref{eq:pALVQ3} and the subscripts $e\eta$ and $\#\eta_{\mathrm{\upsilon}}$ marking the ensemble results with majority vote and average across $\eta$ models and $\upsilon$ clusters. 
}}
\label{tab:hd_5cls13dim}
\begin{tabularx}{\linewidth}{
@{\extracolsep{\fill}} 
@{}>{\raggedright\tiny}p{0.11\textwidth}
@{}*{6}{@{}>{\raggedleft\tiny}p{0.1\textwidth}@{}} >{\tiny\arraybackslash}r@{} }
\toprule
\scriptsize{Method} & \makecell[c]{\scriptsize{Sens}} & \makecell[c]{\scriptsize{MAA}}& \makecell[c]{\scriptsize{Healthy}} &  \makecell[c]{\scriptsize{HD1}} & \makecell[c]{\scriptsize{HD2}} & \makecell[c]{\scriptsize{HD3}} & \makecell[c]{\scriptsize{HD4}}  \\
\toprule
$RF_{100}$ & 73.48 (0.03) & 31.82 (0.04) & \textbf{86.37} (0.03) & 15.64 (0.07) & 19.36 (0.10) & 25.71 (0.11) & 12.0 (0.12) \\
$RF_{150}$ & 74.45 (0.02) & 33.60 (0.03) & 86.25 (0.03) & 15.27 (0.07) & 23.14 (0.10) & 28.00 (0.10) & 15.33 (0.12) \\
$RF_{200}$ & 74.63 (0.02) & 31.68 (0.04) & 86.12 (0.02) & 14.18 (0.07) & 19.36 (0.09) & 27.43 (0.12) & 11.33 (0.10) \\
\midrule
$PLVQ^{A_{12}^2}
$ & 89.08 (0.05) & 50.23 (0.04) & 67.52 (0.03) & 26.67 (0.09) &	32.10 (0.03) & 51.34 (0.15) & 73.50 (0.13) 	\\
$cPLVQ^{A_{12}^2}
$ &	88.91 (0.05) & 51.34 (0.06) & 71.09 (0.03) & 25.71 (0.09) &	29.05 (0.07) & 52.20 (0.15)	& 78.67 (0.17)	\\
$cLVQ^{A^1_{13}}
$ &	83.26 (0.07) & 30.94 (0.07) & 63.97 (0.15) & 21.29 (0.12) & 21.47 (0.14) & 15.49 (0.14) & 32.50 (0.22)	\\
\midrule
$PLVQ^{A_{12}^2}_{e100}$  & 88.33 (0.08) & \textbf{58.55} (0.08) & 74.41 (0.03) & 32.73 (0.14) & 30.36 (0.11) & \textbf{68.57} (0.27) & \textbf{86.67} (0.18)\\ 
$cPLVQ^{A_{12}^2}_{e100}$   & 87.59 (0.08) & 57.74 (0.04) & 75.04 (0.04) & 30.91 (0.19) & \textbf{33.21} (0.07) & 62.86 (0.26) & \textbf{86.67} (0.18)\\
$cLVQ^{A_{13}^1}_{e100}$    & 82.69 (0.09) & 33.15 (0.10) & 82.35 (0.06) & 16.36 (0.12) & 21.79 (0.18) & 8.57 (0.08) & 36.67 (0.41)\\
$PLVQ^{A_{12}^2}_{\#100_1}$ & 91.24 (0.07) & 50.07 (0.08) & 61.02 (0.08) & 29.09 (0.15) & 25.00 (0.12) & 48.57 (0.28) & \textbf{86.67} (0.18) \\
$cPLVQ^{A_{12}^2}_{\#100_1}$& \textbf{92.75} (0.04)	& 50.95 (0.10)	& 65.34 (0.10) & \textbf{34.55} (0.17) & 28.21 (0.11) & 40.00 (0.27) & \textbf{86.67} (0.18) \\
$cLVQ^{A_{13}^1}_{\#100_1}$ & 84.78 (0.09) & 33.62 (0.12) & 76.23 (0.05) & 18.18 (0.14) & 21.79 (0.18) & 8.57 (0.08) & 43.33 (0.43)\\
\bottomrule
\end{tabularx}
\end{table}
\renewcommand{\arraystretch}{1}

Table \ref{tab:hd_5cls13dim} summarizes the mean and standard deviations of the method performances measured in Sensitivity (Healthy versus all diseases combined), macro averaged and class-wise accuracies (Healthy and HD1-4). It can be seen immediately that the five class problem is challenging with a macro averaged accuracy not much more than 30\% achieved by the RF.
The two class problem of all heart disease versus healthy is easier, showing a sensitivity and specificity (class-wise accuracy of the healthy class) of $\approx 74\%$ and $\approx 86\%$. 
Interestingly this data set shows a clear difference between the two different cost functions for the angle LVQ. 
Ensembles of the version inspired by generalized LVQ (updating only the closest correct and wrong prototypes) exhibit only slightly better performance than RF, while the probabilistic version improves the sensitivity and macro averaged accuracy by more than 10\%.
This effect might be caused by the influence of all classes in every update due to the use of the parameterized softmax.
Both strategies to handle the imbalance, namely cost weighted and geodesic SMOTE, demonstrate similar performance. 

\begin{figure}[t]
\centering
\includegraphics[width=0.5\textwidth]{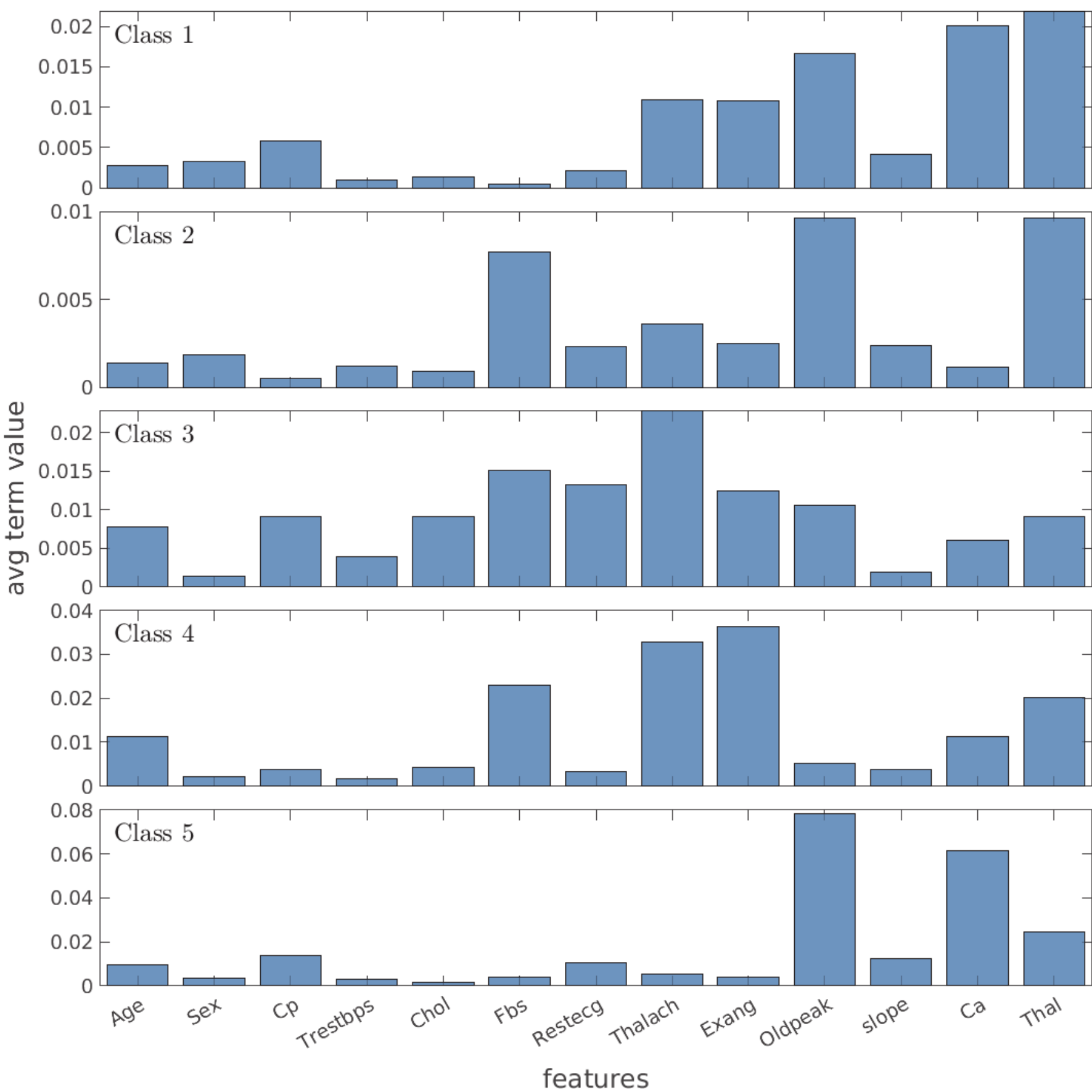}
\includegraphics[width=0.49\textwidth]{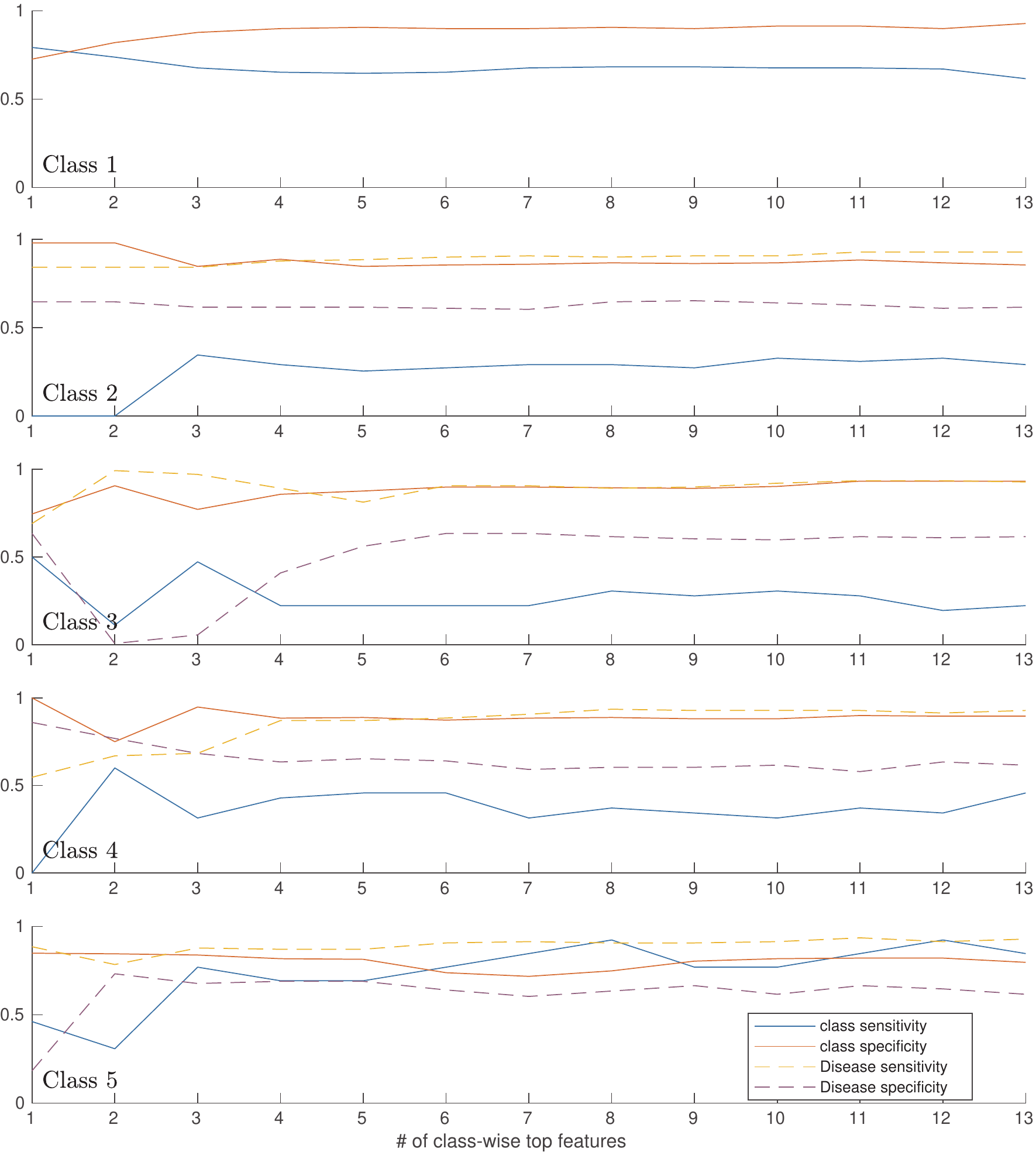}
\caption{UCI HD: Class-wise feature importance for the classification of each class extracted from the 
terms (left) and the 
sensitivity and specificity considering healthy (Class 1) versus disease (Class 2-5) as well as the respective class versus all others combined (right). 
}
\label{fig:CTvsF1_UCI}
\end{figure}

The cost weighting has several benefits, such as, it is faster, operates on fewer samples and allows the user to indicate priorities for the classification and can therefore be recommended. We observe that increase of accuracy for one class is usually accompanied with a decrease of accuracy of another class. RF exhibits the best accuracy for the healthy controls at the expense of accuracy for the diseases. However, performance with respect to the class-wise accuracies of the disease classes are clearly surpassed by the angle LVQ variants, which in turn demonstrates a high sensitivity for the two-class problem. Similar to the previous section we can analyze the statistics of the contribution of the features to the decision of the trained classifiers for each class using the classification terms of the average model across all folds of the $cPLVQ^{A^2_{12}}_{\#100_{1}}$ experiment.
Figure \ref{fig:CTvsF1_UCI} illustrates the performance of each class versus all others combined, as well as Healthy versus disease, when including only the top features for the samples of the respective class in the right panel. 
Since we do not consider ratios we can show the classification term contribution of the features for each class directly as shown in the left panel.
Especially the Healthy (Class 1) and HD1 (Class 2) exhibit a very similar pattern of important features, which explains the difficulty to distinguish them. Figure \ref{fig:CTvsF1_UCI} shows that the 'Oldpeak', referring to ST depression in the ECG signal, induced by exercise relative to rest, is an important feature to identify Healthy, HD1 and HD5 from the rest. Contrarily, 'Thalach' which refers to the maximum heart rate achieved, is important to discriminate HD3 and HD4 from the rest. While RF can find the overall feature importance and there is a good overlap with the findings from prototype based methods as illustrated in \cite{ghosh2020visualisation}, the classification terms from LVQ models help in extraction of class-specific feature relevance as seen in figure \ref{fig:CTvsF1_UCI}.

\section{Conclusion and Future work}
Real-world data from certain domains, such as healthcare, often exhibit a multitude of complications, such as heterogeneous measurements, imbalanced classes, limited size of available data for training, and missing values which are often systematically missing. 
Furthermore, the lack of explainability of many machine learning techniques has fueled an ongoing debate for the responsible use of such methods in critical domains. 
All of these aspects typically hinder the straightforward application of Machine Learning models. 
In this contribution we present, analyze and demonstrate strategies for performing classification in the presence of the aforementioned challenges. 
Our framework delivers robust and comparable or superior performance, while maintaining the interpretability, which is considered essential in an anthropocentric application of machine learning, as desired in healthcare.
This paper extends our work of \cite{ghosh2017comparison, ghosh2020visualisation} with three major contributions:
(1) A comprehensive framework of \emph{intrinsically explainable}, angle-dissimilarity based nearest prototype classifiers of varying complexity (Angle Learning Vector Quantization ($LVQ^A$)), that demonstrates excellent performance even when facing heterogeneous measurements, imbalanced classes, limited data for training and systematically missing values. Additionally to training with and output of crisp labels, the framework includes a probabilistic cost function that can be trained with a combination of crisp (sure) as well as uncertain labels (for example when experts disagree), while providing a probabilistic class output, reflecting the uncertainty of the classification and hence providing additional insight into the model.
(2) The introduction of strategies to harness the power of ensembling of models, by computation of the geodesic average LVQ model and a clustering strategy in case of the presence of multiple local optima, which serve as more \emph{transparent alternatives to a traditional ensemble} and extends to other members of the adaptive metric family, such as the Euclidean LVQ variants and Large Margin Nearest Neighbor \cite{weinberger2009distance, ramanan2010local}.
(3) Furthermore, we demonstrate and compare the performance of the framework and strategies to alternative techniques applicable in situations involving the multitude of complications considered here. 
We provide our code and a detailed analysis and demonstrate the transparency of our framework and how knowledge is extracted from real-world medical datasets. 
Especially the findings for the inborn disorders of steroidogenesis are of significant interest for the healthcare domain and additional details of the analysis of the metabolites and their importance for every disease type will be presented in a separate paper written for the medical community.

In the presence of heterogeneous measurements and systematic missingness the choice of dissimilarity measure has a significant influence on the performance of the LVQ classifiers. A cosine-based adaptive dissimilarity measure appears more robust than the parameterized Euclidean distance in the presence of such complications. . 
We compared a state-of-the-art model agnostic approach of handling missingness via multiple imputation and demonstrated that it can have adverse impact on distance-based classification. 
Angle LVQ when applied to data without imputation, displays similar or superior performance to applicable alternative techniques. However, the $LVQ^A$ variants allow detailed insight into the influence of features on the classification of every sample, every class, the data as a whole and, in special cases, direct visualization of the data and decision boundaries.
Random Forest (RF), which is an ensemble of decision trees, generally shows quite good performance despite the difficulties of the data. 
This fact inspired the development of a strategy to harness the power of ensembling, which is typically accompanied with a loss of model transparency, while preserving interpretability. 

Through extensive examination of our real-life medical data we demonstrate that an average model as replacement of an ensemble of 100 LVQ classifiers shows a comparable or superior performance to RF with 100 DTs, with the additional benefit of increased interpretability and eased knowledge extraction. Especially for the inborn disorders of steroidogenesis our framework provided significant insights into the classification of
 six rare disorders by comprehensive data-driven analysis of 496 pair-wise ratios of 32 urinary biomarkers. The detailed medical findings will be presented in a forthcoming paper written for the medical community. Additionally we presented the competitive performance of our models on a publicly available real-world dataset and briefly illustrate the type of knowledge that could potentially be extracted from it, 
in section \ref{sec: Exp HD5Cls} for reproducibility and verification. Even though we have shown an application in medical domain, these novel strategies  can be applied to data from any other domain, and are particularly relevant where there is missing data of any type. In the current era when the demand for algorithmic transparency, interpretability, and explainability are rising alongside the demand for higher performance, from industry, judiciary, medicine, and society in general, the presented classifiers have shown the potential to efficiently address both these needs.

\acks{We thank (1) the Center for Information Technology of the University of
Groningen for their support and for providing access to the Peregrine High
Performance Computing Cluster, (2) the Rosalind Franklin fellowship, co-
funded by the European Union’s Seventh Framework Programme for research,
technological development and demonstration under grant agreement \textbf{600211},
and H2020-MSCA-IF-2014, project ID \textbf{659104}, (3) Dr. Michael Wilkinson and Dr. Mohammad Mohammadi for proof-reading  this paper at its various stages of development, (4) the LWP team at University of Groningen who made it possible to work from home during the pandemic, and (5) former BSc. student from Short programming project, Ethan Waterink, for his contributions in improving the Mollweide projections.
}
\bibliography{journal1}

\begin{thebibliography}{85}
\providecommand{\natexlab}[1]{#1}
\providecommand{\url}[1]{\texttt{#1}}
\expandafter\ifx\csname urlstyle\endcsname\relax
  \providecommand{\doi}[1]{doi: #1}\else
  \providecommand{\doi}{doi: \begingroup \urlstyle{rm}\Url}\fi

\bibitem[Afsari(2011)]{Afsari2011}
Bijan Afsari.
\newblock Riemannian {$L_p$} center of mass: existence, uniqueness, and
  convexity.
\newblock \emph{Proceedings of the American Mathematical Society}, 139\penalty0
  (2):\penalty0 655--673, 2011.
\newblock ISSN 00029939, 10886826.
\newblock URL \url{http://www.jstor.org/stable/41059320}.

\bibitem[Alfeld et~al.(1996)Alfeld, Neamtu, and
  Schumaker]{Alfeld1996BernsteinBzierPO}
P.~Alfeld, M.~Neamtu, and L.~Schumaker.
\newblock Bernstein-{B{\'e}zier} polynomials on spheres and sphere-like
  surfaces.
\newblock \emph{Comput. Aided Geom. Des.}, 13:\penalty0 333--349, 1996.

\bibitem[Alpaydin(2020)]{parsons2005introduction}
Ethem Alpaydin.
\newblock \emph{Introduction to machine learning}.
\newblock MIT press, 2020.

\bibitem[Amari(2016)]{Amari}
Shun-ichi Amari.
\newblock \emph{Information Geometry and Its Applications}.
\newblock Springer Publishing Company, Incorporated, 1st edition, 2016.
\newblock ISBN 4431559779.

\bibitem[Ando et~al.(2004)Ando, Li, and Mathias]{ANDO2004305}
T.~Ando, Chi-Kwong Li, and Roy Mathias.
\newblock Geometric means.
\newblock \emph{Linear Algebra and its Applications}, 385:\penalty0 305 -- 334,
  2004.
\newblock ISSN 0024-3795.
\newblock \doi{https://doi.org/10.1016/j.laa.2003.11.019}.
\newblock URL
  \url{http://www.sciencedirect.com/science/article/pii/S0024379503008693}.
\newblock Special Issue in honor of Peter Lancaster.

\bibitem[Arlt et~al.(2004)Arlt, Walker, Draper, Ivison, Ride, Hammer, Chalder,
  Borucka-Mankiewicz, Hauffa, Malunowicz, et~al.]{arlt2004congenital}
Wiebke Arlt, Elizabeth~A Walker, Nicole Draper, Hannah~E Ivison, Jon~P Ride,
  Fabian Hammer, Susan~M Chalder, Maria Borucka-Mankiewicz, Berthold~P Hauffa,
  Ewa~M Malunowicz, et~al.
\newblock Congenital adrenal hyperplasia caused by mutant p450 oxidoreductase
  and human androgen synthesis: analytical study.
\newblock \emph{The Lancet}, 363\penalty0 (9427):\penalty0 2128--2135, 2004.

\bibitem[Arnab(2017)]{arnab2017survey}
Raghunath Arnab.
\newblock \emph{Survey sampling theory and applications}.
\newblock Academic Press, 2017.

\bibitem[Arnaudon et~al.(2012)Arnaudon, Dombry, Phan, and
  Yang]{ARNAUDON20121437}
Marc Arnaudon, Clément Dombry, Anthony Phan, and Le~Yang.
\newblock Stochastic algorithms for computing means of probability measures.
\newblock \emph{Stochastic Processes and their Applications}, 122\penalty0
  (4):\penalty0 1437 -- 1455, 2012.
\newblock ISSN 0304-4149.
\newblock \doi{https://doi.org/10.1016/j.spa.2011.12.011}.
\newblock URL
  \url{http://www.sciencedirect.com/science/article/pii/S030441491100319X}.

\bibitem[Arrieta et~al.(2020)Arrieta, D{\'\i}az-Rodr{\'\i}guez, Del~Ser,
  Bennetot, Tabik, Barbado, Garc{\'\i}a, Gil-L{\'o}pez, Molina, Benjamins,
  et~al.]{arrieta2020explainable}
Alejandro~Barredo Arrieta, Natalia D{\'\i}az-Rodr{\'\i}guez, Javier Del~Ser,
  Adrien Bennetot, Siham Tabik, Alberto Barbado, Salvador Garc{\'\i}a, Sergio
  Gil-L{\'o}pez, Daniel Molina, Richard Benjamins, et~al.
\newblock Explainable artificial intelligence (xai): Concepts, taxonomies,
  opportunities and challenges toward responsible ai.
\newblock \emph{Information Fusion}, 58:\penalty0 82--115, 2020.

\bibitem[Azur et~al.(2011)Azur, Stuart, Frangakis, and Leaf]{azur2011multiple}
Melissa~J Azur, Elizabeth~A Stuart, Constantine Frangakis, and Philip~J Leaf.
\newblock Multiple imputation by chained equations: what is it and how does it
  work?
\newblock \emph{International journal of methods in psychiatric research},
  20\penalty0 (1):\penalty0 40--49, 2011.

\bibitem[Backhaus and Seiffert(2014)]{backhaus2014classification}
Andreas Backhaus and Udo Seiffert.
\newblock Classification in high-dimensional spectral data: Accuracy vs.
  interpretability vs. model size.
\newblock \emph{Neurocomputing}, 131:\penalty0 15--22, 2014.

\bibitem[Baranowski et~al.(2018)Baranowski, Arlt, and
  Idkowiak]{baranowski2018monogenic}
Elizabeth~S Baranowski, Wiebke Arlt, and Jan Idkowiak.
\newblock Monogenic disorders of adrenal steroidogenesis.
\newblock \emph{Hormone research in paediatrics}, 89\penalty0 (5):\penalty0
  292--310, 2018.

\bibitem[Bibal and Fr{\'e}nay(2016)]{bibal2016interpretability}
Adrien Bibal and Beno{\^\i}t Fr{\'e}nay.
\newblock Interpretability of machine learning models and representations: an
  introduction.
\newblock In M.Verleysen, editor, \emph{European Symposium on Artificial Neural
  Networks (ESANN)}, pages 17--22. i6doc.com, 2016.

\bibitem[Biehl et~al.(2013)Biehl, Hammer, and Villmann]{biehl2013distance}
Michael Biehl, Barbara Hammer, and Thomas Villmann.
\newblock Distance measures for prototype based classification.
\newblock In \emph{International Workshop on Brain-Inspired Computing}, pages
  100--116. Springer, 2013.

\bibitem[BINI et~al.(2010)BINI, MEINI, and POLONI]{Bini}
DARIO~A. BINI, BEATRICE MEINI, and FEDERICO POLONI.
\newblock An effective matrix geometric mean satisfying the ando-li-mathias
  properties.
\newblock \emph{Mathematics of Computation}, 79\penalty0 (269):\penalty0
  437--452, 2010.
\newblock ISSN 00255718, 10886842.
\newblock URL \url{http://www.jstor.org/stable/40590410}.

\bibitem[Bonnabel and Sepulchre(2010)]{Bonnabel2009}
Silv\'{e}re Bonnabel and Rodolphe Sepulchre.
\newblock Riemannian metric and geometric mean for positive semidefinite
  matrices of fixed rank.
\newblock \emph{SIAM Journal on Matrix Analysis and Applications}, 31\penalty0
  (3):\penalty0 1055--1070, 2010.
\newblock \doi{10.1137/080731347}.
\newblock URL \url{https://doi.org/10.1137/080731347}.

\bibitem[Bonnabel et~al.(2013)Bonnabel, Collard, and
  Sepulchre]{BONNABEL20133202}
Silvère Bonnabel, Anne Collard, and Rodolphe Sepulchre.
\newblock Rank-preserving geometric means of positive semi-definite matrices.
\newblock \emph{Linear Algebra and its Applications}, 438\penalty0
  (8):\penalty0 3202 -- 3216, 2013.
\newblock ISSN 0024-3795.
\newblock \doi{https://doi.org/10.1016/j.laa.2012.12.009}.
\newblock URL
  \url{http://www.sciencedirect.com/science/article/pii/S0024379512008646}.

\bibitem[Breiman(1996)]{breiman1996bagging}
Leo Breiman.
\newblock Bagging predictors.
\newblock \emph{Machine learning}, 24\penalty0 (2):\penalty0 123--140, 1996.

\bibitem[Breiman(2001)]{breiman2001random}
Leo Breiman.
\newblock Random forests.
\newblock \emph{Machine learning}, 45\penalty0 (1):\penalty0 5--32, 2001.

\bibitem[Bunte et~al.(2012)Bunte, Schneider, Hammer, Schleif, Villmann, and
  Biehl]{Bunte2011_LiRaMLVQ}
K.~Bunte, P.~Schneider, B.~Hammer, F.-M. Schleif, T.~Villmann, and M.~Biehl.
\newblock {Limited Rank Matrix Learning -- Discriminative Dimension Reduction
  and Visualization}.
\newblock \emph{Neural Networks}, 26\penalty0 (4):\penalty0 159--173, February
  2012.
\newblock \doi{10.1016/j.neunet.2011.10.001}.
\newblock URL \url{http://dx.doi.org/10.1016/j.neunet.2011.10.001}.

\bibitem[Buss and Fillmore(2001)]{Buss01sphericalaverages}
Samuel~R. Buss and Jay~P. Fillmore.
\newblock Spherical averages and applications to spherical splines and
  interpolation.
\newblock \emph{ACM Transactions on Graphics}, 20:\penalty0 95--126, 2001.

\bibitem[{Carson} et~al.(2017){Carson}, {Mixon}, and {Villar}]{Carson2017}
T.~{Carson}, D.~G. {Mixon}, and S.~{Villar}.
\newblock Manifold optimization for k-means clustering.
\newblock In \emph{2017 International Conference on Sampling Theory and
  Applications (SampTA)}, pages 73--77, 2017.
\newblock \doi{10.1109/SAMPTA.2017.8024388}.

\bibitem[Carvalho et~al.(2019)Carvalho, Pereira, and
  Cardoso]{carvalho2019machine}
Diogo~V Carvalho, Eduardo~M Pereira, and Jaime~S Cardoso.
\newblock Machine learning interpretability: A survey on methods and metrics.
\newblock \emph{Electronics}, 8\penalty0 (8):\penalty0 832, 2019.

\bibitem[Chawla et~al.(2002)Chawla, Bowyer, Hall, and
  Kegelmeyer]{chawla2002smote}
Nitesh~V. Chawla, Kevin~W. Bowyer, Lawrence~O. Hall, and W.~Philip Kegelmeyer.
\newblock Smote: synthetic minority over-sampling technique.
\newblock \emph{Journal of artificial intelligence research}, 16:\penalty0
  321--357, 2002.

\bibitem[Chechik et~al.(2008)Chechik, Heitz, Elidan, Abbeel, and
  Koller]{chechik2008max}
Gal Chechik, Geremy Heitz, Gal Elidan, Pieter Abbeel, and Daphne Koller.
\newblock Max-margin classification of data with absent features.
\newblock \emph{Journal of Machine Learning Research}, 9\penalty0
  (Jan):\penalty0 1--21, 2008.

\bibitem[Clark and Thompson(1984)]{Clark1984}
R.~M. Clark and R.~Thompson.
\newblock Statistical comparison of palaeomagnetic directional records from
  lake sediments.
\newblock \emph{Geophysical Journal International}, 76\penalty0 (2):\penalty0
  337--368, 02 1984.
\newblock ISSN 0956-540X.
\newblock \doi{10.1111/j.1365-246X.1984.tb05050.x}.
\newblock URL \url{https://doi.org/10.1111/j.1365-246X.1984.tb05050.x}.

\bibitem[Dixon(1979)]{Dixon}
John~K. Dixon.
\newblock Pattern recognition with partly missing data.
\newblock \emph{IEEE Transactions on Systems, Man, and Cybernetics}, 9\penalty0
  (10):\penalty0 617--621, 1979.
\newblock \doi{10.1109/TSMC.1979.4310090}.

\bibitem[Doquire and Verleysen(2012)]{doquire2012feature}
Gauthier Doquire and Michel Verleysen.
\newblock Feature selection with missing data using mutual information
  estimators.
\newblock \emph{Neurocomputing}, 90:\penalty0 3--11, 2012.
\newblock ISSN 0925-2312.
\newblock \doi{https://doi.org/10.1016/j.neucom.2012.02.031}.

\bibitem[Eirola et~al.(2013)Eirola, Doquire, Verleysen, and
  Lendasse]{EIROLA2013115}
Emil Eirola, Gauthier Doquire, Michel Verleysen, and Amaury Lendasse.
\newblock Distance estimation in numerical data sets with missing values.
\newblock \emph{Information Sciences}, 240:\penalty0 115--128, 2013.
\newblock ISSN 0020-0255.
\newblock \doi{https://doi.org/10.1016/j.ins.2013.03.043}.
\newblock URL
  \url{https://www.sciencedirect.com/science/article/pii/S0020025513002570}.

\bibitem[Fisher et~al.(2019)Fisher, Rudin, and Dominici]{fisher2019all}
Aaron Fisher, Cynthia Rudin, and Francesca Dominici.
\newblock All models are wrong, but many are useful: Learning a variable's
  importance by studying an entire class of prediction models simultaneously.
\newblock \emph{Journal of Machine Learning Research}, 20\penalty0
  (177):\penalty0 1--81, 2019.

\bibitem[Fletcher et~al.(2004)Fletcher, Lu, Pizer, and Joshi]{Fletcher2004}
P.~T. Fletcher, C.~Lu, S.~M. Pizer, and S.~Joshi.
\newblock Principal geodesic analysis for the study of nonlinear statistics of
  shape.
\newblock \emph{IEEE Trans. on Medical Imaging}, 23\penalty0 (8):\penalty0
  995--1005, 2004.
\newblock ISSN 0278-0062.
\newblock \doi{10.1109/TMI.2004.831793}.

\bibitem[Garc{\'i}a-Laencina et~al.(2010)Garc{\'i}a-Laencina, Sancho-G{\'o}mez,
  and Figueiras-Vidal]{Garcia-Laencina2010}
Pedro~J. Garc{\'i}a-Laencina, Jos{\'e}-Luis Sancho-G{\'o}mez, and An{\'i}bal~R.
  Figueiras-Vidal.
\newblock Pattern classification with missing data: a review.
\newblock \emph{Neural Computing and Applications}, 19\penalty0 (2):\penalty0
  263--282, Mar 2010.
\newblock ISSN 1433-3058.
\newblock \doi{10.1007/s00521-009-0295-6}.
\newblock URL \url{https://doi.org/10.1007/s00521-009-0295-6}.

\bibitem[Ghosh(2021)]{sghoshPhDthesis2021}
Sreejita Ghosh.
\newblock \emph{Intrinsically Interpretable Machine Learning In Computer Aided
  Diagnosis}.
\newblock PhD thesis, University of Groningen, 2021.

\bibitem[Ghosh et~al.(2017)Ghosh, Baranowski, van Veen, de~Vries, Biehl, Arlt,
  Tino, and Bunte]{ghosh2017comparison}
Sreejita Ghosh, Elizabeth Baranowski, Rick van Veen, G~de~Vries, Michael Biehl,
  Wiebke Arlt, Peter Tino, and Kerstin Bunte.
\newblock Comparison of strategies to learn from imbalanced classes for
  computer aided diagnosis of inborn steroidogenic disorders.
\newblock In M.~Verleysen, editor, \emph{Proc. of the European Symposium on
  Artificial Neural Networks}, pages 199--204. i6doc.com, 2017.

\bibitem[Ghosh et~al.(2020)Ghosh, Tino, and Bunte]{ghosh2020visualisation}
Sreejita Ghosh, Peter Tino, and Kerstin Bunte.
\newblock Visualisation and knowledge discovery from interpretable models.
\newblock In \emph{International Joint Conference on Neural Networks, {IJCNN}
  2020 Glasgow, UK, July 19-24, 2020}. {IEEE}, 2020.
\newblock URL \url{https://doi.org/}.

\bibitem[Hammer and Villmann(2002)]{Hammer20021059}
B.~Hammer and T.~Villmann.
\newblock Generalized relevance learning vector quantization.
\newblock \emph{Neural Networks}, 15\penalty0 (8--9):\penalty0 1059 -- 1068,
  2002.
\newblock ISSN 0893-6080.
\newblock \doi{http://dx.doi.org/10.1016/S0893-6080(02)00079-5}.
\newblock URL
  \url{http://www.sciencedirect.com/science/article/pii/S0893608002000795}.

\bibitem[Hammer et~al.(2005)Hammer, Strickert, and Villmann]{Hammer2005}
B.~Hammer, M.~Strickert, and T.~Villmann.
\newblock On the generalization ability of {GRLVQ} networks.
\newblock \emph{Neural Processing Letters}, 21\penalty0 (2):\penalty0 109--120,
  2005.
\newblock ISSN 1573-773X.
\newblock \doi{10.1007/s11063-004-1547-1}.

\bibitem[Hegde et~al.(2019)Hegde, Shimpi, Panny, Glurich, Christie, and
  Acharya]{HEGDE2019100275}
Harshad Hegde, Neel Shimpi, Aloksagar Panny, Ingrid Glurich, Pamela Christie,
  and Amit Acharya.
\newblock Mice vs ppca: Missing data imputation in healthcare.
\newblock \emph{Informatics in Medicine Unlocked}, 17:\penalty0 100275, 2019.
\newblock ISSN 2352-9148.
\newblock \doi{https://doi.org/10.1016/j.imu.2019.100275}.
\newblock URL
  \url{http://www.sciencedirect.com/science/article/pii/S2352914819302783}.

\bibitem[Holzinger et~al.(2017)Holzinger, Biemann, Pattichis, and
  Kell]{holzinger2017we}
Andreas Holzinger, Chris Biemann, Constantinos~S Pattichis, and Douglas~B Kell.
\newblock What do we need to build explainable ai systems for the medical
  domain?
\newblock \emph{arXiv preprint arXiv:1712.09923}, 2017.

\bibitem[Janosi et~al.(1988)Janosi, Steinbrunn, Pfisterer, and
  Detrano]{DavidWAha1988}
Andras Janosi, William Steinbrunn, Matthias Pfisterer, and Robert Detrano.
\newblock Heart disease data set, {UCI} machine learning repository, 1988.
\newblock URL \url{https://archive.ics.uci.edu/ml/datasets/heart+Disease}.

\bibitem[Jr.(1963)]{Ward1963}
Joe H.~Ward Jr.
\newblock Hierarchical grouping to optimize an objective function.
\newblock \emph{Journal of the American Statistical Association}, 58\penalty0
  (301):\penalty0 236--244, 1963.
\newblock \doi{10.1080/01621459.1963.10500845}.

\bibitem[Karcher(1977)]{Karcher1977}
H.~Karcher.
\newblock Riemannian center of mass and mollifier smoothing.
\newblock \emph{Communications on Pure and Applied Mathematics}, 30\penalty0
  (5):\penalty0 509--541, 1977.
\newblock \doi{10.1002/cpa.3160300502}.
\newblock URL
  \url{https://onlinelibrary.wiley.com/doi/abs/10.1002/cpa.3160300502}.

\bibitem[Kendall(1990)]{Kendall1990}
Wilfrid~S. Kendall.
\newblock {Probability, Convexity, and Harmonic Maps with Small Image I:
  Uniqueness and Fine Existence}.
\newblock \emph{Proceedings of the London Mathematical Society}, s3-61\penalty0
  (2):\penalty0 371--406, 09 1990.
\newblock ISSN 0024-6115.
\newblock \doi{10.1112/plms/s3-61.2.371}.
\newblock URL \url{https://doi.org/10.1112/plms/s3-61.2.371}.

\bibitem[Krakowski et~al.(2007)Krakowski, H\"uper, and
  Manton]{Krakowski07onthe}
Krzysztof~A. Krakowski, Knut H\"uper, and Jonathan~H. Manton.
\newblock On the computation of the karcher mean on spheres and special
  orthogonal groups.
\newblock In \emph{in Proc. Workshop Robot. Math. (RoboMat) ’07}, September
  2007.

\bibitem[Kubat(2017)]{kubat2017introduction}
Miroslav Kubat.
\newblock \emph{An introduction to machine learning}.
\newblock Springer, 2017.

\bibitem[Lall and Sharma(1996)]{lall1996nearest}
Upmanu Lall and Ashish Sharma.
\newblock A nearest neighbor bootstrap for resampling hydrologic time series.
\newblock \emph{Water Resources Research}, 32\penalty0 (3):\penalty0 679--693,
  1996.

\bibitem[Lawson and Lim(2013)]{Lawson2016}
Jimmie Lawson and Yongdo Lim.
\newblock Weighted means and {Karcher} equations of positive operators.
\newblock \emph{Proceedings of the National Academy of Sciences}, 110\penalty0
  (39):\penalty0 15626--15632, 2013.
\newblock ISSN 0027-8424.
\newblock \doi{10.1073/pnas.1313640110}.
\newblock URL \url{https://www.pnas.org/content/110/39/15626}.

\bibitem[Little(1988{\natexlab{a}})]{little1988missing}
Roderick~JA Little.
\newblock Missing-data adjustments in large surveys.
\newblock \emph{Journal of Business \& Economic Statistics}, 6\penalty0
  (3):\penalty0 287--296, 1988{\natexlab{a}}.

\bibitem[Little(1988{\natexlab{b}})]{little1988test}
Roderick~JA Little.
\newblock A test of missing completely at random for multivariate data with
  missing values.
\newblock \emph{Journal of the American statistical Association}, 83\penalty0
  (404):\penalty0 1198--1202, 1988{\natexlab{b}}.

\bibitem[Little and Rubin(2019)]{little2019statistical}
Roderick~JA Little and Donald~B Rubin.
\newblock \emph{Statistical analysis with missing data}, volume 793.
\newblock John Wiley \& Sons, 2019.

\bibitem[Lundberg and Lee(2017)]{lundberg2017unified}
Scott~M Lundberg and Su-In Lee.
\newblock A unified approach to interpreting model predictions.
\newblock In \emph{Advances in neural information processing systems}, pages
  4765--4774, 2017.

\bibitem[Luo et~al.(2019)Luo, Tseng, Cui, Wei, Ten~Haken, and
  El~Naqa]{luo2019balancing}
Yi~Luo, Huan-Hsin Tseng, Sunan Cui, Lise Wei, Randall~K Ten~Haken, and Issam
  El~Naqa.
\newblock Balancing accuracy and interpretability of machine learning
  approaches for radiation treatment outcomes modeling.
\newblock \emph{BJR| Open}, 1\penalty0 (1):\penalty0 20190021, 2019.

\bibitem[Maaten and Hinton(2008)]{maaten2008visualizing}
Laurens van~der Maaten and Geoffrey Hinton.
\newblock Visualizing data using t-sne.
\newblock \emph{Journal of Machine Learning Research}, 9:\penalty0 2579--2605,
  2008.

\bibitem[Marlin(2008)]{Marlin:2008:MDP:1925592}
Benjamin~M. Marlin.
\newblock \emph{Missing Data Problems in Machine Learning}.
\newblock PhD thesis, University of Toronto, Toronto, Ont., Canada, Canada,
  2008.
\newblock AAINR57898.

\bibitem[Marrinan et~al.(2014)Marrinan, Draper, Beveridge, Kirby, and
  Peterson]{Marrinan2014}
Tim Marrinan, Bruce Draper, J.~Ross Beveridge, Michael Kirby, and Chris
  Peterson.
\newblock Finding the subspace mean or median to fit your need.
\newblock In \emph{Proceedings of the 2014 IEEE Conference on Computer Vision
  and Pattern Recognition}, CVPR '14, page 1082–1089, USA, 2014. IEEE
  Computer Society.
\newblock ISBN 9781479951185.
\newblock \doi{10.1109/CVPR.2014.142}.
\newblock URL \url{https://doi.org/10.1109/CVPR.2014.142}.

\bibitem[Mohammadi et~al.(2019)Mohammadi, Petkov, Bunte, Peletier, and
  Schleif]{mohammadi2019globular}
Mohammad Mohammadi, Nicolai Petkov, Kerstin Bunte, Reynier~F Peletier, and F-M
  Schleif.
\newblock Globular cluster detection in the gaia survey.
\newblock \emph{Neurocomputing}, 342:\penalty0 164--171, 2019.

\bibitem[Mujalli et~al.(2016)Mujalli, L{\'o}pez, and Garach]{mujalli2016bayes}
Randa~Oqab Mujalli, Griselda L{\'o}pez, and Laura Garach.
\newblock Bayes classifiers for imbalanced traffic accidents datasets.
\newblock \emph{Accident Analysis \& Prevention}, 88:\penalty0 37--51, 2016.

\bibitem[Murtagh and Legendre(2014)]{ward2}
Fionn Murtagh and Pierre Legendre.
\newblock Ward’s hierarchical agglomerative clustering method: Which
  algorithms implement {Ward’s} criterion?
\newblock \emph{Journal of Classification}, 31:\penalty0 274--295, 2014.
\newblock \doi{10.1007/s00357-014-9161-z}.
\newblock URL \url{https://doi.org/10.1007/s00357-014-9161-z}.

\bibitem[Pazzani et~al.(1994)Pazzani, Merz, Murphy, Ali, Hume, and
  Brunk]{Pazzani1994}
M.~Pazzani, C.~Merz, P.~Murphy, K.~Ali, T.~Hume, and C.~Brunk.
\newblock Reducing misclassification costs.
\newblock In \emph{Proc. of the 11th ICML}, San Francisco, 1994. Morgan
  Kauffmann.

\bibitem[Pfannschmidt et~al.(2019)Pfannschmidt, G{\"o}pfert, Neumann, Heider,
  and Hammer]{pfannschmidt2019fri}
Lukas Pfannschmidt, Christina G{\"o}pfert, Ursula Neumann, Dominik Heider, and
  Barbara Hammer.
\newblock Fri-feature relevance intervals for interpretable and interactive
  data exploration.
\newblock In \emph{2019 IEEE Conference on Computational Intelligence in
  Bioinformatics and Computational Biology (CIBCB)}, pages 1--10. IEEE, 2019.

\bibitem[Ramanan and Baker(2011)]{ramanan2010local}
Deva Ramanan and Simon Baker.
\newblock Local distance functions: A taxonomy, new algorithms, and an
  evaluation.
\newblock \emph{IEEE Transactions on Pattern Analysis and Machine
  Intelligence}, 33\penalty0 (4):\penalty0 794--806, 2011.
\newblock \doi{10.1109/TPAMI.2010.127}.

\bibitem[Royston et~al.(2011)Royston, White, et~al.]{royston2011multiple}
Patrick Royston, Ian~R White, et~al.
\newblock Multiple imputation by chained equations (mice): implementation in
  stata.
\newblock \emph{J Stat Softw}, 45\penalty0 (4):\penalty0 1--20, 2011.

\bibitem[Sato and Yamada(1996)]{sato}
A.~S. Sato and K.~Yamada.
\newblock Generalized learning vector quantization.
\newblock In \emph{Advances in Neural Information Processing Systems},
  volume~8, pages 423--429, 1996.

\bibitem[Schneider et~al.(2007)Schneider, Biehl, and Hammer]{Schneider2007}
P.~Schneider, M.~Biehl, and B.~Hammer.
\newblock Relevance matrices in learning vector quantization.
\newblock In M.~Verleysen, editor, \emph{Proc. of the 15th European Symposium
  on Artificial Neural Networks (ESANN)}, pages 37--43, Bruges, Belgium, 2007.
  d-side publishing.

\bibitem[Schneider et~al.(2009)Schneider, Biehl, and Hammer]{schneider2009}
Petra Schneider, Michael Biehl, and Barbara Hammer.
\newblock Adaptive relevance matrices in learning vector quantization.
\newblock \emph{Neural Computation}, 21\penalty0 (12):\penalty0 3532--3561,
  2009.
\newblock \doi{10.1162/neco.2009.11-08-908}.
\newblock URL \url{https://doi.org/10.1162/neco.2009.11-08-908}.
\newblock PMID: 19764875.

\bibitem[Schneider et~al.(2011)Schneider, Geweniger, Schleif, Biehl, and
  Villmann]{schneider2011multivariate}
Petra Schneider, Tina Geweniger, Frank-Michael Schleif, Michael Biehl, and
  Thomas Villmann.
\newblock Multivariate class labeling in {Robust Soft LVQ}.
\newblock In \emph{ESANN}, 2011.

\bibitem[Schulz et~al.(2020)Schulz, Hinder, and Hammer]{schulz2020deepview}
Alexander Schulz, Fabian Hinder, and Barbara Hammer.
\newblock Deepview: Visualizing classification boundaries of deep neural
  networks as scatter plots using discriminative dimensionality reduction.
\newblock In \emph{Proceedings of the Twenty-Ninth International Joint
  Conference on Artificial Intelligence, IJCAI}, 2020.

\bibitem[Scornet(2020)]{scornet2020trees}
Erwan Scornet.
\newblock Trees, forests, and impurity-based variable importance.
\newblock \emph{arXiv preprint arXiv:2001.04295}, 2020.

\bibitem[Severson et~al.(2017)Severson, Molaro, and
  Braatz]{severson2017principal}
Kristen~A Severson, Mark~C Molaro, and Richard~D Braatz.
\newblock Principal component analysis of process datasets with missing values.
\newblock \emph{Processes}, 5\penalty0 (3):\penalty0 38, 2017.

\bibitem[{Shirazi} et~al.(2012){Shirazi}, {Harandi}, {Sanderson}, {Alavi}, and
  {Lovell}]{Shirazi2012}
S.~{Shirazi}, M.~T. {Harandi}, C.~{Sanderson}, A.~{Alavi}, and B.~C. {Lovell}.
\newblock Clustering on {Grassmann} manifolds via kernel embedding with
  application to action analysis.
\newblock In \emph{2012 19th IEEE International Conference on Image
  Processing}, pages 781--784, 2012.
\newblock \doi{10.1109/ICIP.2012.6466976}.

\bibitem[Storbeck et~al.(2019)Storbeck, Schiffer, Baranowski, Chortis, Prete,
  Barnard, Gilligan, Taylor, Idkowiak, Arlt, et~al.]{storbeck2019steroid}
Karl-Heinz Storbeck, Lina Schiffer, Elizabeth~S Baranowski, Vasileios Chortis,
  Alessandro Prete, Lise Barnard, Lorna~C Gilligan, Angela~E Taylor, Jan
  Idkowiak, Wiebke Arlt, et~al.
\newblock Steroid metabolome analysis in disorders of adrenal steroid
  biosynthesis and metabolism.
\newblock \emph{Endocrine Reviews}, 40\penalty0 (6):\penalty0 1605--1625, 2019.

\bibitem[Tan et~al.(2016)Tan, Steinbach, and Kumar]{tan2016introduction}
Pang-Ning Tan, Michael Steinbach, and Vipin Kumar.
\newblock \emph{Introduction to data mining}.
\newblock Pearson Education India, 2016.

\bibitem[Tipping and Bishop(1999)]{tipping1999mixtures}
Michael~E Tipping and Christopher~M Bishop.
\newblock Mixtures of probabilistic principal component analyzers.
\newblock \emph{Neural Computation}, 11\penalty0 (2):\penalty0 443--482, 1999.

\bibitem[Tjoa and Guan(2021)]{XAItowardsMedicine}
Erico Tjoa and Cuntai Guan.
\newblock A survey on explainable artificial intelligence (xai): Toward medical
  xai.
\newblock \emph{IEEE Transactions on Neural Networks and Learning Systems},
  32\penalty0 (11):\penalty0 4793—4813, November 2021.
\newblock ISSN 2162-237X.
\newblock \doi{10.1109/tnnls.2020.3027314}.

\bibitem[Tse and Viswanath(2005)]{tse2005appendix}
D~Tse and P~Viswanath.
\newblock Appendix {B}: Information theory from first principles.
\newblock In \emph{Fundamentals of Wireless Communication}, pages 516--536.
  Cambridge University Press, 2005.

\bibitem[Turaga et~al.(2011)Turaga, Veeraraghavan, Srivastava, and
  Chellappa]{Turaga2011}
P.~Turaga, A.~Veeraraghavan, A.~Srivastava, and R.~Chellappa.
\newblock Statistical computations on {Grassmann} and {Stiefel} manifolds for
  image and video-based recognition.
\newblock \emph{IEEE Trans. Pattern Analysis and Machine Intelligence},
  33\penalty0 (11):\penalty0 2273--2286, 2011.
\newblock \doi{10.1109/TPAMI.2011.52}.

\bibitem[Van~Buuren(2018)]{van2018flexible}
Stef Van~Buuren.
\newblock \emph{Flexible imputation of missing data}.
\newblock CRC press, 2018.

\bibitem[van Veen(2016)]{NaNLVQ}
Rick van Veen.
\newblock Analysis of missing data imputation applied to heart failure data.
\newblock Masters thesis, University of Groningen, 2016.

\bibitem[Villmann et~al.(2018)Villmann, Kaden, Saralajew, and
  Villmann]{villmann2018probabilistic}
Andrea Villmann, Marika Kaden, Sascha Saralajew, and Thomas Villmann.
\newblock Probabilistic learning vector quantization with cross-entropy for
  probabilistic class assignments in classification learning.
\newblock In \emph{International Conference on Artificial Intelligence and Soft
  Computing}, pages 724--735. Springer, 2018.

\bibitem[Wagner(1990)]{wagner1990}
Gerold Wagner.
\newblock On means of distances on the surface of a sphere (lower bounds).
\newblock \emph{Pacific J. Math.}, 144\penalty0 (2):\penalty0 389--398, 1990.
\newblock URL \url{https://projecteuclid.org:443/euclid.pjm/1102645739}.

\bibitem[Wagner(1992)]{wagner1992}
Gerold Wagner.
\newblock On means of distances on the surface of a sphere. ii. upper bounds.
\newblock \emph{Pacific J. Math.}, 154\penalty0 (2):\penalty0 381--396, 1992.
\newblock URL \url{https://projecteuclid.org:443/euclid.pjm/1102635628}.

\bibitem[Wang et~al.(2020)Wang, Kaushal, and Khullar]{wang2020should}
Fei Wang, Rainu Kaushal, and Dhruv Khullar.
\newblock Should health care demand interpretable artificial intelligence or
  accept “black box” medicine?
\newblock \emph{Ann Intern Med}, 172\penalty0 (1):\penalty0 59--60, 2020.

\bibitem[Watson(1983)]{Watson1983}
Geoffrey~S. Watson.
\newblock \emph{Statistics on spheres}.
\newblock University of Arkansas lecture notes in the mathematical sciences.v.
  6. Wiley, New York, 1983.
\newblock ISBN 0471888664 (pbk.).
\newblock URL \url{http://hdl.handle.net/2027/mdp.39015017408140}.
\newblock "A Wiley-Interscience publication.".

\bibitem[Weinberger and Saul(2009)]{weinberger2009distance}
Kilian~Q Weinberger and Lawrence~K Saul.
\newblock Distance metric learning for large margin nearest neighbor
  classification.
\newblock \emph{Journal of machine learning research}, 10\penalty0 (2), 2009.

\bibitem[Wilson et~al.(2014)Wilson, Hancock, Pekalska, and
  Duin]{WilsonHPD14pami}
R.~C. Wilson, E.~R. Hancock, E.~Pekalska, and R.~P.~W. Duin.
\newblock Spherical and hyperbolic embeddings of data.
\newblock \emph{{IEEE} Trans. Pattern Anal. Mach. Intell.}, 36\penalty0
  (11):\penalty0 2255--2269, 2014.
\newblock \doi{10.1109/TPAMI.2014.2316836}.
\newblock URL \url{http://dx.doi.org/10.1109/TPAMI.2014.2316836}.

\end{thebibliography}

\end{document}